%% file: main.tex
\documentclass[acmtog,screen]{acmart} %

\copyrightyear{2025}
\acmYear{2025}
\setcopyright{cc}
\setcctype{by}
\acmConference[SA Conference Papers '25]{SIGGRAPH Asia 2025 Conference Papers}{December 15--18, 2025}{Hong Kong, Hong Kong}
\acmBooktitle{SIGGRAPH Asia 2025 Conference Papers (SA Conference Papers '25), December 15--18, 2025, Hong Kong, Hong Kong}
\acmDOI{10.1145/3757377.3763946}
\acmISBN{979-8-4007-2137-3/2025/12}

\usepackage{graphicx}
\usepackage{subfig}
\usepackage{multirow}
\usepackage{cleveref}
\usepackage{enumitem}

\newcommand{\OURS}{WorldExplorer}

\let\titleold\title
\renewcommand{\title}[1]{\titleold{#1}\newcommand{\thetitle}{#1}}
\def\maketitlesupplementary
   {
   \newpage
       \twocolumn[
        \centering
        \Large
        \textbf{\thetitle}\\
        \vspace{0.5em}Supplementary Material \\
        \vspace{1.0em}
       ] %
   }

\begin{document}

\input{tables/fig_teaser}

\author{Manuel-Andreas Schneider}
\authornote{denotes equal contribution. Website: \url{https://mschneider456.github.io/world-explorer}}
\affiliation{%
  \institution{Technical University of Munich}
  \country{Germany}
}
\email{manuel.schneider@tum.de}

\author{Lukas H\"ollein}
\authornotemark[1]
\affiliation{%
  \institution{Technical University of Munich}
  \country{Germany}
}
\email{lukas.hoellein@tum.de}

\author{Matthias Nie{\ss}ner}
\affiliation{%
  \institution{Technical University of Munich}
  \country{Germany}
}
\email{niessner@tum.de}

\title{\OURS: Towards Generating Fully Navigable 3D Scenes}

\ccsdesc[500]{Computing methodologies~Rendering}
\ccsdesc[500]{Computing methodologies~Volumetric models}
\ccsdesc[500]{Computing methodologies~Reconstruction}

\keywords{scene generation, video diffusion, 3D Gaussian Splatting}

\input{sections/0abstract}

\maketitle

\input{sections/1intro}
\input{sections/2rw}
\input{sections/3method}
\input{sections/4results}
\input{sections/5conclusion}

\begin{acks}
This project was funded by the ERC Consolidator Grant Gen3D (101171131) as well as the German Research Foundation (DFG) Research Unit ``Learning and Simulation in Visual Computing''.
We also thank Angela Dai for the video voice-over and Simon Giebenhain for help on the Gaussian overlay visualization.
\end{acks}

\bibliographystyle{ACM-Reference-Format}
\bibliography{sample-base}

\input{sections/Xsuppl}

\end{document}

%% file: tables/fig_teaser.tex
\begin{teaserfigure}
\centering
\setlength\tabcolsep{1pt}
\begin{tabular}{cc}
\includegraphics[width=0.779\textwidth]{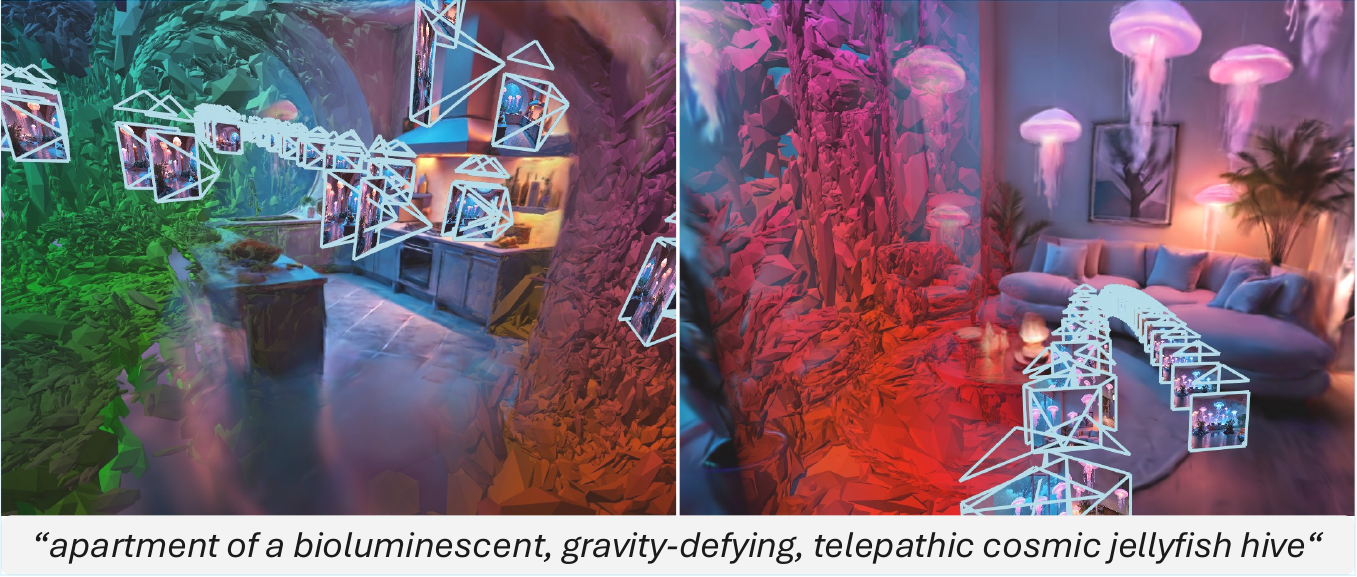} &
\includegraphics[width=0.220\textwidth]{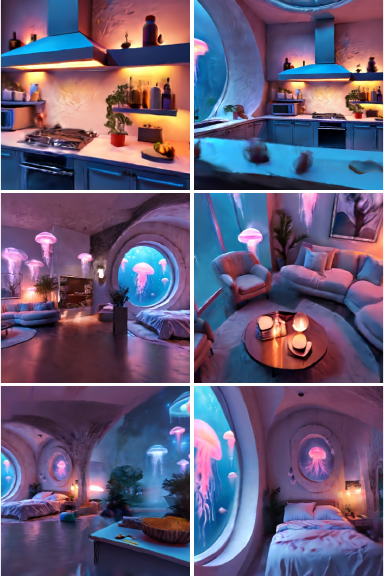} \\
(a) 3D Gaussian Splatting Generation from Text & (b) Rendered Novel Views
\end{tabular}
\caption{
\textbf{3D world generation from text input.} 
We exploit camera-guided video diffusion models~\cite{zhou2025stable} to iteratively generate multi-view observations of a scene.
To this end, we define multiple trajectories (cameras marked in blue) that explore different areas.
Finally, we reconstruct a 3D Gaussian Splatting scene~\cite{kerbl20233d} from all generated images (we visualize the individual Gaussians on top of the renderings).
Our scenes can be rendered under a large variety of viewpoints, i.e., they can be fully navigated beyond centered perspectives.
}
\Description{
\textbf{3D world generation from text input.} 
We exploit camera-guided video diffusion models~\cite{zhou2025stable} to iteratively generate multi-view observations of a scene.
To this end, we define multiple trajectories (cameras marked in blue) that explore different areas.
Finally, we reconstruct a 3D Gaussian Splatting scene~\cite{kerbl20233d} from all generated images (we visualize the individual Gaussians on top of the renderings).
Our scenes can be rendered under a large variety of viewpoints, i.e., they can be fully navigated beyond centered perspectives.
}
\label{fig:teaser}
\end{teaserfigure}

%% file: sections/0abstract.tex
\begin{abstract}
Generating 3D worlds from text is a highly anticipated goal in computer vision.
Existing works are limited by the degree of exploration they allow inside of a scene, i.e., produce streched-out and noisy artifacts when moving beyond central or panoramic perspectives.
To this end, we propose \OURS, a novel method based on autoregressive video trajectory generation, which builds fully navigable 3D scenes with consistent visual quality across a wide range of viewpoints.
We initialize our scenes by creating multi-view consistent images corresponding to a 360 degree panorama.
Then, we expand it by leveraging video diffusion models in an iterative scene generation pipeline.
Concretely, we generate multiple videos along short, pre-defined trajectories, that explore the scene in depth, including motion around objects.
Our novel scene memory conditions each video on the most relevant prior views, while a collision-detection mechanism prevents degenerate results, like moving into objects.
Finally, we fuse all generated views into a unified 3D representation via 3D Gaussian Splatting optimization.
Compared to prior approaches, {\OURS} produces high-quality scenes that remain stable under large camera motion, enabling for the first time realistic and unrestricted exploration.
We believe this marks a significant step toward generating immersive and truly explorable virtual 3D environments.
\end{abstract}

%% file: sections/1intro.tex
\section{Introduction}

High-quality and diverse 3D environments are a key requirement for applications ranging from AR/VR asset creation to computer graphics in gaming and film.  
As creating these worlds manually requires significant time and expertise, automated generation offers a scalable and accessible alternative for producing rich 3D content.  
The advent of text-to-image (T2I) and text-to-video (T2V) models \cite{flux2023, yang2024cogvideox} enables users to synthesize varied and visually compelling images directly from text input.  
Recent methods have begun leveraging these advances in 2D generation for 3D tasks such as object synthesis \cite{DreamFusion, AssetGen} and scene construction \cite{sun2024dimensionx, yu2024viewcrafter}. 
However, existing approaches to 3D scene generation still exhibit several limitations that hinder their practicality for world creation.

One of the core challenges in generating 3D environments is enabling users to \textit{freely explore} them in an interactive fashion.
That is, to create immersive surroundings, users must be able to navigate the scene and look around objects, all while maintaining realistic view synthesis.  
One line of existing methods leverages $\text{360}^\circ$ panoramas and converts them into 3D scenes \cite{yang2024layerpano3d, zhou2024dreamscene360}.  
Despite high-quality view synthesis near the scene's center, their ability to support broader exploration is very limited; occlusions and missing multi-view information result in distorted geometry and inconsistent appearance under larger viewpoint changes.  
Alternative approaches generate scenes through iterative \textit{render-refine-repeat} pipelines \cite{shriram2024realmdreamer, hollein2023text2room, yu2024wonderworld}, often relying on T2I models and monocular depth estimators \cite{yang2024depth}.  
While this enables them to produce entire scenes beyond panoramas, their iterative nature often leads to stretching or distortion artifacts, since the object geometry is expanded frame-by-frame.
More recently, methods using video diffusion models have demonstrated locally plausible object geometry~\cite{sun2024dimensionx, chen2025flexworld, ren2025gen3c}, i.e., their world prior over continuous motions helps to mitigate such artifacts.
However, the generated worlds cannot yet be interactively explored, as they either lack full $360^\circ$ scene coverage or exhibit distortion artifacts under novel views beyond centered perspectives.

To address these challenges, we propose a novel scene generation method leveraging camera-guided video generation, that enables free exploration of text-generated 3D environments (\Cref{fig:teaser}).
Our approach combines the strengths of panorama-based initialization, video diffusion models, and iterative scene generation. 
First, we construct a $360^\circ$ panorama from multiple images, describing separate areas of a diverse environment.
By utilizing depth estimation~\cite{yang2024depth} and T2I inpainting~\cite{flux2023}, we create this initial scene scaffold.
Next, we leverage video diffusion models in an iterative generation scheme to explore and expand the areas described by these initial observations.
Concretely, we generate multiple small videos, that move into different scene parts and thus provide more multi-view information about their contained objects.
We define the video trajectories \textit{a-priori}, i.e., a diverse set of directions and perspectives iteratively explores the entire $360^\circ$ scene.
To ensure view-consistency between the separate videos, we propose a scene memory mechanism.
It builds a conditional input for the video model from our scene scaffold and the most suitable views among all previously generated images.
We further develop a collision detection mechanism, that dynamically adjusts the trajectory length based on the actual generated content, i.e., we avoid degenerate outputs like moving the camera into walls or objects.
Finally, all generated views are fused into a single 3D scene representation via 3D Gaussian Splatting optimization \cite{kerbl20233d}.
We initialize this reconstruction by estimating a point cloud from all generated images using ~\cite{wang2025vggt} and rigidly align it with our generated cameras.
This enables interactive exploration of generated 3D worlds from arbitrary viewpoints in real-time based on rasterization.

\noindent To summarize, our contributions are:
\begin{itemize}[leftmargin=*,topsep=0pt, noitemsep]
    \item We introduce the first method for generating 3D scenes from text that supports high-quality view synthesis while enabling exploration across a wide range of camera poses.
    \item We propose an autoregressive scene expansion via video diffusion, driven by trajectory sampling and adaptive collision detection.
    \item We design a scene memory to condition each generation step on key prior frames, ensuring view consistency and scene coherence.
\end{itemize}

%% file: sections/2rw.tex
\section{Related Work}

\paragraph{Image- and Video-Generation Models}
Text-to-image (T2I) models can create high-quality, diverse images from text \cite{rombach2022high, SaharCSLWDGAMLSHFN2022, ramesh2022hierarchical}.
Their visual quality continuously improved \cite{podell2023sdxl, flux2023, xie2024sana, zhang2023text}.
As they are trained on billions of images \cite{schuhmann2022laion}, they represent strong 2D world priors, that can be adopted for many downstream tasks like inpainting/editing \cite{brooks2023instructpix2pix, lugmayr2022repaint}, semantic controls \cite{zhang2023adding, ye2023ip, mou2024t2i}, or personalization \cite{DreamBooth, LoRA}.
We leverage T2I models in our panorama initialization stage to create large and diverse environments.
Video models can generate worlds under continuously changing perspectives \cite{blattmann2023stable, yang2024cogvideox, seawead2025seaweed}.
A popular approach trains diffusion models \cite{ho2020denoising} via flow-matching \cite{lipman2022flow} with U-Net \cite{ronneberger2015u} or DiT \cite{peebles2023scalable, vaswani2017attention} backbones.
We leverage these models to build entire 3D scenes autoregressively.

\paragraph{3D Scene Generation}
Several approaches use 2D models as priors for 3D generation.
Score distillation sampling \cite{DreamFusion} optimizes for diverse objects \cite{AssetGen, chen2023fantasia3d, wang2023prolificdreamer} or scenes \cite{cohen2023set, zhang2024towards, li2024dreamscene}.
Others create multi-view consistent images for 3D reconstruction \cite{szymanowicz2025bolt3d, hollein2024viewdiff, gao2024cat3d}.
In order to create entire scenes, \cite{hollein2023text2room, schult2024controlroom3d, yu2024wonderjourney, shriram2024realmdreamer} utilize depth estimation \cite{yang2024depth, ke2023repurposing} and inpainting in an iterative generation pipeline \cite{liu2021infinite}.
Alternatively, \cite{wang2024perf, schwarz2025recipe, li2024scenedreamer360} turn $\text{360}^\circ$ panoramas into 3D Gaussian Splatting \cite{kerbl20233d} scenes.
Recently, camera-guided video diffusion models are utilized in similar pipelines \cite{sun2024dimensionx, yu2024viewcrafter, ren2025gen3c, liu2024reconxreconstructscenesparse}.
We leverage such video models in a novel iterative pipeline to generate large-scale 3D scenes that allow interactive exploration.

\paragraph{Camera-Controlled Video Generation}
Camera control in video generation is beneficial for world generation.
Recent approaches finetune models with additional pose input \cite{bahmani2024vd3d, bahmani2024ac3d, liang2024wonderland, he2025cameractrl, bai2025recammaster}.
Others render point clouds into novel views to condition the next frame generation \cite{liu2024novel, wang2025videoscene, zhang2025scene}.
While the motion of the videos can be precisely controlled, their length is often limited (e.g., small rotations around objects).
By using T2V models in a sliding-window fashion, \cite{song2025history, zhou2025stable} generate longer and consistent videos.
However, to create entire 3D scenes, it is crucial to generate consistent outputs even when revisiting previously generated areas or when viewing them from different perspectives.
To this end, we exploit these models in a novel iterative scheme.
Backed by a scene memory mechanism, we generate multiple small videos in a 3D-consistent fashion.

%% file: sections/3method.tex
\input{tables/fig_pipeline}

\section{Method}

Our method creates a 3D scene from text as input.
It is divided into three stages (see \Cref{fig:pipeline}).
First, we generate a panorama initialization to define a rough scaffold of the scene layout (\Cref{subsec:pano}).
Then, we utilize camera-guided video diffusion models (\Cref{subsec:cam-vid-diff}) to expand the scene.
The core idea of our approach is a scene generation scheme, which creates multiple small videos that progressively explore the scene (\Cref{subsec:iterative-gen}).
Each video contributes multi-view information about individual objects, i.e., we move ``inside and around'' of different parts of our scene scaffold. 
To ensure 3D-consistency between videos, we propose a novel scene memory mechanism, that selects optimal conditioning images for the next video, and a collision detection mechanism.
Finally, we reconstruct the entire 3D scene from all generated images (\Cref{subsec:3d-rec}).

\subsection{Review of Video Diffusion Models}
\label{subsec:cam-vid-diff}

Video diffusion \cite{blattmann2023stable, yang2024cogvideox, seawead2025seaweed} models the conditional probability distribution $p_{\theta}(\mathbf{I}^{\text{tgt}} | \mathbf{I}^{\text{src}})$ over data $(\mathbf{I}^\text{tgt}, \mathbf{I}^\text{src}) {\sim} q(\mathbf{V})$, where $\mathbf{V} {=} \{ \mathbf{I}_1, \mathbf{I}_2, ..., \mathbf{I}_N \}$ is a dataset of continuous videos with $\mathbf{I} {\in} \mathbb{R}^{3 \times H \times W}$ being the individual images.
To achieve more control over the generated video samples, a camera condition $(\mathbf{\pi}^\text{src}, \mathbf{\pi}^\text{tgt})$ is added to each image, which typically is encoded into dense Plücker embeddings \cite{zhou2025stable, liang2024wonderland}.
The network $\theta$ then parameterizes the conditional diffusion denoising probability $p_{\theta}(\mathbf{I}^{\text{tgt}} | \mathbf{c}) {=} \int p_{\phi}(\mathbf{I}^{\text{tgt}}_{0{:}T} | \mathbf{c})d\mathbf{I}^{\text{tgt}}_{1{:}T}$ with $\mathbf{c} {=} (\mathbf{I}^{\text{src}}, \mathbf{\pi}^\text{src}, \mathbf{\pi}^\text{tgt})$ and we progressively add more Gaussian noise to the latent variables $\mathbf{I}^{\text{tgt}}_{1{:}T}$ following \cite{ho2020denoising, lipman2022flow}.
We leverage the recent camera-guided video diffusion model from \cite{zhou2025stable} as a strong prior over the real world.
In other words, we exploit the model to generate entire 3D scenes autoregressively, i.e., we generate a large set of images expressed as multiple shorter videos $\{\mathbf{\hat{V}_1}, ..., \mathbf{\hat{V}_M}\}$.
To ensure high-quality and 3D-consistent video generation, we develop a novel conditioning mechanism that selects the optimal $\mathbf{c}_k$ from all previously generated data $\{\mathbf{\hat{V}_1}, ..., \mathbf{\hat{V}_{k-1}}\}$ (see \Cref{subsec:iterative-gen}).

\subsection{Panorama Initialization}
\label{subsec:pano}

In the first step of our method, we create a panoramic overview of the scene (see \Cref{fig:pipeline} left).
We utilize the T2I model Flux~\cite{flux2023} to generate four initial images, denoted as $\{ \mathbf{I}_i \}_{i=1}^4$.
Each of these images describes separate areas of the scene, e.g., a kitchen or a living room.
Then, we define four fixed camera poses, each consisting of a rotation $R_i {\in} \mathrm{SO}(3)$ and a translation $t_i {\in} \mathbb{R}^3$, with $t_i {=} \mathbf{0}$ for all $i$. 
The rotations correspond to uniformly distributed yaw angles, i.e.,
\begin{equation}
\label{eq:pano-rot}
    R_i = R_y(\theta_i), \quad \theta_i \in \{0^\circ, 90^\circ, 180^\circ, 270^\circ\},
\end{equation}
where $R_y(\theta)$ denotes a rotation of angle $\theta$ about the global $y$-axis. 
In other words, each camera is oriented outward from the origin.
Next, we utilize these poses together with a pretrained monocular depth estimator~\cite{yang2024depth} to unproject each image into a point cloud as $\mathcal{P}_i = R_i K^{-1}[u, v, 1]^\top \cdot \mathbf{D}_i(u, v)$, where $\mathbf{K}$ are the pinhole camera intrinsics with a field of view of $60^\circ$ and $\mathbf{D}_i$ is the estimated depth map for image $\mathbf{I}_i$.
We now similarly define four additional camera poses at $\theta_i \in \{45^\circ, 135^\circ, 225^\circ, 315^\circ\}$ and project the point clouds into these views using point splatting~\cite{ravi2020accelerating, wiles2020synsin}.
Finally, we inpaint the unobserved regions using ~\cite{flux2023} to obtain our panoramic scene overview $\{ \mathbf{I}_i \}_{i=1}^8$.

In contrast to existing text-to-panorama generators like ~\cite{Tang2023mvdiffusion, yang2024layerpano3d}, this approach can create more diverse scenes.
For example, the four initial images can represent drastically different areas (e.g., an indoor kitchen and outdoor forest), whereas panorama generators tend to produce $360^\circ$ views of a single scene type only (see \Cref{fig:ours-baselines} and the supplementary material).
It also offers users an easy interface for interactive generation of diverse scenes by only having to generate and arrange four images according to personal preferences.

\subsection{Iterative Video Trajectory Generation}
\label{subsec:iterative-gen}
\input{tables/fig_scene_mem}

Given the initial scene scaffold $\{ \mathbf{I}_i \}_{i=1}^8$, we create additional observations with the video diffusion model.
Concretely, we propose an iterative generation scheme: starting from an initial image, we utilize four pre-defined trajectories $\{ \mathbf{\pi}^\text{tgt}_k \}_{k=1}^4$ that explore the scene by moving ``into and around'' that area.
In other words, we sequentially generate the videos $\mathbf{\hat{V}}_{4i+k} {\sim} p_{\theta}(\mathbf{I}^{\text{tgt}} | \mathbf{I}^{\text{src}}, \mathbf{\pi}^\text{src}, \mathbf{\pi}^\text{tgt}_k)$ for all $i,k$, where suitable condition frames $\mathbf{I}^{\text{src}}$ are selected per video from all previous frames.
We iterate this principle multiple times to create in total 32 videos.
Next, we explain the generation in more detail.

\paragraph{Trajectory Sampling}

Camera-guided video diffusion models such as ~\cite{zhou2025stable} are powerful priors for generating \textit{locally plausible} multi-view observations (e.g., rotating around objects).
In contrast, generating entire 3D scenes along a single, \textit{large} camera path remains challenging due to several reasons.
First, defining a continuous path a-priori, that covers an entire 3D scene is ambiguous.
In other words, we have to adapt the trajectory on-the-fly based on the generated content (e.g., prevent running into walls, resolve occlusions, discover novel generated areas).
Second, to achieve 3D-consistency in large 3D worlds, the model must memorize previously generated objects/structures across a large history of previous frames.
While approaches like ~\cite{zhou2025stable, song2025history} show promising results, their models still suffer from inconsistencies, collisions, and catastrophic forgetting, when re-visiting previously observed areas (also see \Cref{fig:ours-baselines} and the supplementary material).

To this end, we rely on a different strategy to turn video diffusion models into 3D-consistent generators for large 3D scenes.
Inspired by iterative generation pipelines like ~\cite{liu2021infinite, hollein2023text2room}, we split scene generation into multiple smaller trajectories.
Concretely, our four pre-defined trajectories exploit video diffusion models to generate additional observations within a specific local scene area, starting from one of the 8 initial panoramic images.
As can be seen in \Cref{fig:scene-mem}, we create camera paths that zoom-in along a straight line (1), rotate to the left/right (2,3) and elevate up/look down (4).
These trajectories allow us to rotate around arbitrary generated content in a brute-force fashion, i.e., we use the same camera paths starting from each initial image.

Crucially, resolving collisions becomes a much easier task this way: since we already split scene generation into discontinuous and small camera paths, we only need to prevent such degenerate cases for each trajectory separately.
In other words, the next trajectory is not impacted by a collision in previous trajectories: it can still be generated starting from the scene scaffold.
Concretely, we first generate the entire video and then utilize a video depth estimator~\cite{video_depth_anything} to predict normalized depth maps $\mathbf{\hat{D}} \in [0,1]^{N \times H \times W}$ for all generated frames.
We define a fixed area of interest as a binary mask $\mathbf{M} \in \{0,1\}^{H \times W}$, where pixels $(u, v)$ within a center crop of extent $(w, h)$ are set to 1:
\begin{equation}
\label{eq:mask}
\mathbf{M}(u,v) = 
\begin{cases}
1 & \text{if } u \in \left[\frac{W - w}{2}, \frac{W + w}{2}\right),\ v \in \left[\frac{H - h}{2}, \frac{H + h}{2}\right) \\
0 & \text{otherwise}
\end{cases}
\end{equation}
We then compute a masked average depth per frame \( t {\in} \{1 \dots N\} \) as
\begin{equation}
\bar{d}_t = \frac{\sum_{u,v} \mathbf{M}(u,v) \cdot \mathbf{\hat{D}}_t(u,v)}{\sum_{u,v} \mathbf{M}(u,v)}.
\end{equation}
We discard the trajectory from the first frame $t^*$ onward for which $\bar{d}_{t^*} < 0.4$, and retain only the generated frames before that point.
Empirically, we define the center crop to extend to 20\% of the image resolution.
This avoids ending trajectories too early if they move closely besides objects/walls, which usually leads to low depth values on the edges of images.

\paragraph{Scene Memory}

Our scene generation is divided into multiple short video segments, denoted as $\{\mathbf{\hat{V}}_{4i+k}\}_{i,k}$, where each segment corresponds to a separate camera trajectory.
To ensure 3D-consistency across videos, we introduce a novel scene memory mechanism (summarized in \Cref{fig:scene-mem}).
Assuming each trajectory produces $N$ frames, we further divide them into video batches of $21$ frames.
Similar to~\cite{zhou2025stable}, we condition each batch on a set of $13$ previously generated frames. 
The first 8 are fixed and correspond to the initial panoramic images $\{ \mathbf{I}_i \}_{i=1}^8$, which ensures global scene layout awareness at all times.
The remaining 5 context frames are dynamically selected based on rotational similarity to all previously generated frames across earlier trajectories.
Concretely, we select the top-5 nearest camera poses w.r.t. all $21$ frames in a manner similar to~\cite{zhou2025stable}.
Different from their approach, we consider all previously generated images across all trajectories as potential candidates.
Additionally, we leverage our known trajectory layout to discard cameras, that are directly opposite to the initial image of the current trajectory.
After the video generation completed, we add the newly generated images and their associated cameras into our scene memory, i.e, the next trajectory can be conditioned on them.
For improved diversity, we avoid saving the first two generated frames of any trajectory, as their images are still similar to the initial panoramic images.

Our method leverages the two-stage video generation approach of~\cite{zhou2025stable}, where anchor frames are first generated uniformly along the trajectory, followed by interpolation between these anchors.
Our scene memory mechanism ensures consistency of the anchor frames across trajectories.
The subsequent interpolation then naturally produces scene-consistent intermediate frames.

\subsection{3D Scene Optimization}
\label{subsec:3d-rec}

The final step of our method is to reconstruct the 3D scene from all generated images.
To this end, we utilize 3D Gaussian Splatting (3DGS) \cite{kerbl20233d} as our scene representation, since it offers real-time, high-quality novel-view-synthesis (NVS) capabilities on standard computer graphics pipelines.
This makes it possible to create immersive and fully navigable 3D scene experiences.

First, we obtain a point cloud initialization from our generated images.
In traditional NVS, Structure-from-Motion typically provides such a point cloud when regressing camera poses from the input images.
However, we generate more than 1K images, which makes running methods like COLMAP~\cite{schonberger2016structure} prohibitively slow.
To this end, we utilize the recent approach VGGT~\cite{wang2025vggt} to obtain a sparse point cloud from our generated images.
We downsample the prediction to 200K points and use those as initialization for 3DGS training.

We perform an additional alignment step to our known camera poses, since VGGT predicts the point cloud $\mathcal{\hat{P}}_{GS}$ and the corresponding camera poses in its own world coordinate system.
First, we obtain the rigid transformation between VGGT and our coordinate system by comparing the predicted camera pose of $\mathbf{I}_1$ with our known pose: $\mathbf{\pi}_t {=} \mathbf{\pi}_1 \mathbf{\hat{\pi}_1}^{-1}$.
Next, we calculate a scaling factor as $\mathbf{s} {=} \frac{\text{hull}(\{ \mathbf{\pi_i} \}_{i=1}^N)}{\text{hull}(\{ \mathbf{\hat{\pi}_i} \}_{i=1}^N)}$, where $\textit{hull}$ denotes the length of the bounding box around all camera centers.
Finally, we transform the VGGT point cloud into our known world coordinate system as $\mathcal{P}_{GS} = \mathbf{s} \cdot \mathbf{\pi}_t \cdot \mathcal{\hat{P}}_{GS}$.

We optimize the 3DGS scene following the loss function introduced in ~\cite{kerbl20233d}:
\begin{equation}
\label{eq:3dgs-loss}
\mathcal{L}(\mathbf{x}) {=} \frac{1}{N} \sum_{i=1}^{N} (\lambda_{1} |\mathbf{\hat{I}_i} {-} \mathbf{I_i}| {+} \lambda_{2} (1 {-} \text{SSIM}(\mathbf{\hat{I}_i}, \mathbf{I_i})))
\end{equation}
where $\mathbf{\hat{I}_i}$ is the rendered image from the 3DGS scene, $\mathbf{I_i}$ our generated frames from the second stage, and $\lambda_{1}, \lambda_{2}$ the loss weights.

%% file: tables/fig_pipeline.tex
\begin{figure*}[ht]
\centering
\includegraphics[width=\linewidth]{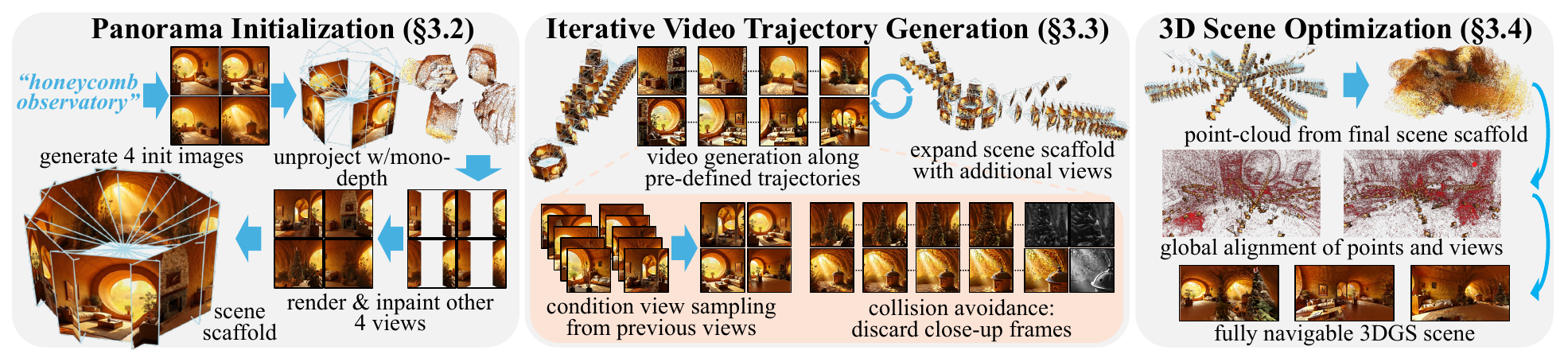}
\caption{\textbf{Method overview}.
We generate a fully navigable 3D scene from text input in three stages.
\textbf{Left}: we create four initial images with a T2I model~\cite{flux2023}, and we define that they look outwards, distributed uniformly around the scene center (without overlap).
We leverage monocular depth estimators~\cite{yang2024depth} and inpainting to create the remaining four images of our scene scaffold.
\textbf{Mid}: we propose a novel iterative video trajectory generation pipeline.
It exploits camera-guided video diffusion models~\cite{zhou2025stable} to generate additional images of the scene along pre-defined trajectories.
For each video generation, we select optimal conditioning views from all previously generated frames, which encourages 3D-consistent content generation.
We further discard frames (marked as grey), that are too close to (running into) objects to avoid degenerate outputs.
We repeat this trajectory generation multiple times to obtain a final set of posed images.
\textbf{Right}: we optimize a 3D Gaussian-Splatting scene~\cite{kerbl20233d} from our generated images.
To this end, we first obtain a point-cloud initialization using~\cite{wang2025vggt}.
Then we align the predicted points to our camera poses via rigid transformation and scaling.
Finally, we can explore the scene, i.e., synthesize novel views from perspectives beyond the scene center, in real-time using standard computer graphics rasterization.
}
\Description{\textbf{Method overview}.
We generate a fully navigable 3D scene from text input in three stages.
\textbf{Left}: we create four initial images with a T2I model~\cite{flux2023}, and we define that they look outwards, distributed uniformly around the scene center (without overlap).
We leverage monocular depth estimators~\cite{yang2024depth} and inpainting to create the remaining four images of our scene scaffold.
\textbf{Mid}: we propose a novel iterative video trajectory generation pipeline.
It exploits camera-guided video diffusion models~\cite{zhou2025stable} to generate additional images of the scene along pre-defined trajectories.
For each video generation, we select optimal conditioning views from all previously generated frames, which encourages 3D-consistent content generation.
We further discard frames (marked as grey), that are too close to (running into) objects to avoid degenerate outputs.
We repeat this trajectory generation multiple times to obtain a final set of posed images.
\textbf{Right}: we optimize a 3D Gaussian-Splatting scene~\cite{kerbl20233d} from our generated images.
To this end, we first obtain a point-cloud initialization using~\cite{wang2025vggt}.
Then we align the predicted points to our camera poses via rigid transformation and scaling.
Finally, we can explore the scene, i.e., synthesize novel views from perspectives beyond the scene center, in real-time using standard computer graphics rasterization.
}
\label{fig:pipeline}
\end{figure*}

%% file: tables/fig_scene_mem.tex
\begin{figure}
\centering
\includegraphics[width=0.5\textwidth]{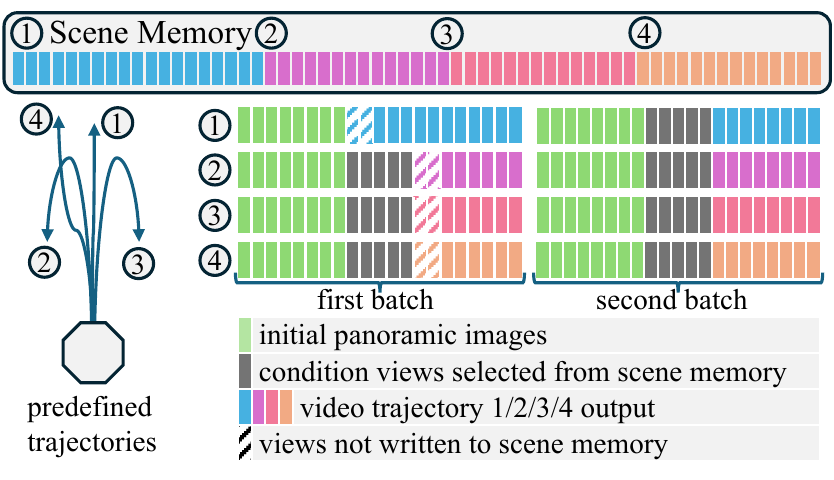}
\caption{\textbf{Conditional video generation via scene memory}. 
We sequentially generate videos in multiple batches of 21 frames along 4 trajectories (blue, purple, red, orange).
These trajectories start from each of our initial panoramic images (bottom left).
To ensure 3D-consistent outputs, we prefill the first 8 frames with our initial scene scaffold (marked in green).
The remaining 5 are sampled based on rotational similarity from our scene memory, which stores the images and poses of all previously generated frames (top).
After each trajectory is generated, we append all frames starting from the third image to our scene memory for conditioning of later frames.
}
\Description{\textbf{Conditional video generation via scene memory}. 
We sequentially generate videos in multiple batches of 21 frames along 4 trajectories (blue, purple, red, orange).
These trajectories start from each of our initial panoramic images (bottom left).
To ensure 3D-consistent outputs, we prefill the first 8 frames with our initial scene scaffold (marked in green).
The remaining 5 are sampled based on rotational similarity from our scene memory, which stores the images and poses of all previously generated frames (top).
After each trajectory is generated, we append all frames starting from the third image to our scene memory for conditioning of later frames.
}
\label{fig:scene-mem}
\end{figure}

%% file: sections/4results.tex
\section{Results}

\paragraph{Implementation Details}

We generate all results with our default trajectory sampling scheme of 8 initial start images and 4 pre-defined trajectories.
All 32 videos are generated sequentially in a clockwise order of starting images. 
Within each we first generate the ``zoom-in'' trajectory (1) followed by ``left'' (2), ``right'' (3), and ``up'' (4).
We generate 44 frames per video for the trajectory types (1) and (4), and 134 frames for (2) and (3), respectively.
Depending on the collision detection, we generate between 1.5K-2.5K images per scene at resolution $576 \times 576$.
The complete three-stage generation process takes on average 7h on a single RTX3090 GPU.
Our runtime is dominated by the second stage (\textasciitilde 10min/trajectory), while the first stage (\textasciitilde 5min) and third stage (\textasciitilde 11min) are comparatively faster.

We experiment with multiple different prompts and scene types in our main results.
Concretely, we generate prompts with LLMs and use a different room type for each of the four start images (e.g., living room, bathroom, bedroom, office, kitchen) for our indoor scene results.
We also demonstrate the capability of our method to generate outdoor scenes in the supplementary material.

\paragraph{Baselines}
We compare against multiple types of text-to-3D scene generation methods.
We provide additional details about baselines in the supplementary material.
\begin{itemize}[leftmargin=*,topsep=0pt, noitemsep]

    \item \emph{Panorama-To-3D}: we compare against DreamScene360~\cite{zhou2024dreamscene360} and LayerPano3D~\cite{yang2024layerpano3d} as state-of-the-art methods that first generate a $360^\circ$ panorama from text and then turn it into a 3DGS~\cite{kerbl20233d} scene.

    \item \emph{Iterative T2I Lifting}: we compare against Text2Room~\cite{hollein2023text2room} and WonderWorld~\cite{yu2024wonderworld}, which generate 3D scenes through a render-refine-repeat pattern based on T2I models and monocular depth estimators. We provide one of our four start images as input to these methods for fair comparisons.
    
    \item \emph{Camera-Guided Video Diffusion}: most related to our approach, we compare against FlexWorld~\cite{chen2025flexworld} and SEVA~\cite{zhou2025stable}. We choose them over other existing models, as they have demonstrated \textit{complete} $360^\circ$ scene generation with video diffusion (instead of only locally looking around objects or scene parts). We provide one of our four start images as input to both methods and add our 3DGS post-optimization step to SEVA.
\end{itemize}

\paragraph{Metrics}

We evaluate our generated 3D scenes both qualitatively and quantitatively.
We calculate CLIP Score (CS)~\cite{radford2021learning} and Inception Score (IS)~\cite{salimans2016improved} on RGB renderings from novel perspectives.
For each scene, we sample cameras \textit{beyond} panoramic/centered perspectives, to assess the quality under arbitrary viewpoints.
Additionally, we conduct a user study and ask $\text{n}{=}64$ participants about the Perceptual Quality ($PQ$) and the 3D-Consistency ($3DC$).
We present them with rendered videos that move throughout the whole 3D scene from novel (non-centered) perspectives and ask to score the scenes on a scale of $1{-}5$.

\subsection{Qualitative Comparisons}
\input{tables/fig_ours_baselines}

We compare the quality of generated scenes against all baselines in \Cref{fig:ours-baselines}.
Also, we provide additional results of our method in \Cref{fig:ours-scenes} and the supplementary material.
In both cases, we split the visualization into two parts.
First, we visualize a scene overview located at the center of the scene to evaluate the quality and diversity of generated 3D scenes.
Second, we render novel views that explore the scene, i.e., they are rotated around objects and moved away from the center.
This allows us to showcase the ability of each method to fully navigate their generated scenes.

The panorama-based methods DreamScene360~\cite{zhou2024dreamscene360} and LayerPano3D~\cite{yang2024layerpano3d} create compelling $360^\circ$ views, but they do not hallucinate unobserved regions beyond the panoramic viewpoint.
Thus, their novel views contain visible distortions and occlusions.
Text2Room~\cite{hollein2023text2room} and WonderWorld~\cite{yu2024wonderworld} create high-detail textures due to their iterative unprojection steps, but suffer from error accumulation, i.e., their object geometry appears stretched out due to erroneous depth estimation.
FlexWorld~\cite{chen2025flexworld} creates complete 3D scenes with compelling hallucinations (e.g., the furnace).
However, their trajectory-based generation does not cover large viewpoint changes, and as such novel views rotated around objects still exhibit missing and distorted geometry (e.g., behind the couch/table).
SEVA~\cite{zhou2025stable} uses a single, continuous trajectory to generate an entire 3D scene given a single image as input.
This enables them to generate novel geometry, that matches the style of the input image.
However, the generated video frames contain noticeable flickering artifacts (see suppl. material), which become visible in the 3DGS renderings as duplicated objects (e.g., multiple overlayed couches/tables) or floating artifacts (e.g., the bottom-left of the overview rendering).

This showcases the advantage of our three-stage approach: by conditioning the video diffusion model on 8 panoramic images, we generate high-quality and fitting novel views for the entire $360^\circ$ scene layout.
Similar to ~\cite{chen2025flexworld, zhou2025stable} our video diffusion stage generates locally plausible multi-view information.
Additionally, our novel scene memory and iterative trajectory generation lift this to a complete 3D scene along multiple, discontinuous trajectories.
Overall, this leads to high-quality renderings even for novel views far beyond the scene center.
This enables users to interactively explore our environments from arbitrary camera viewpoints.
Please see the suppl.~ material for more scenes, animated video results, and intermediate baseline outputs (e.g., video frames from SEVA~\cite{zhou2025stable} and panoramas from DreamScene360~\cite{li2024dreamscene} and LayerPano3d~\cite{yang2024layerpano3d}).

\subsection{Quantitative Comparisons}
\input{tables/tab_ours_baselines}
We present the quantitative results averaged across the novel view renderings of multiple scenes in ~\Cref{tab:ours-baseline}.
Our approach achieves the highest CLIP score, since our generated geometry remains consistent towards the scene prompt even beyond panoramic perspectives.
This highlights the advantage of our panorama initialization: the scene scaffold is already created from the beginning, which makes inpainting around it easier than for the iterative T2I and video diffusion baselines, that have to hallucinate the novel content completely from scratch.
We obtain competitive Inception scores, which signals that our generated images are as sharp as baseline methods.
Users give the highest ratings to our method, which highlights the quality of our generated 3D scenes under challenging, novel perspectives.
We also provide a runtime and memory comparison against baselines in the supplementary material.

\subsection{Ablations}
\input{tables/fig_ours_ablation}

The key components of our method are the panorama initialization (\Cref{subsec:pano}) and iterative video trajectory generation, utilizing our novel scene memory and collision detection mechanisms (\Cref{subsec:iterative-gen}).
We demonstrate their importance in \Cref{fig:ours-ablation} and \Cref{tab:ours-baseline}.

\paragraph{How important is the panorama initialization?}
We generate multiple small videos along pre-defined trajectories, that start from each of the 8 initial panoramic images.
We ablate the importance of having 8 initial images by only providing 4 or 1 of them as condition during video generation (but still generate all 32 trajectories).
Scene diversity and sharpness drop noticeably, i.e., generating entire scene parts from scratch is harder than inpainting around our initial scene scaffold.
Concretely, when providing only a single panoramic image, the video diffusion model needs to hallucinate entire new areas in a 3D-consistent fashion.
This oftentimes leads to oversimplified outputs (if using a low guidance scale) or oversaturated frames (if using a high guidance scale).
While providing 4 images helps to retain large scenes, their 3D-consistency severely drops, as the individual trajectories are less well-conditioned.
This results in blurriness and floating artifacts in the 3D scene reconstruction.

\paragraph{How important is the scene memory?}
We propose a novel scene memory mechanism, that conditions each video generation on the most suitable previous frames across all trajectories.
We ablate this choice by replacing it with the scene memory of~\cite{zhou2025stable}, i.e., we only select condition frames across the individual batches \textit{within} each trajectory.
Since we design multiple small camera paths instead of one large trajectory, the amount of multi-view information beyond the local areas of each trajectory is very limited under this setup.
As a result, each video generates slightly different object types/positions.
We can observe severe degradation of our scenes, as the lack of 3D-consistency results in blurry reconstructions.

\paragraph{How important is the collision detection?}
Our collision detection mechanism avoids degenerate videos due to running into generated content like walls.
It is a necessary component, as we define our trajectories a-priori, i.e., we always use the same trajectories regardless of the actual generated content.
Without it, collisions may occur, which lead to blurry edges around those areas.
For example, the video generation ran into the archway in front of the kitchen, which makes its edges fuzzy when observed from up-close.
Since collisions do not necessarily appear for every trajectory, other areas like the kitchen remain unaffected of these local degenerations.
This further signals the robustness of our iterative generation: later trajectories are not adversely impacted by collisions of previous ones.

\subsection{Limitations}
\input{tables/fig_ours_scenes_updated}
Our method generates high-quality 3D scenes from text as input, that can be fully explored, i.e., we can interactively move into scene parts and rotate around objects without quality degradation.
Nevertheless, there are remaining limitations (see supplementary material).
Our video diffusion model sometimes produces high-frequency flickering artifacts across trajectories, which manifest in blurry renderings.
Extending our 3D scene optimization with uncertainty~\cite{goli2024bayes} or robustness~\cite{sabour2023robustnerf} weighting schemes specifically tailored for these common problems of video generators may further detect and ignore such outliers.
Furthermore, our method uses a set of pre-defined trajectories, which explore and expand the initial scene scaffold.
While it proved robust towards a large variety of room and prompt types, it limits the layout of the generated scenes.
Incorporating semantic layouts as condition~\cite{schult2024controlroom3d, zhang2023adding}, enabling interactive and unlimited scene generations, and generating dynamic 4D worlds are interesting future directions.
Lastly, our approach generates 3D scenes with baked-in material and lighting effects, since our base diffusion models operate in RGB space.
Recent text-to-x generators like~\cite{kocsis2025intrinsix, ke2025marigold} have the potential to replace RGB-space models in downstream tasks such as scene generation.

%% file: tables/fig_ours_baselines.tex
\begin{figure*}
\centering
\setlength\tabcolsep{1pt}         
\renewcommand{\arraystretch}{1}   

\begin{tabular}{cccccc}

& Generated scene overview & \multicolumn{3}{c}{Rendered novel views} \\

\raisebox{0.15\height}{\rotatebox{90}{\parbox{2cm}{\centering DreamScene360 \\ \cite{zhou2024dreamscene360}}}} &
\includegraphics[height=0.15\textwidth]{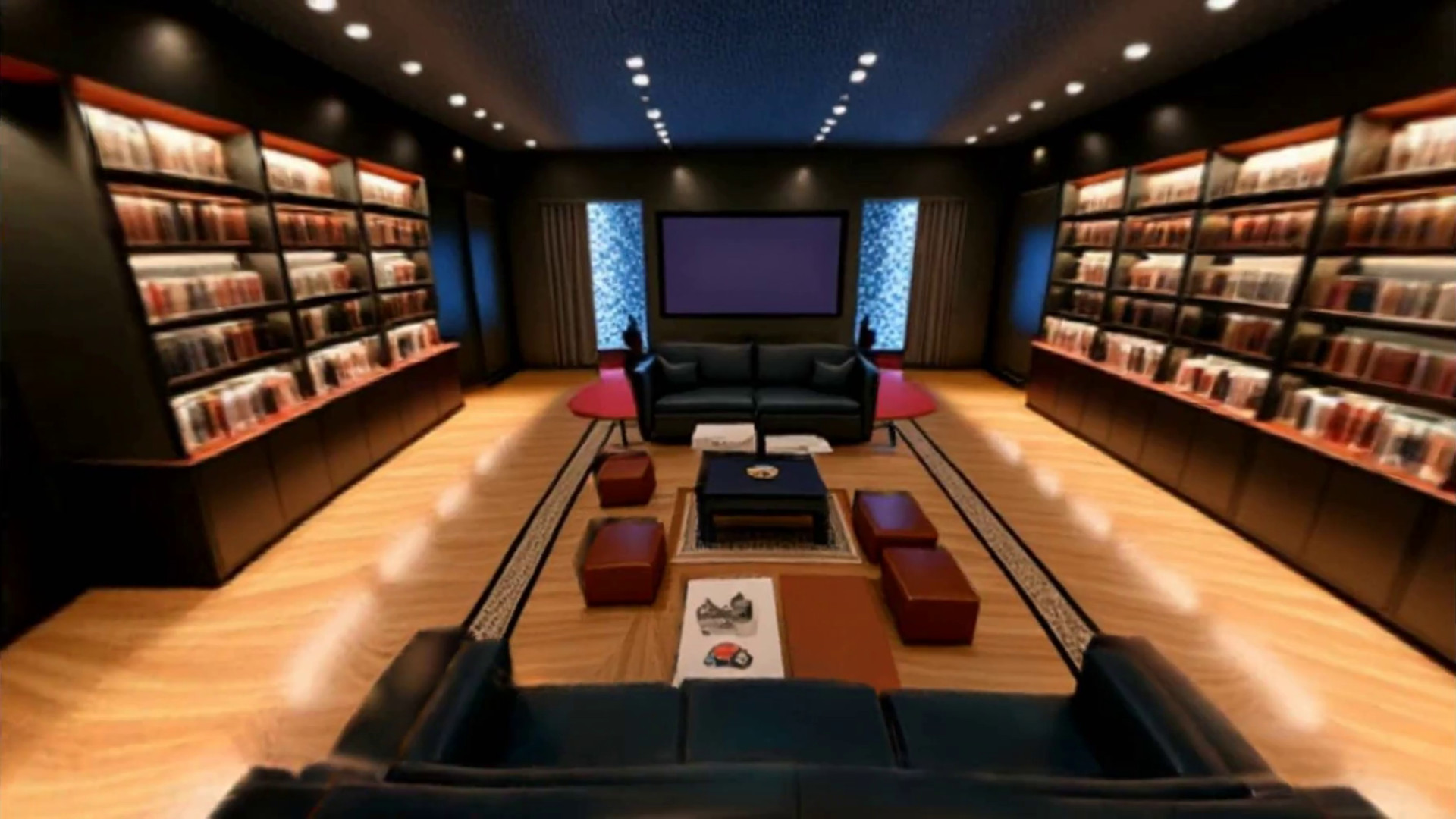} &
\includegraphics[height=0.15\textwidth]{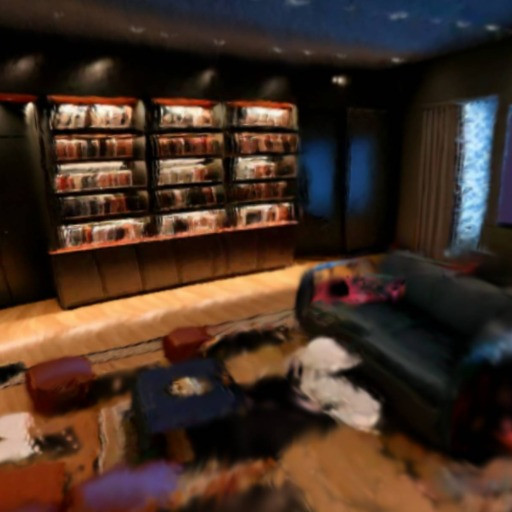} &
\includegraphics[height=0.15\textwidth]{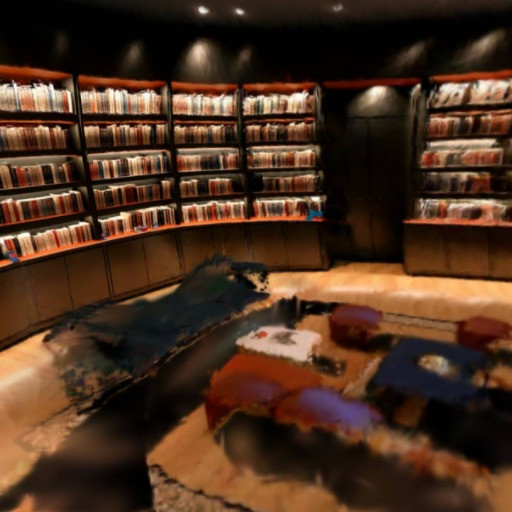} &
\includegraphics[height=0.15\textwidth]{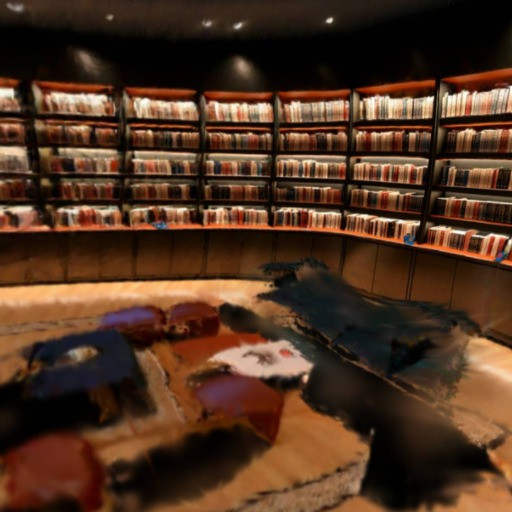} &
\includegraphics[height=0.15\textwidth]{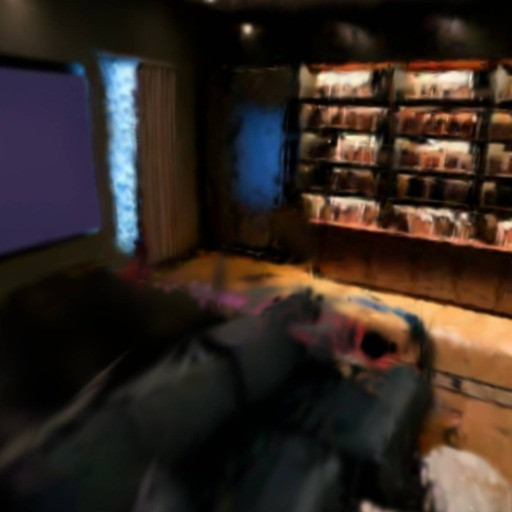} \\

\raisebox{0.15\height}{\rotatebox{90}{\parbox{2cm}{\centering LayerPano3D \\ \cite{yang2024layerpano3d}}}} &
\includegraphics[height=0.15\textwidth]{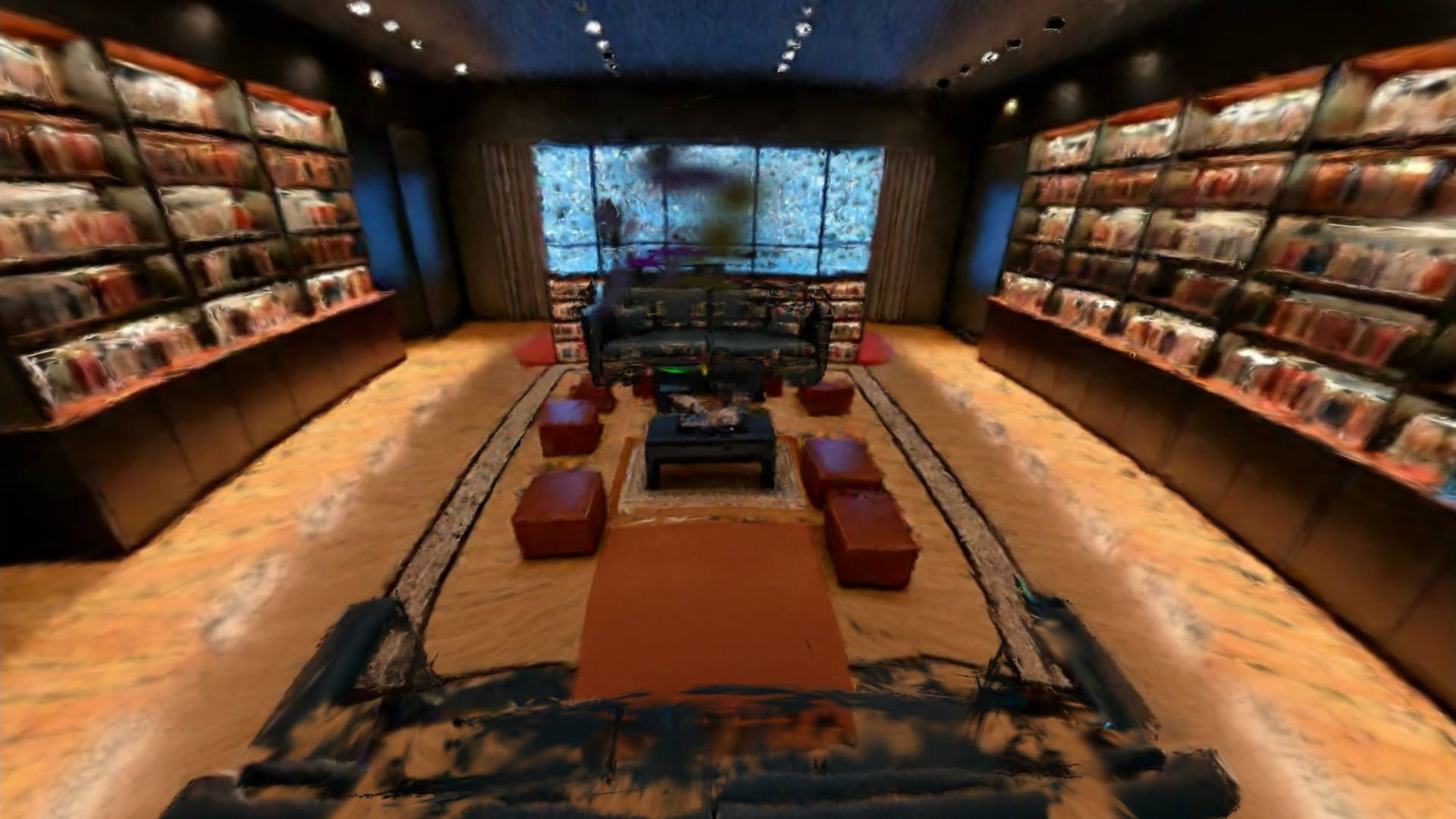} &
\includegraphics[height=0.15\textwidth]{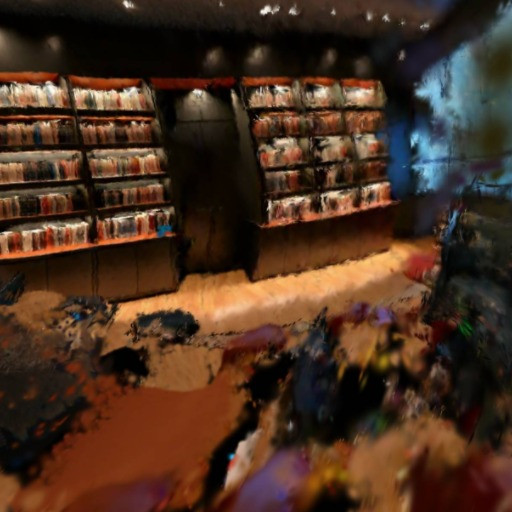} &
\includegraphics[height=0.15\textwidth]{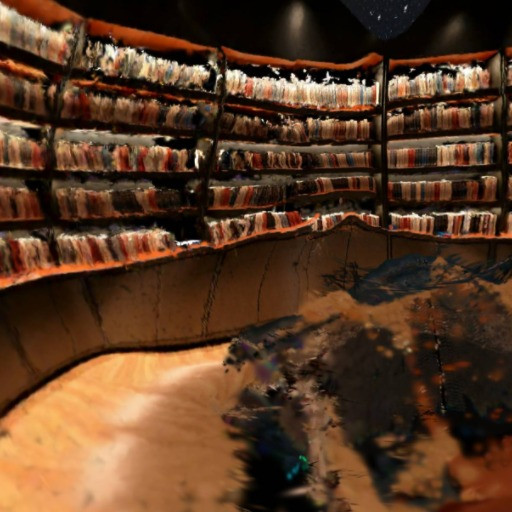} &
\includegraphics[height=0.15\textwidth]{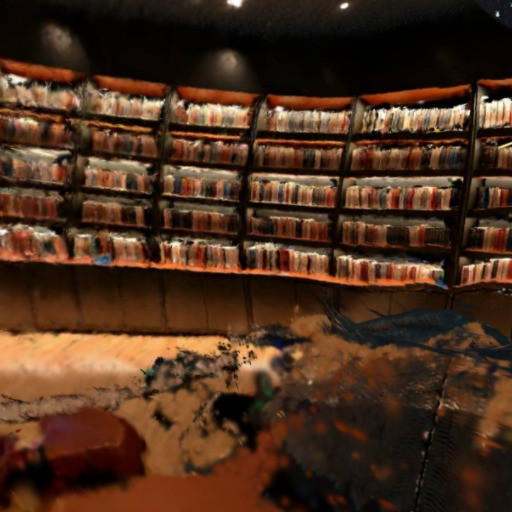} &
\includegraphics[height=0.15\textwidth]{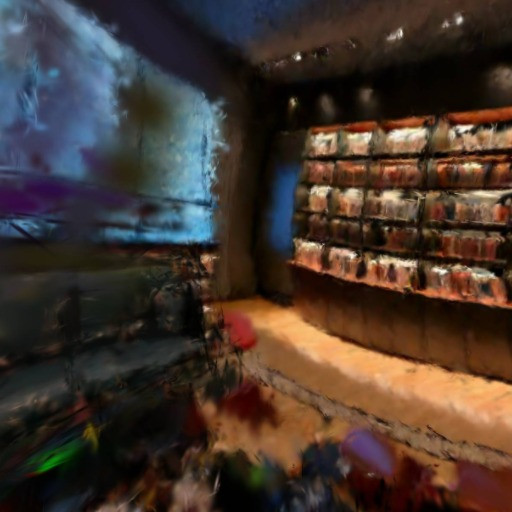} \\

\raisebox{0.15\height}{\rotatebox{90}{\parbox{2cm}{\centering Text2Room \\ \cite{hollein2023text2room}}}} &
\includegraphics[height=0.15\textwidth]{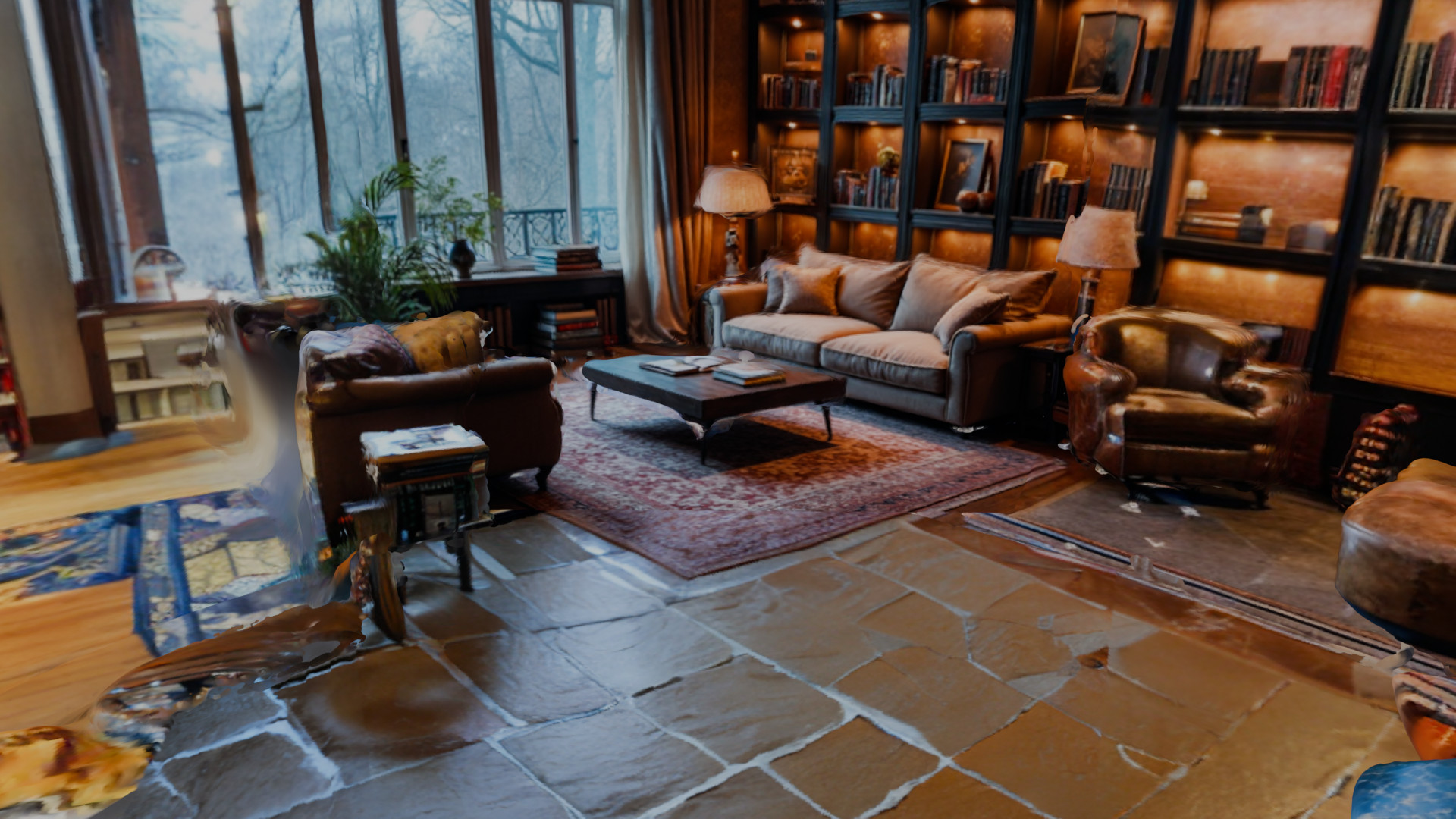} &
\includegraphics[height=0.15\textwidth]{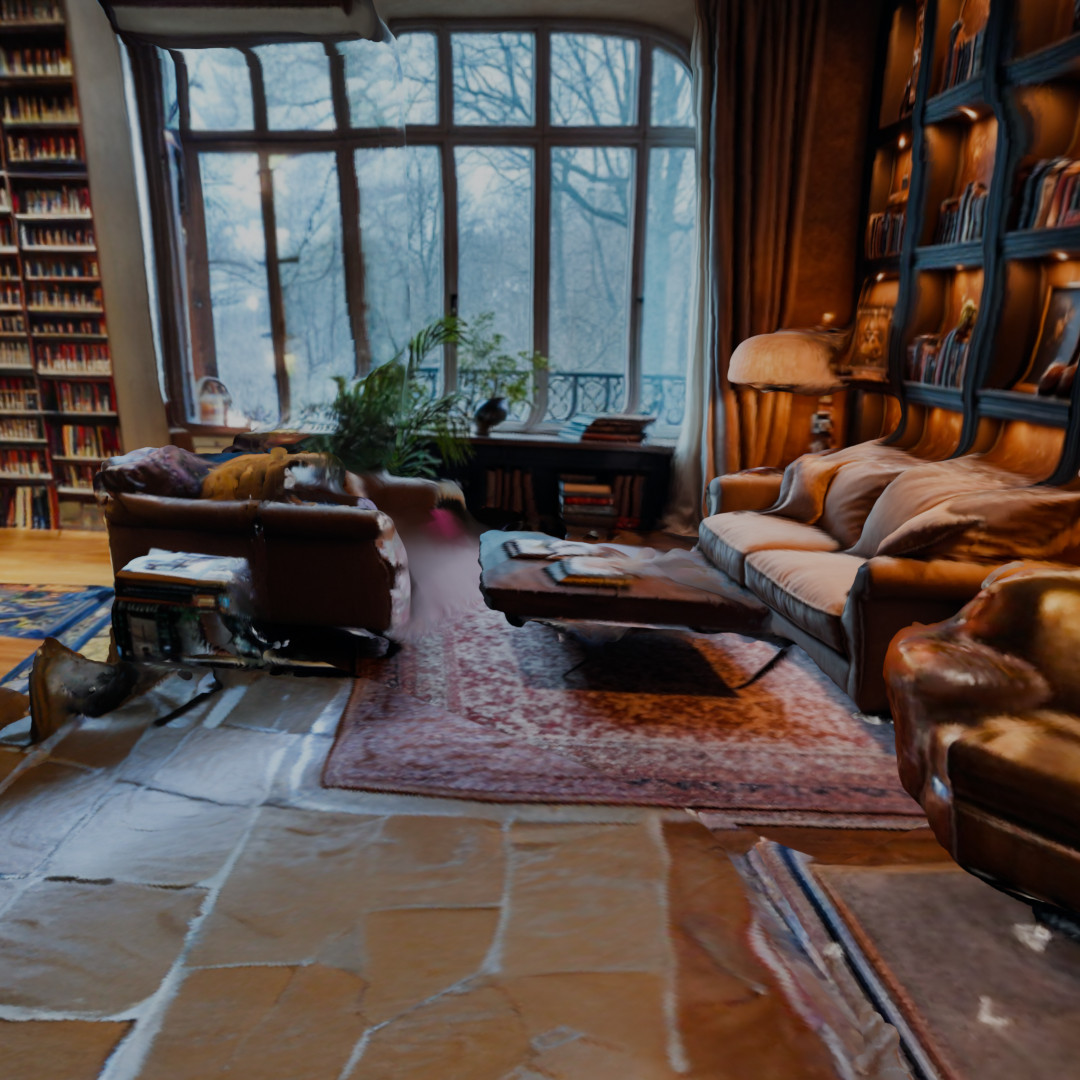} &
\includegraphics[height=0.15\textwidth]{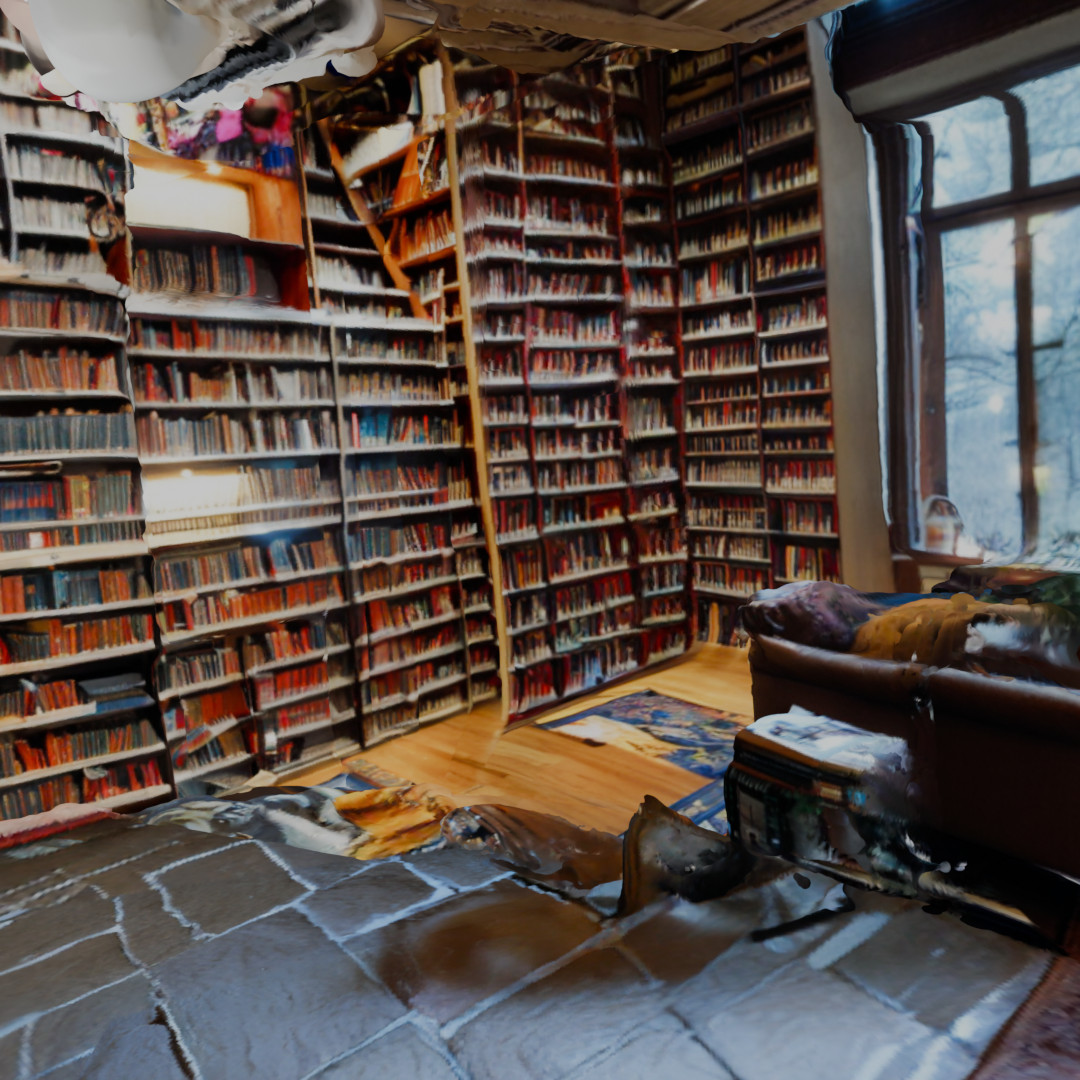} &
\includegraphics[height=0.15\textwidth]{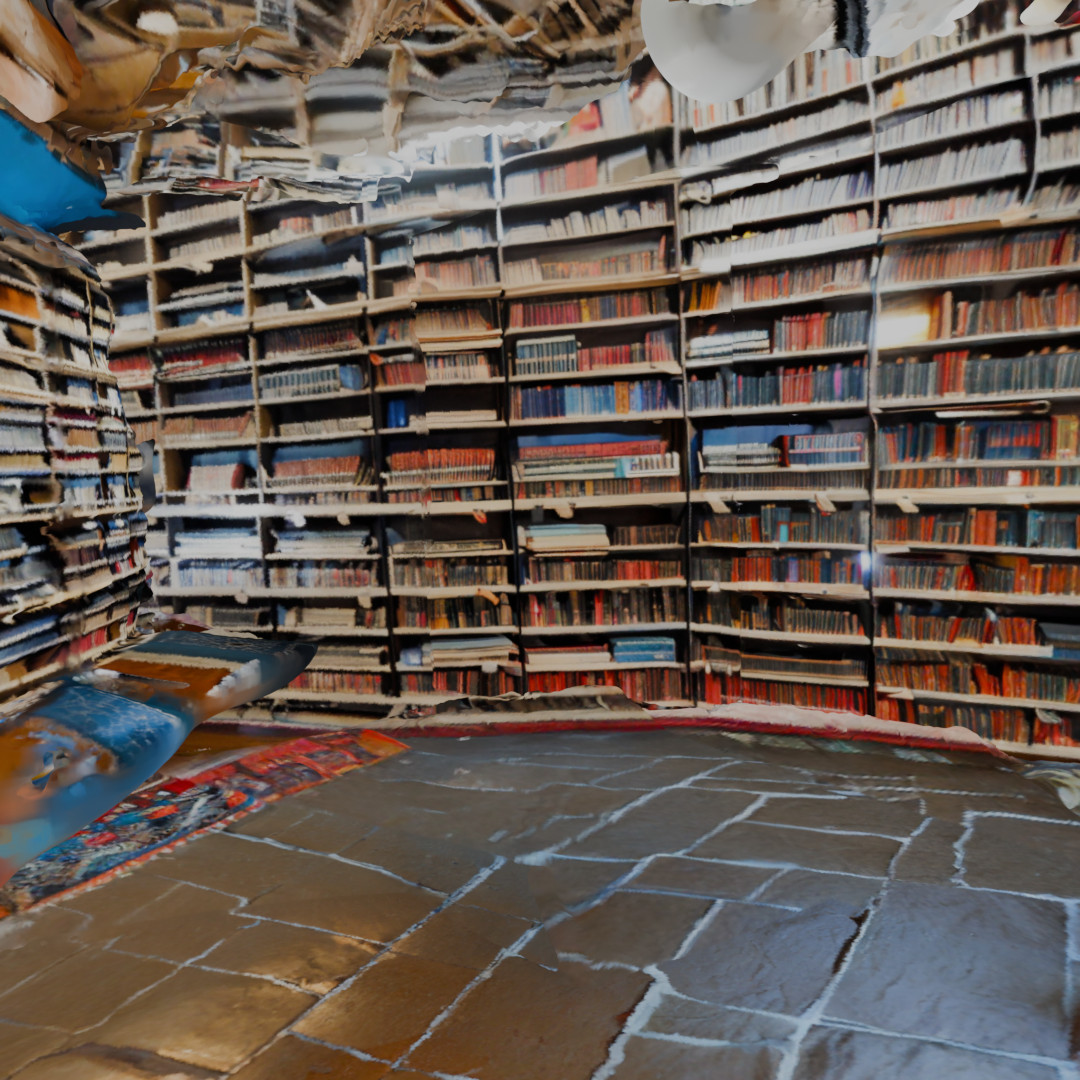} &
\includegraphics[height=0.15\textwidth]{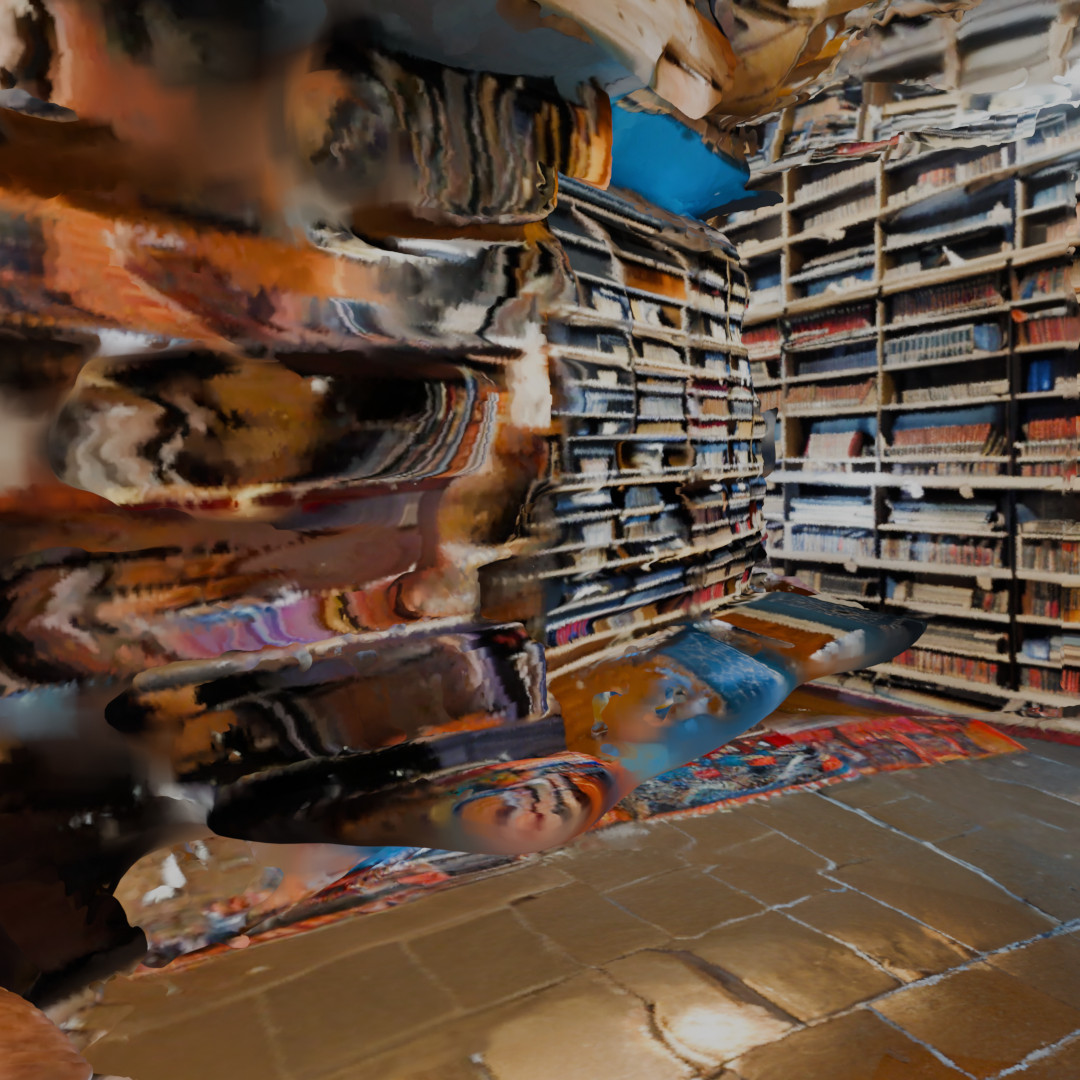} \\

\raisebox{0.15\height}{\rotatebox{90}{\parbox{2cm}{\centering WonderWorld \\ \cite{yu2024wonderworld}}}} &
\includegraphics[height=0.15\textwidth]{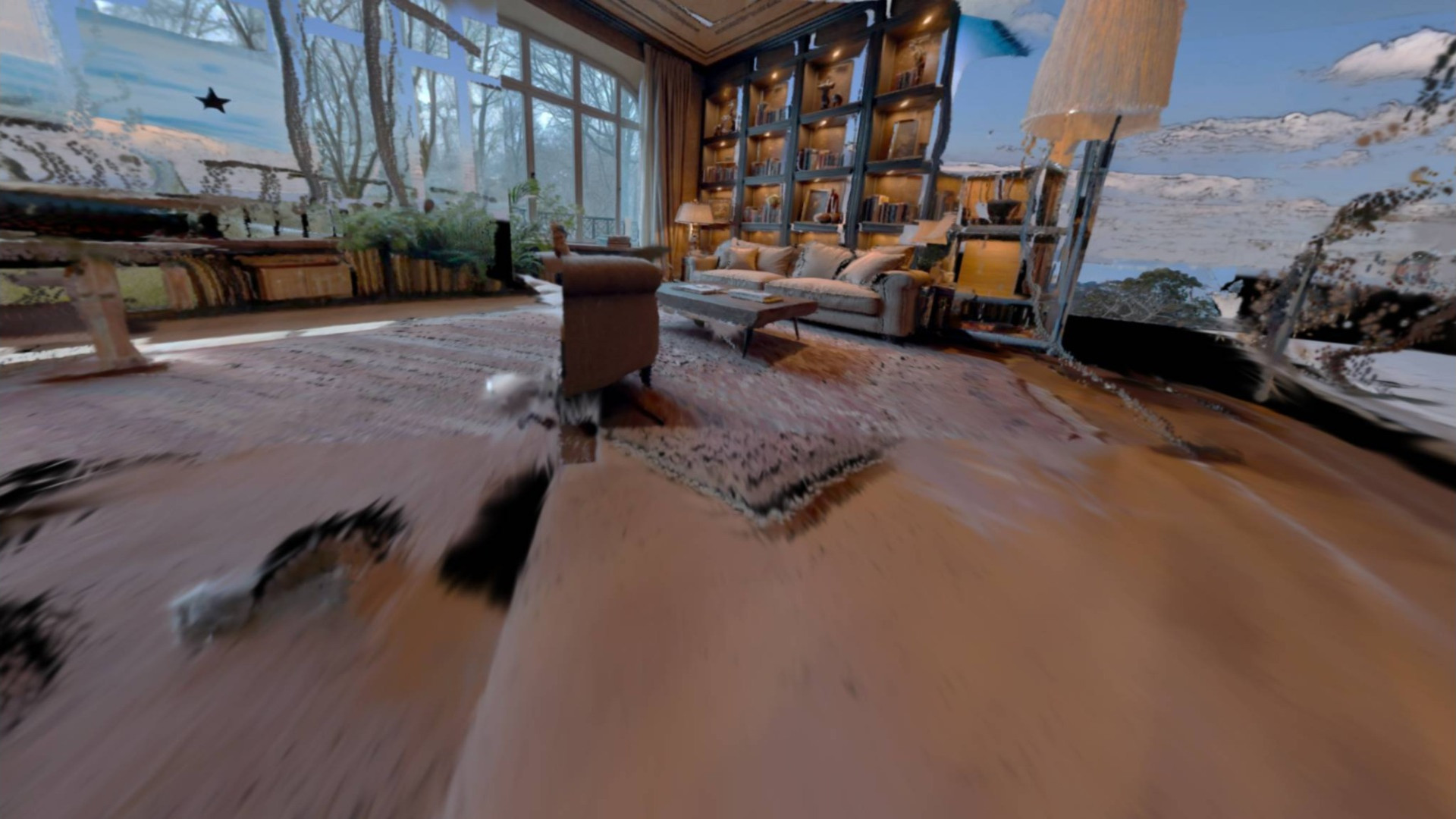} &
\includegraphics[height=0.15\textwidth]{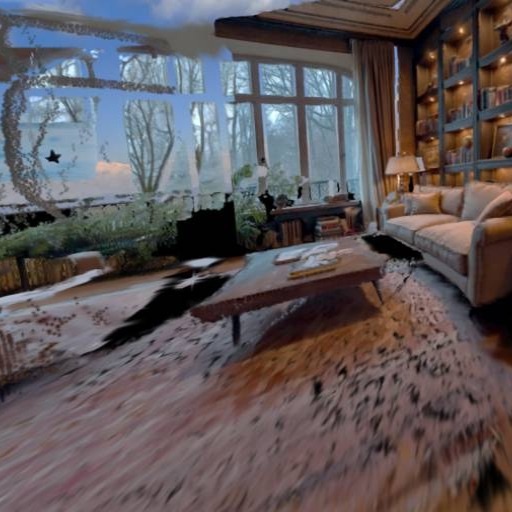} &
\includegraphics[height=0.15\textwidth]{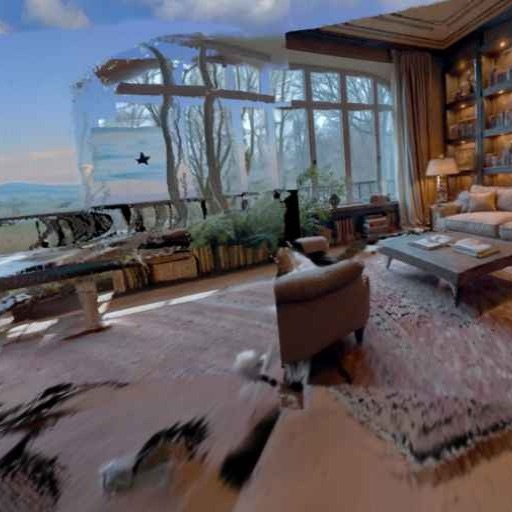} &
\includegraphics[height=0.15\textwidth]{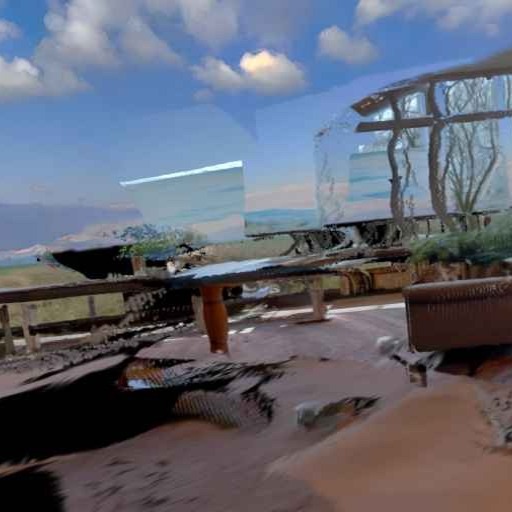} &
\includegraphics[height=0.15\textwidth]{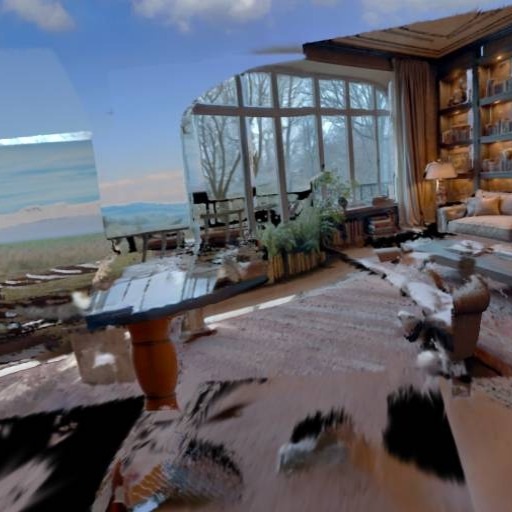} \\

\raisebox{0.15\height}{\rotatebox{90}{\parbox{2cm}{\centering FlexWorld \\ \cite{chen2025flexworld}}}} &
\includegraphics[height=0.15\textwidth]{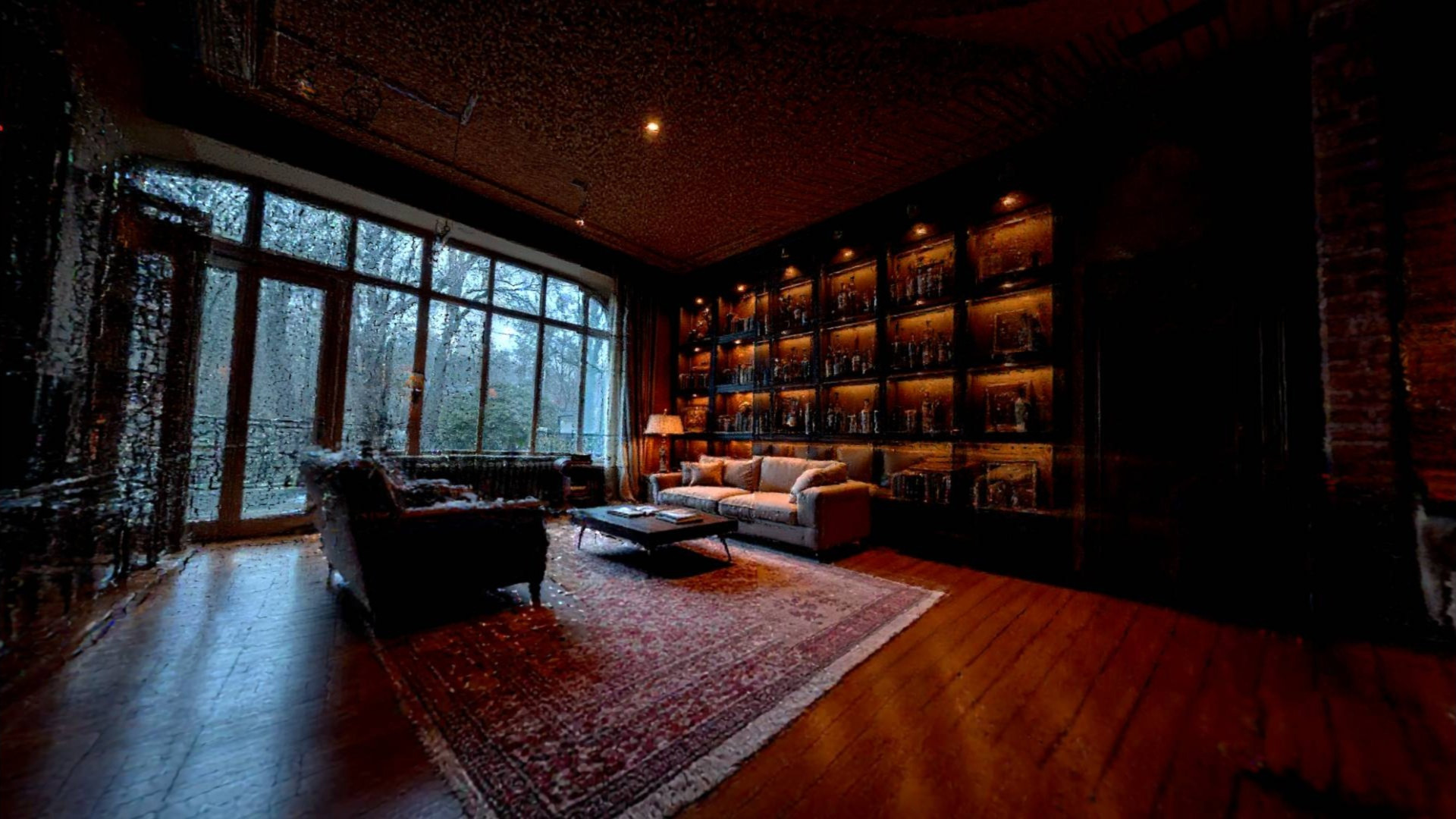} &
\includegraphics[height=0.15\textwidth]{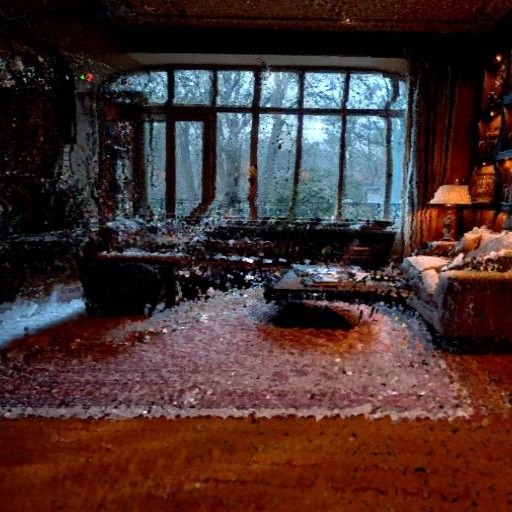} &
\includegraphics[height=0.15\textwidth]{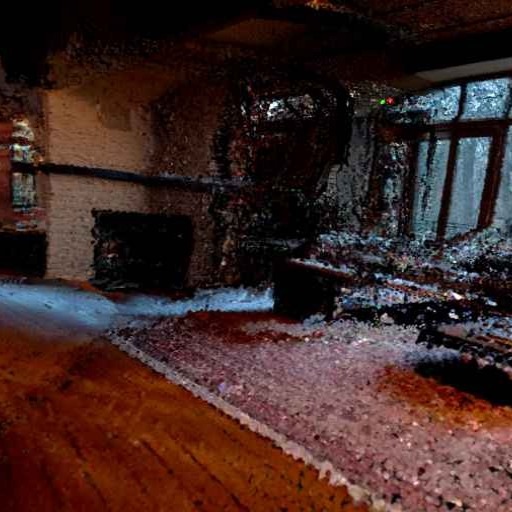} &
\includegraphics[height=0.15\textwidth]{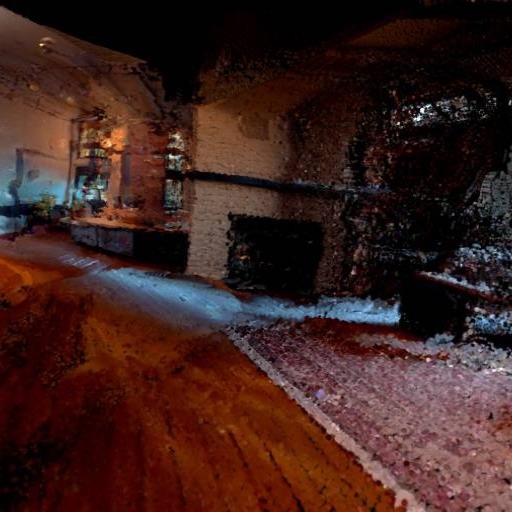} &
\includegraphics[height=0.15\textwidth]{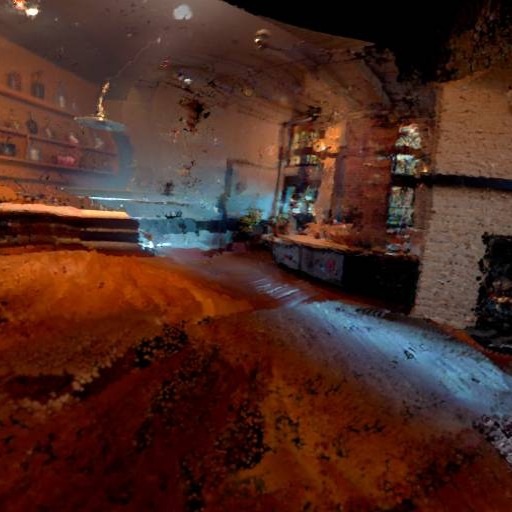} \\

\raisebox{0.15\height}{\rotatebox{90}{\parbox{2cm}{\centering SEVA \\ \cite{zhou2025stable}}}} &
\includegraphics[height=0.15\textwidth]{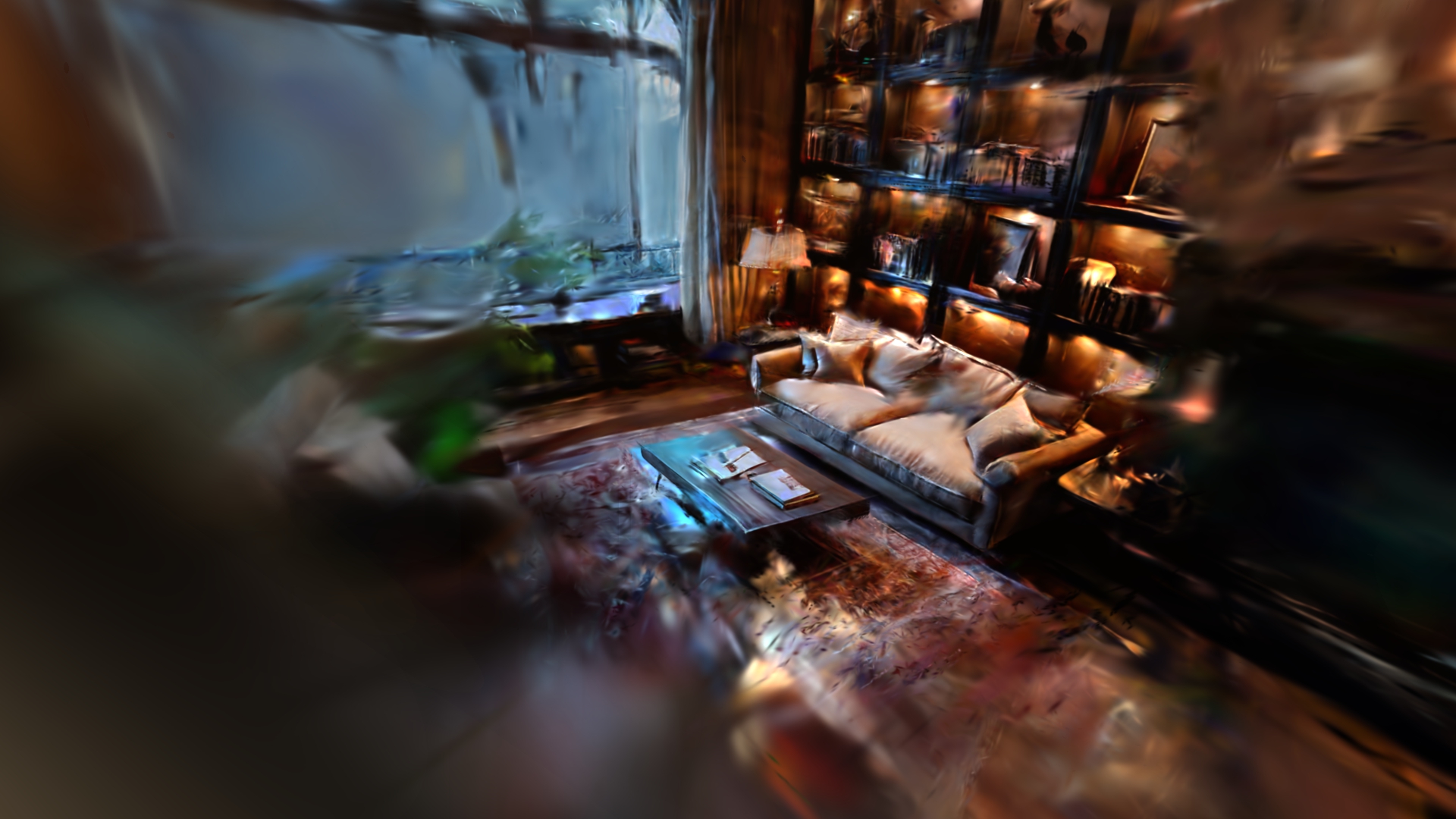} &
\includegraphics[height=0.15\textwidth]{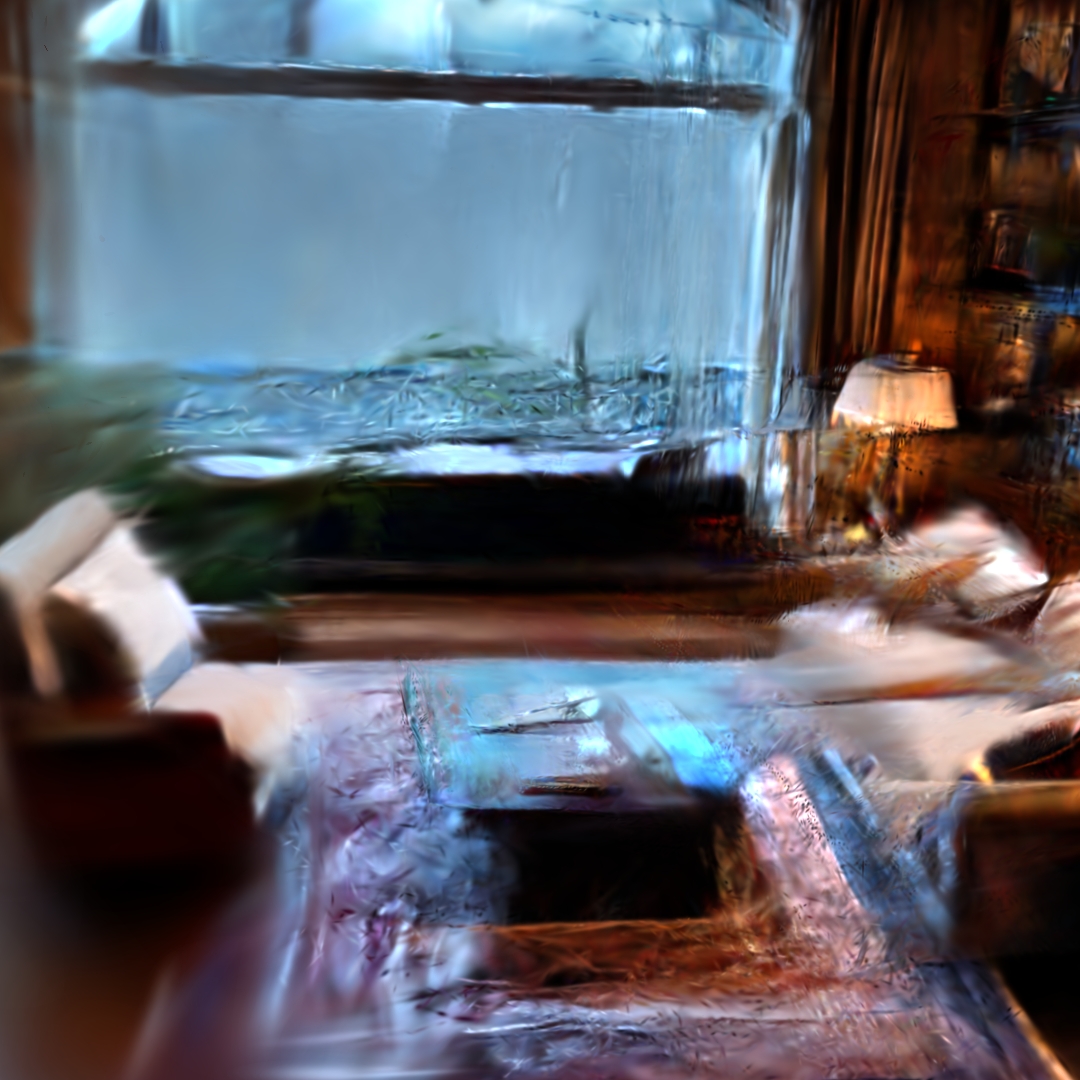} &
\includegraphics[height=0.15\textwidth]{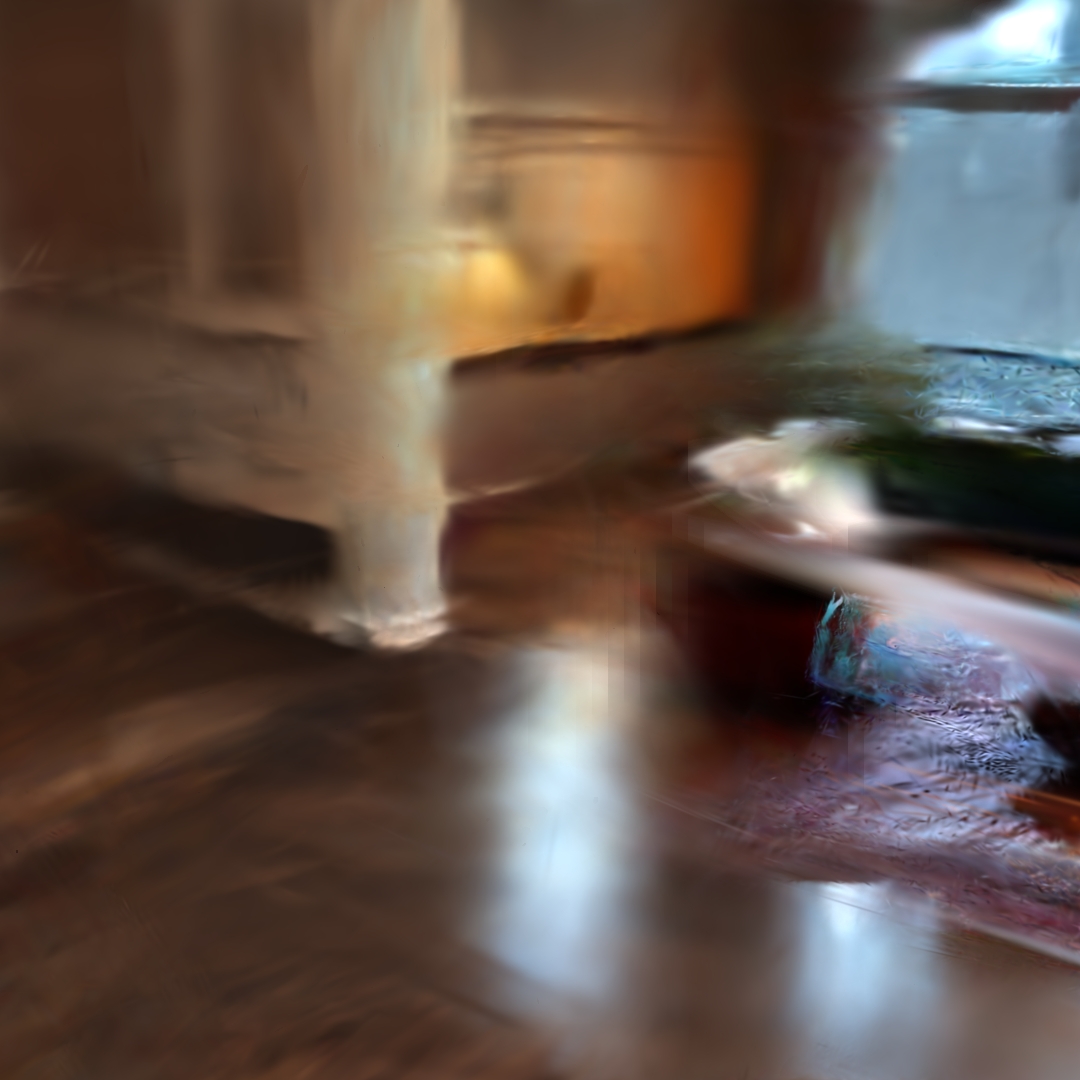} &
\includegraphics[height=0.15\textwidth]{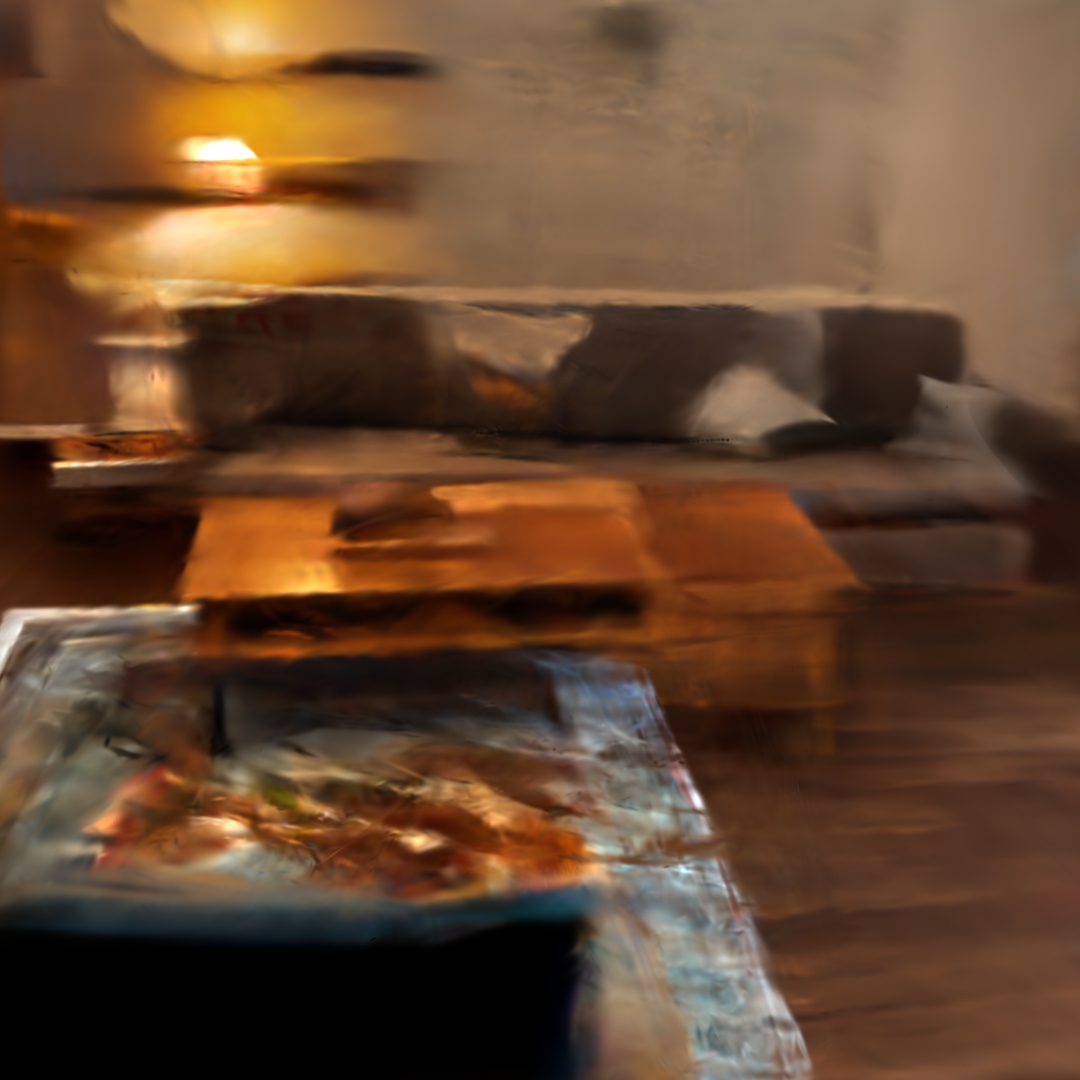} &
\includegraphics[height=0.15\textwidth]{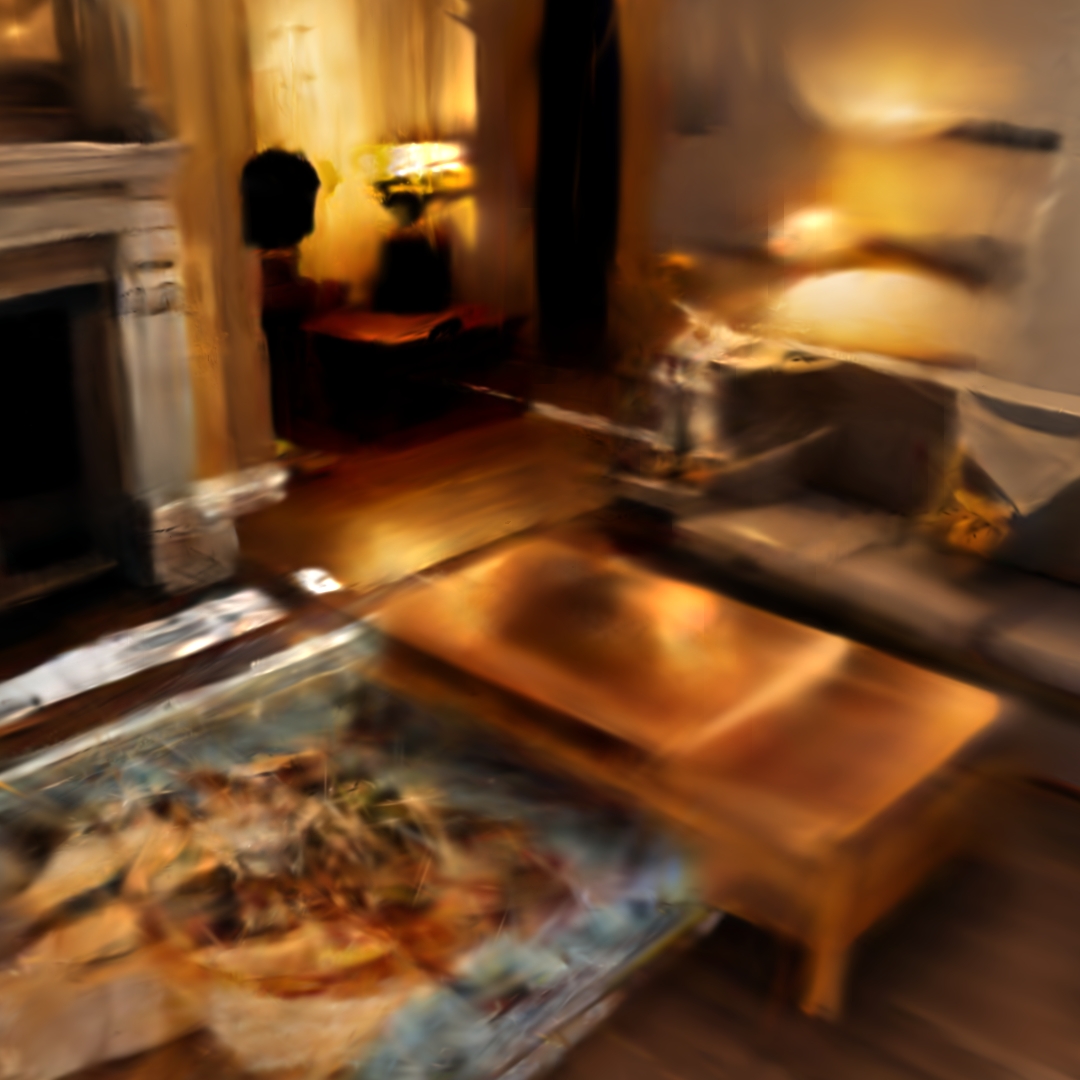} \\

\raisebox{1.5\height}{\rotatebox{90}{Ours}} &
\includegraphics[height=0.15\textwidth]{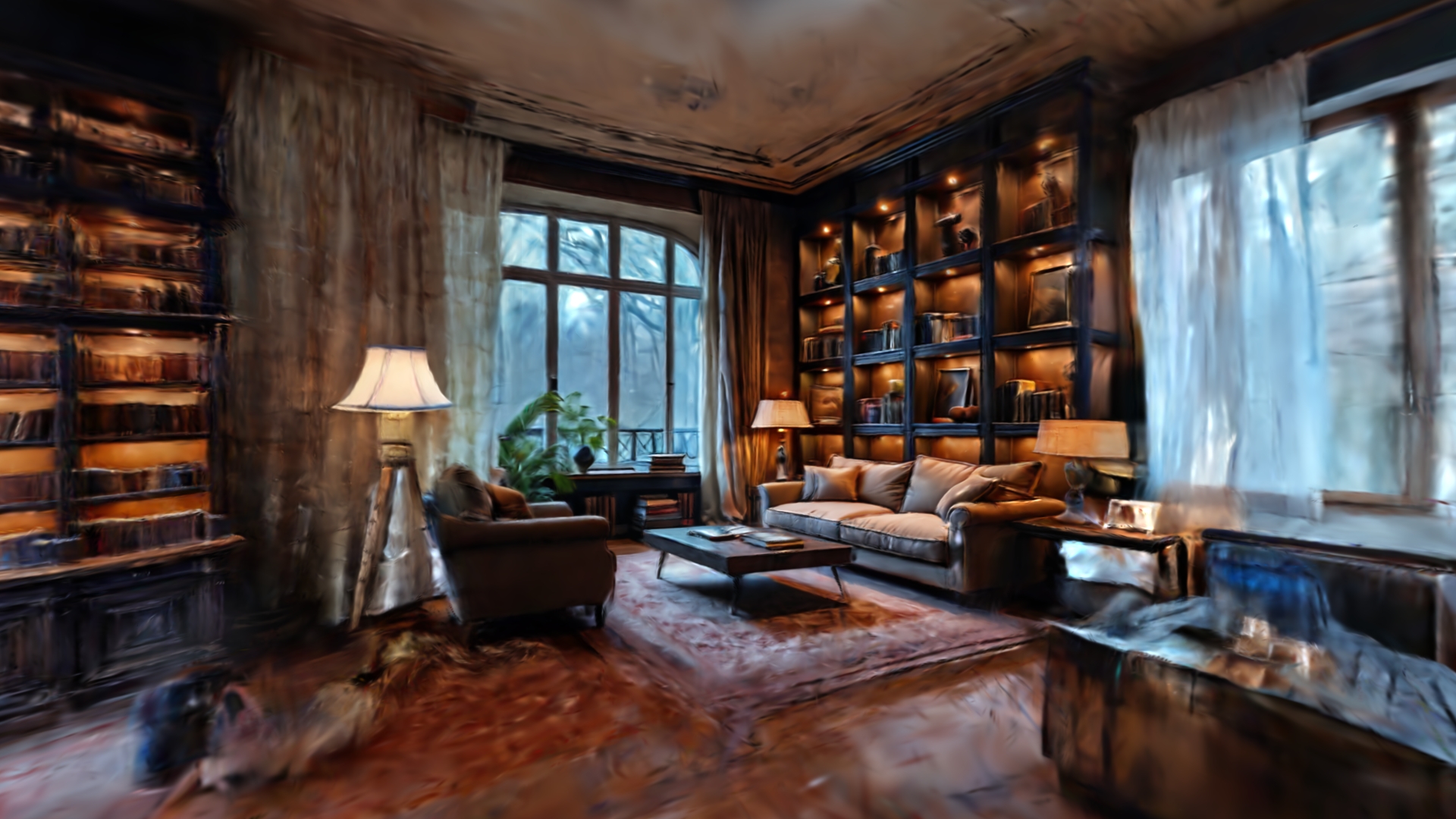} &
\includegraphics[height=0.15\textwidth]{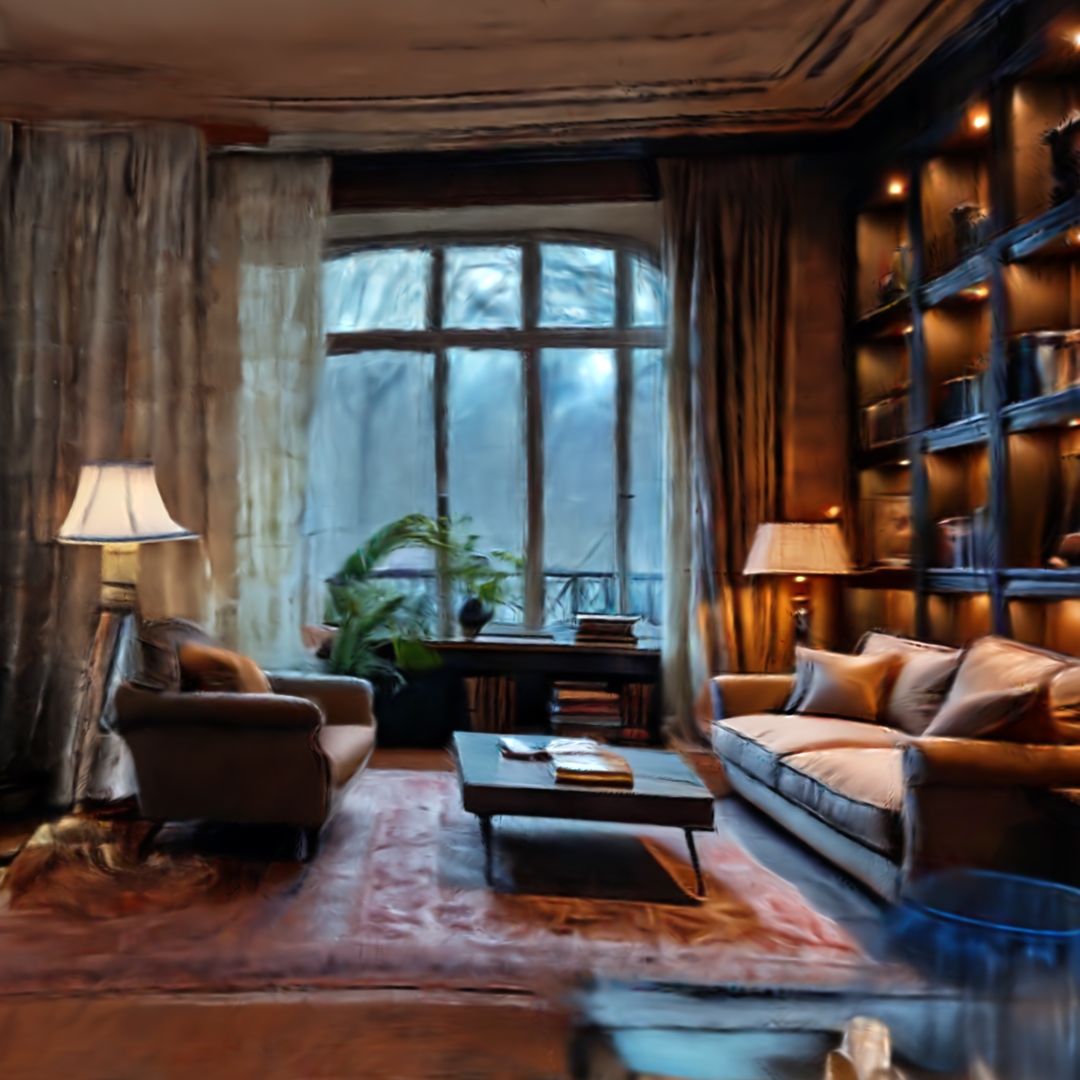} &
\includegraphics[height=0.15\textwidth]{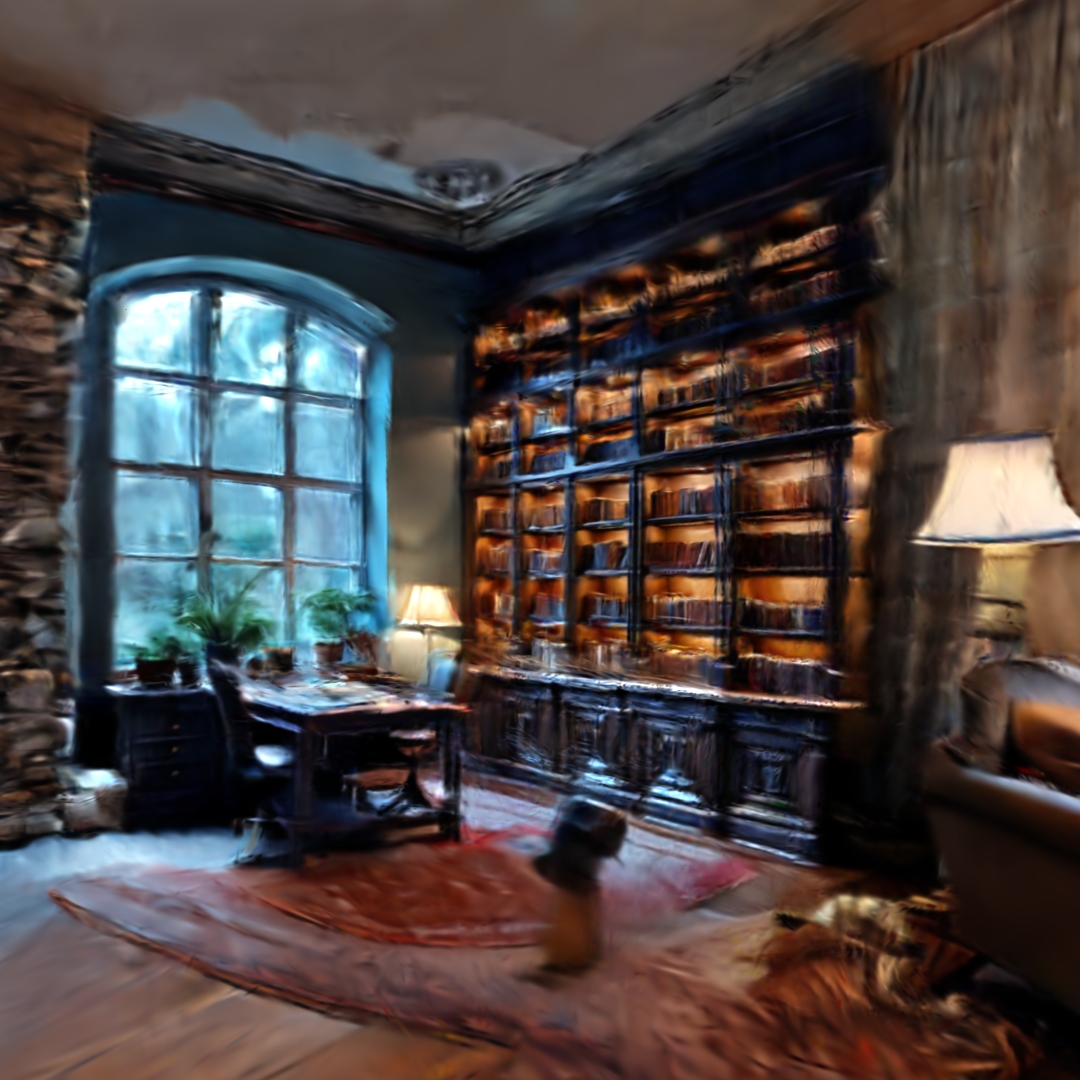} &
\includegraphics[height=0.15\textwidth]{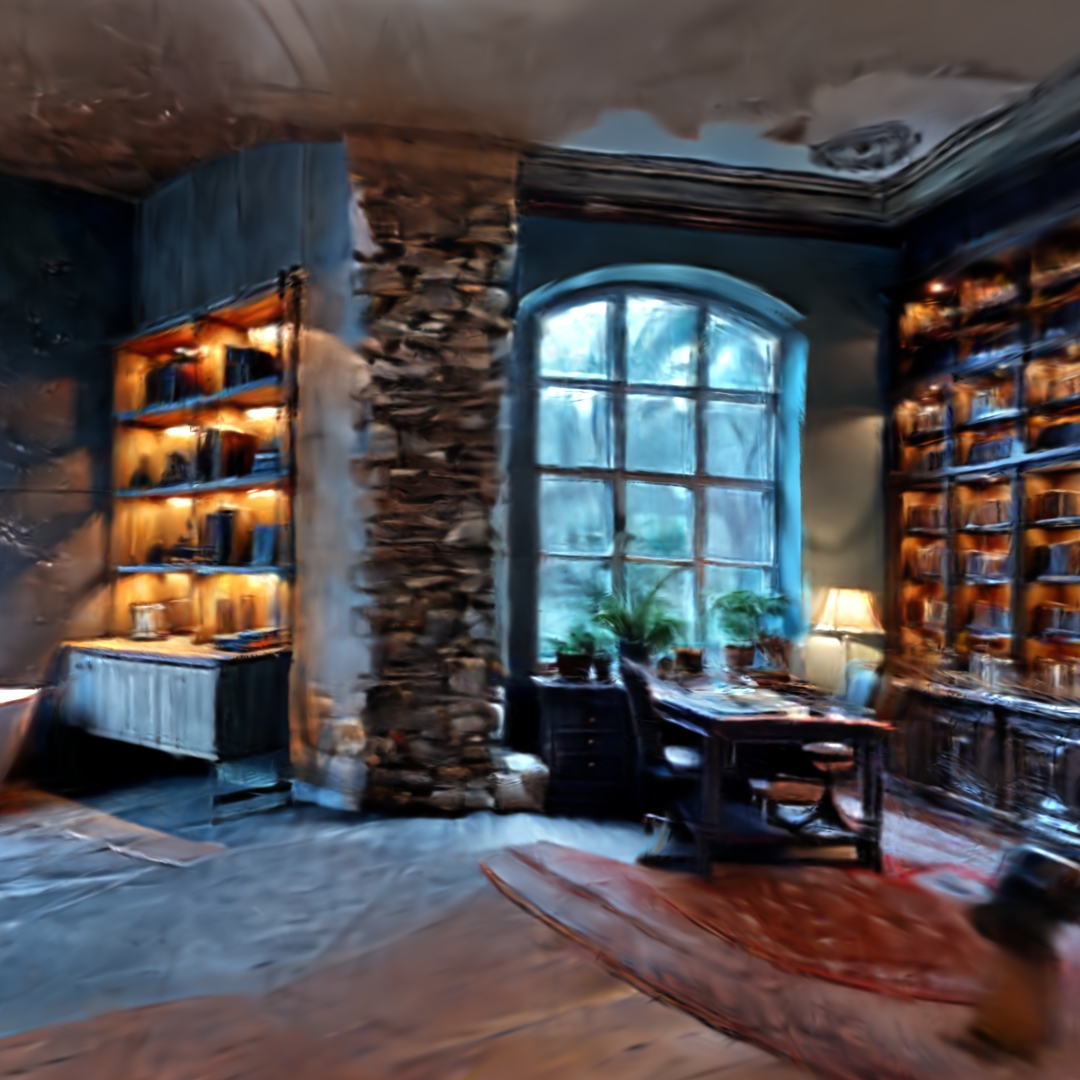} &
\includegraphics[height=0.15\textwidth]{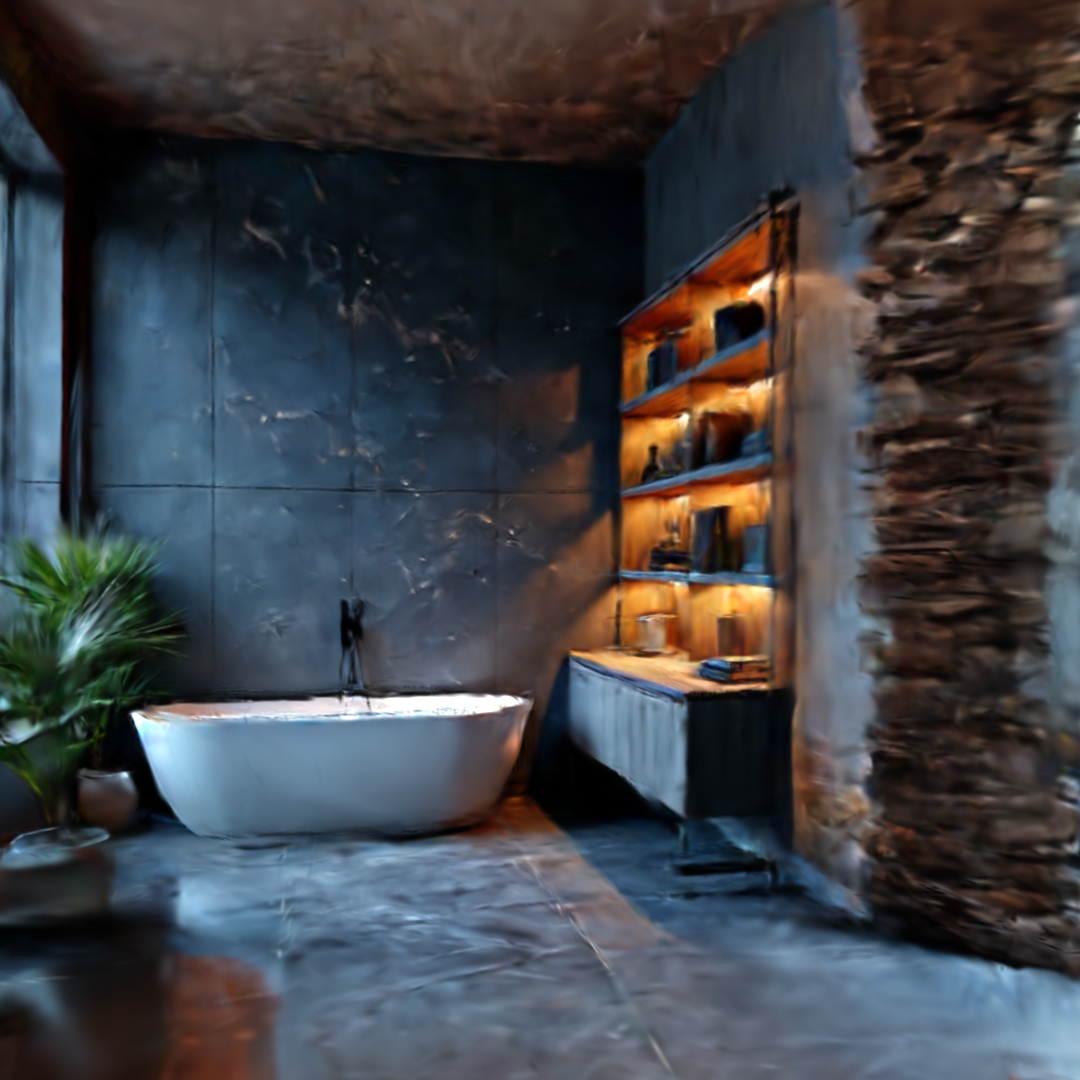} \\
\end{tabular}

\caption{
\textbf{Qualitative comparison of our method and baselines.}
We generate a scene given the text prompt ``star-written, constellation-binding, midnight celestial ink library''.
We visualize a scene overview from a centered camera position and 4 novel views that explore the scene from non-centered perspectives.
Our method retains compelling visual quality and generates plausible continuations of the scene, even under these extreme camera positions.
Please see the supplementary material for animated comparisons against baselines.
}
\Description{
\textbf{Qualitative comparison of our method and baselines.}
We generate a scene given the text prompt ``star-written, constellation-binding, midnight celestial ink library''.
We visualize a scene overview from a centered camera position and 4 novel views that explore the scene from non-centered perspectives.
Our method retains compelling visual quality and generates plausible continuations of the scene, even under these extreme camera positions.
Please see the supplementary material for animated comparisons against baselines.
}
\label{fig:ours-baselines}
\end{figure*}

%% file: tables/tab_ours_baselines.tex
\begin{table}
  \centering
    \caption{
  \textbf{Quantitative comparison.}
  We report 2D metrics and user study results, including: Clip Score~(\emph{CS}), Inception Score~(\emph{IS}), Perceptual Quality (\emph{PQ}), and 3D-Consistency ~(\emph{3DC}).
  We evaluate all methods on novel views rendered from perspectives beyond the scene center, i.e., we quantify their ability to explore the scene arbitrarily.
  Our method creates scenes with the highest text-prompt similarity and is preferred in our user study.
  }
  \begin{tabular}{l cc cc}
    \toprule
        \multirow{2}{*}{Method} & \multicolumn{2}{c}{2D Metrics} & \multicolumn{2}{c}{User Study}\\
                        \cmidrule(l{2pt}r{2pt}){2-3} \cmidrule(l{2pt}r{2pt}){4-5}
    & CS $\uparrow$ & IS $\uparrow$ & PQ $\uparrow$ & 3DC $\uparrow$\\
    \midrule
    DreamScene360~\cite{zhou2024dreamscene360} & 23.51 & 2.06 & 3.01 & 3.04 \\
    LayerPano3D~\cite{yang2024layerpano3d} & 24.30 & 2.04 & 2.16 & 2.14 \\
    \midrule
    Text2Room~\cite{hollein2023text2room} & 24.37 & 1.92 & 2.68 & 2.32 \\
    WonderWorld~\cite{yu2024wonderworld} & 19.74 & 2.23 & 2.34 & 1.75 \\
    \midrule
    FlexWorld~\cite{chen2025flexworld} & 21.61 & \textbf{2.31} & 2.89 & 2.71 \\
    SEVA~\cite{zhou2025stable} & 21.63 & 2.13 & 2.51 & 1.91 \\
    \midrule
    \midrule
    Ours (1 image) & 21.42 & 1.88 & 2.42 & 2.03 \\
    Ours (4 images) & 25.65 & 2.29 & 2.48 & 2.14 \\
    Ours (w/o scene-mem) & 25.91 & 2.21 & 2.61 & 2.36 \\
    Ours (w/o coll-det) & 25.37 & 1.89 & 3.77 & 3.42 \\
    Ours & \textbf{25.94} & 2.27 & \textbf{4.04} & \textbf{4.02} \\
    \bottomrule
  \end{tabular}
  \label{tab:ours-baseline}
\end{table}

%% file: tables/fig_ours_ablation.tex
\begin{figure*}
\centering
\setlength\tabcolsep{1pt}         
\renewcommand{\arraystretch}{0.5}

\begin{tabular}{cccccc}

& Generated scene overview & \multicolumn{3}{c}{Rendered novel views} \\

\raisebox{0.15\height}{\rotatebox{90}{\parbox{2cm}{\centering Ours\\1 image}}} &
\includegraphics[height=0.15\textwidth]{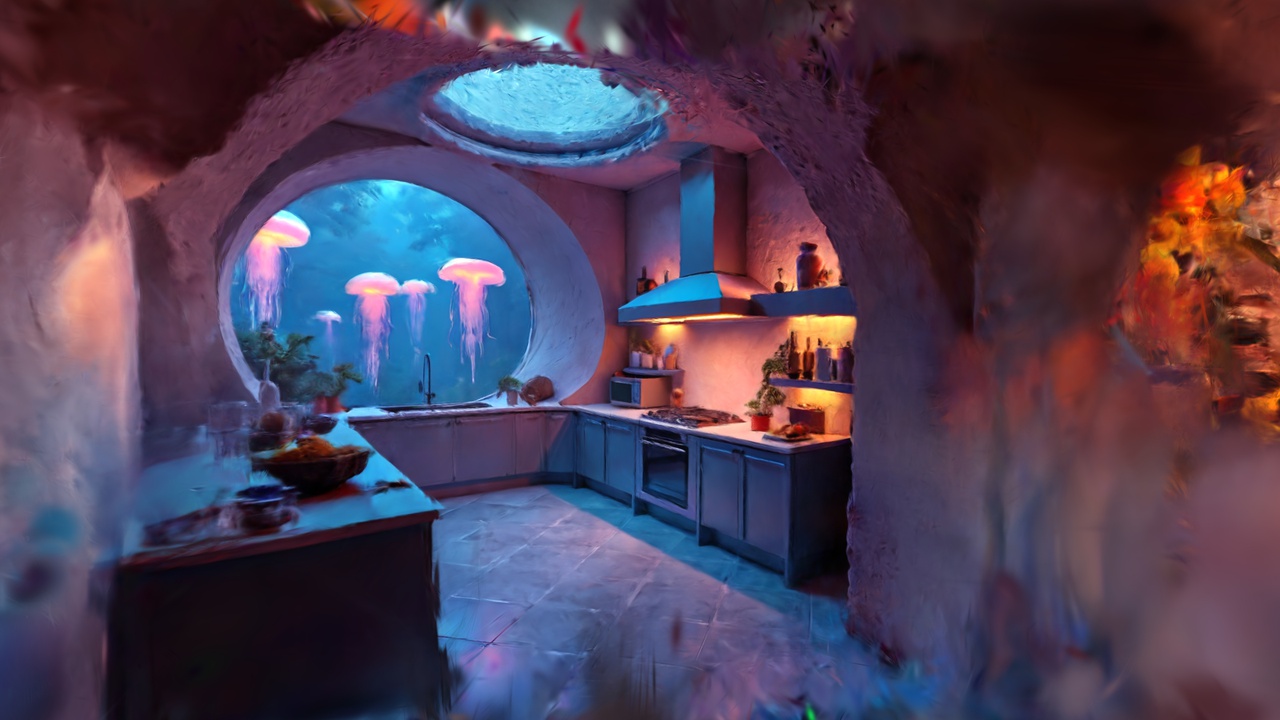} &
\includegraphics[height=0.15\textwidth]{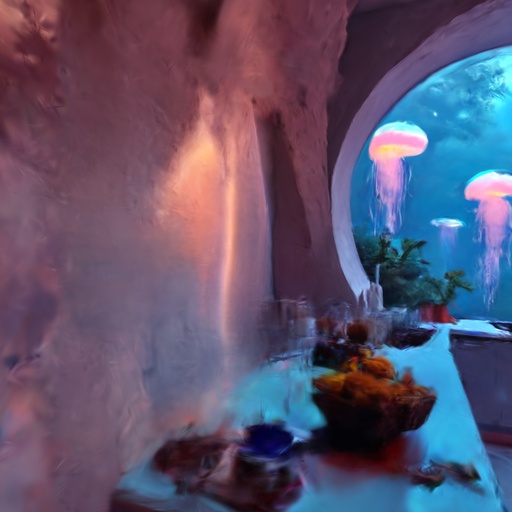} &
\includegraphics[height=0.15\textwidth]{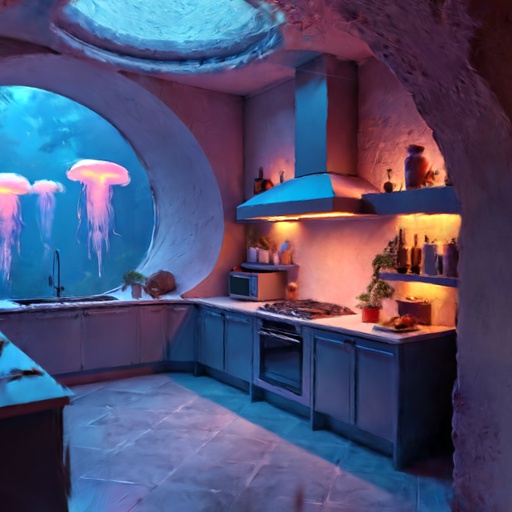} &
\includegraphics[height=0.15\textwidth]{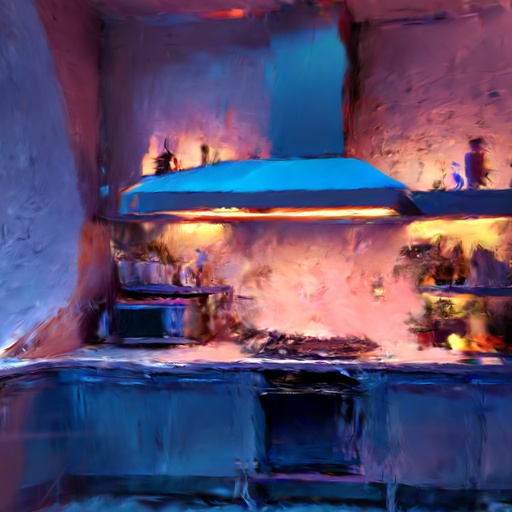} &
\includegraphics[height=0.15\textwidth]{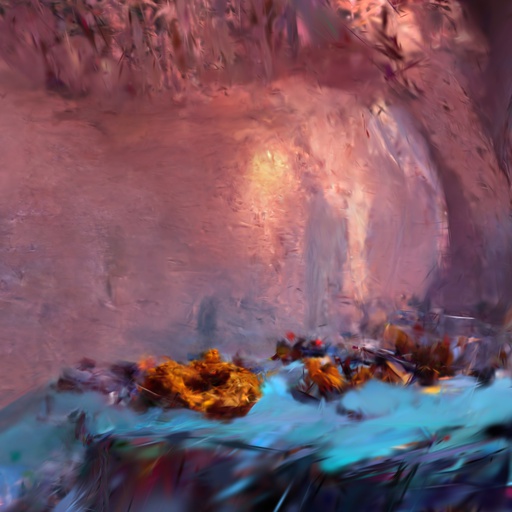} \\

\raisebox{0.15\height}{\rotatebox{90}{\parbox{2cm}{\centering Ours\\4 images}}} &
\includegraphics[height=0.15\textwidth]{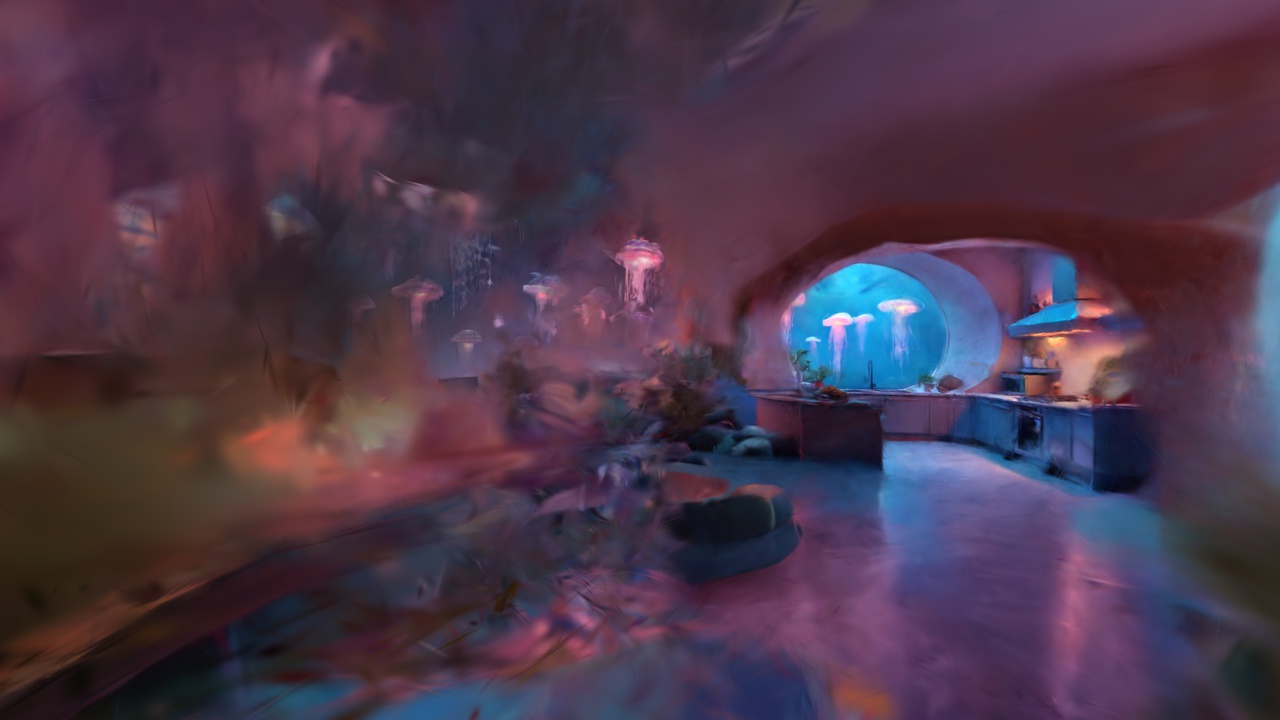} &
\includegraphics[height=0.15\textwidth]{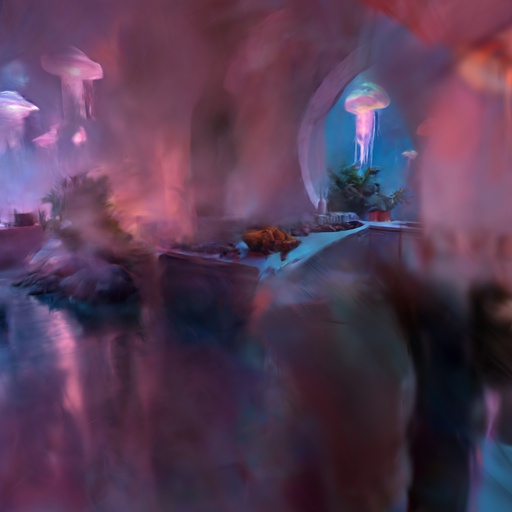} &
\includegraphics[height=0.15\textwidth]{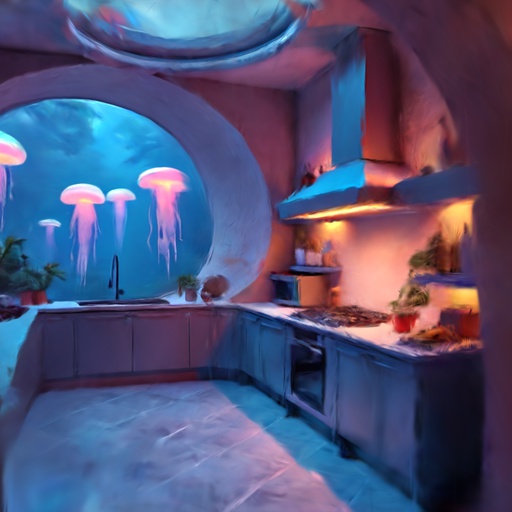} &
\includegraphics[height=0.15\textwidth]{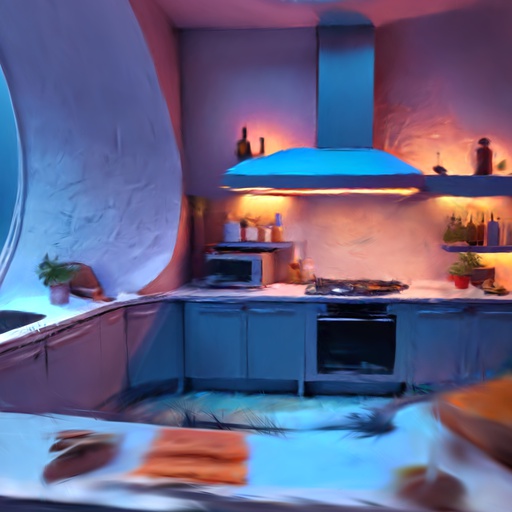} &
\includegraphics[height=0.15\textwidth]{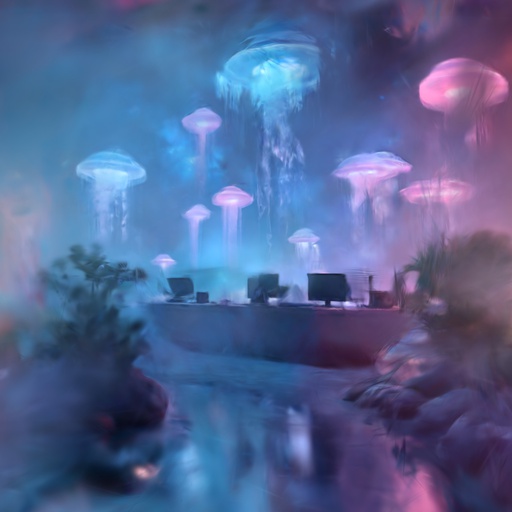} \\

\raisebox{0.15\height}{\rotatebox{90}{\parbox{2cm}{\centering Ours\\w/o scene-mem}}} &
\includegraphics[height=0.15\textwidth]{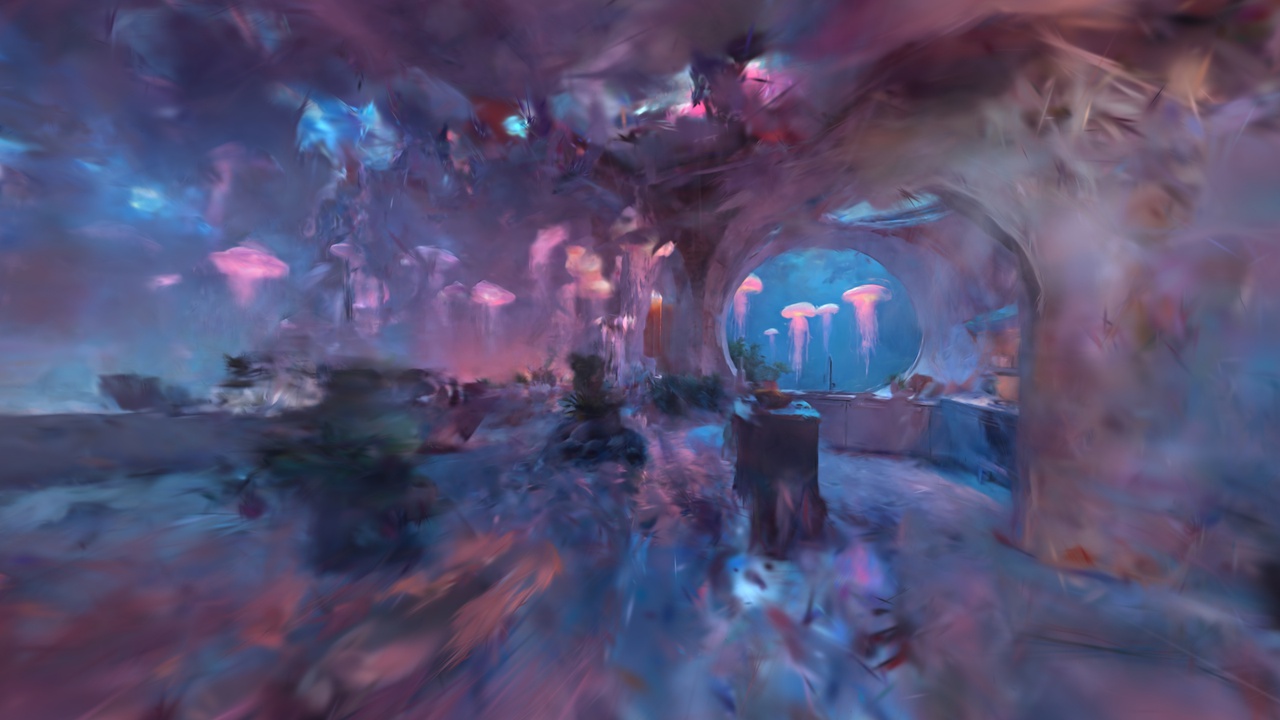} &
\includegraphics[height=0.15\textwidth]{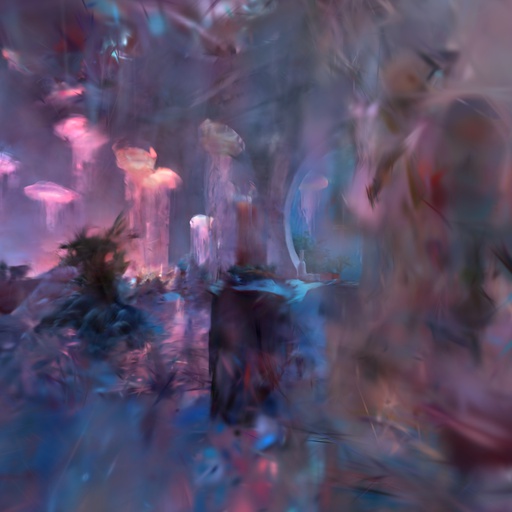} &
\includegraphics[height=0.15\textwidth]{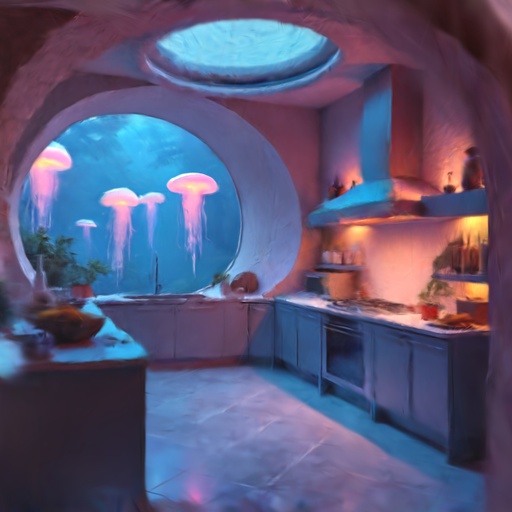} &
\includegraphics[height=0.15\textwidth]{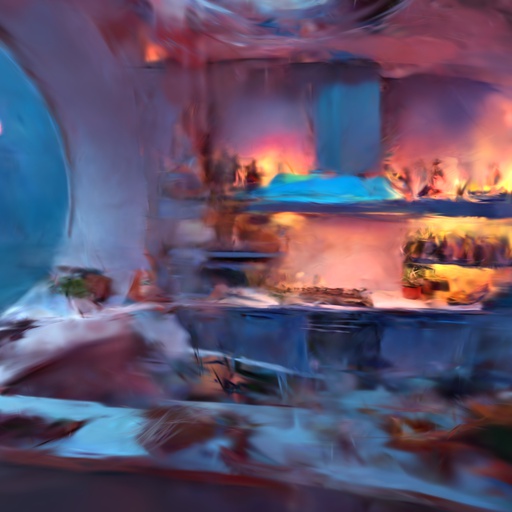} &
\includegraphics[height=0.15\textwidth]{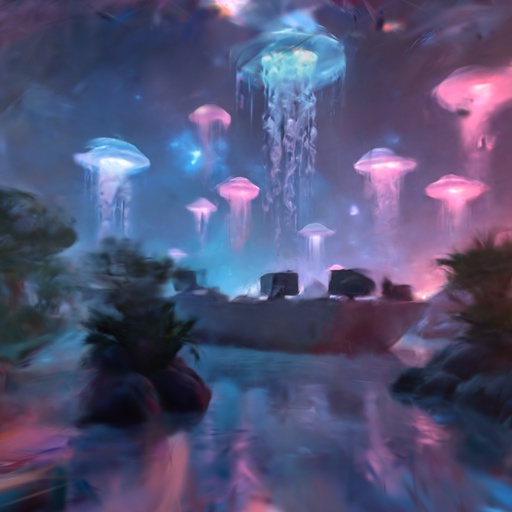} \\

\raisebox{0.2\height}{\rotatebox{90}{\parbox{2cm}{\centering Ours\\w/o coll-det}}} &
\includegraphics[height=0.15\textwidth]{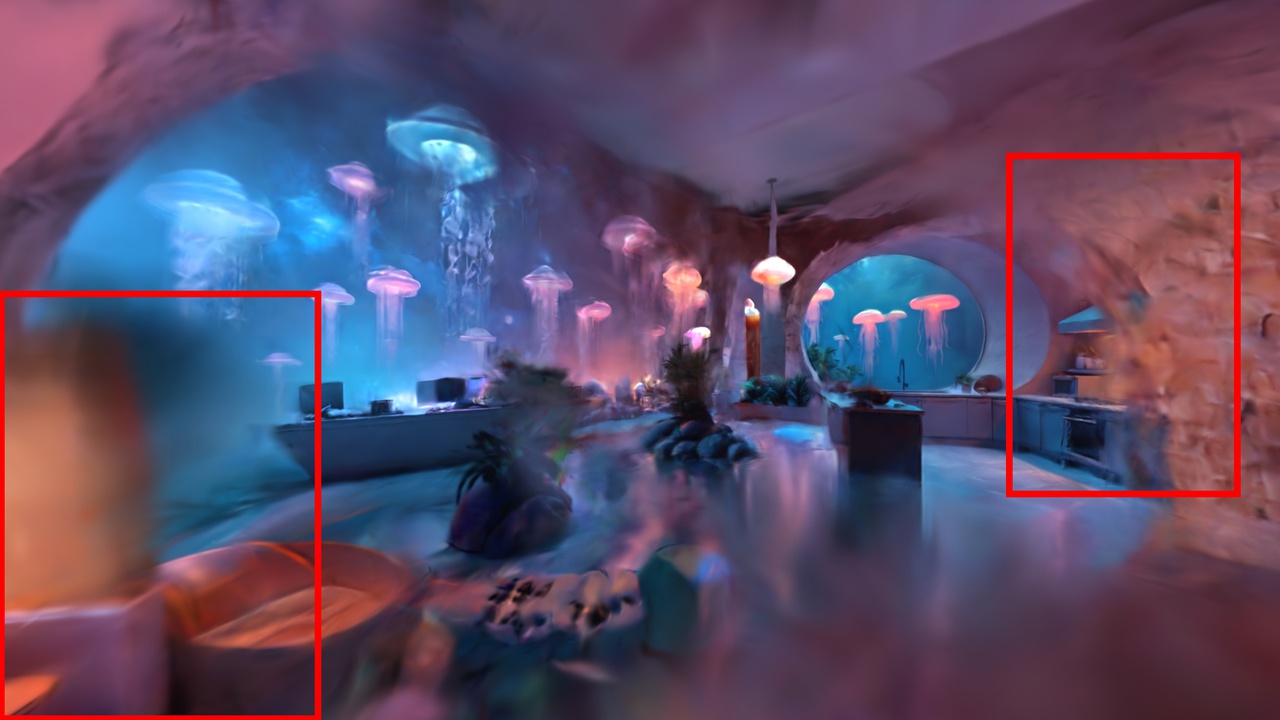} &
\includegraphics[height=0.15\textwidth]{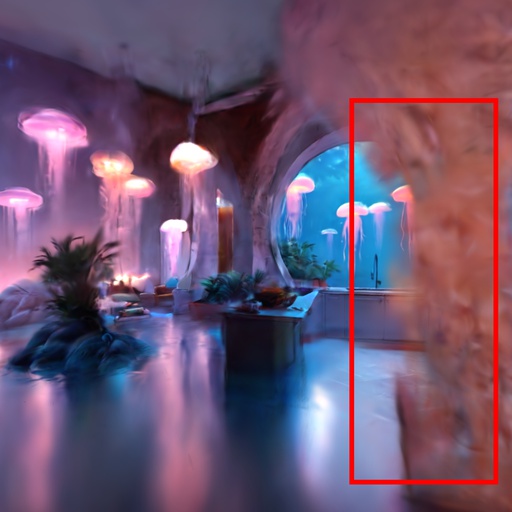} &
\includegraphics[height=0.15\textwidth]{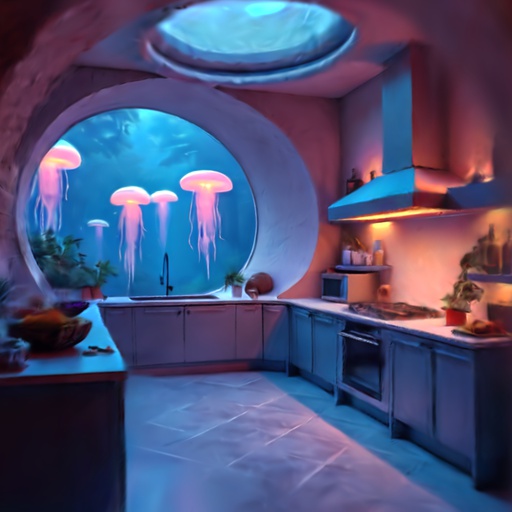} &
\includegraphics[height=0.15\textwidth]{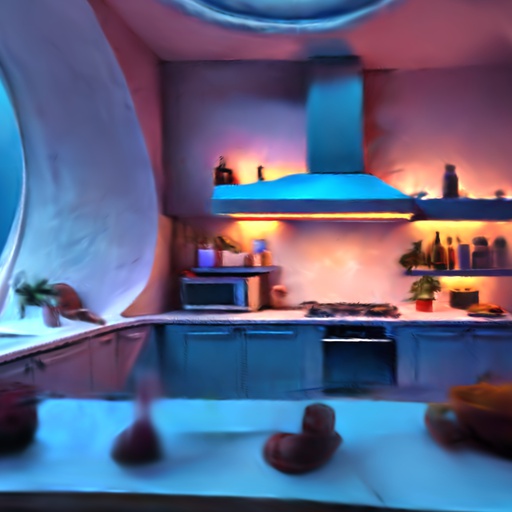} &
\includegraphics[height=0.15\textwidth]{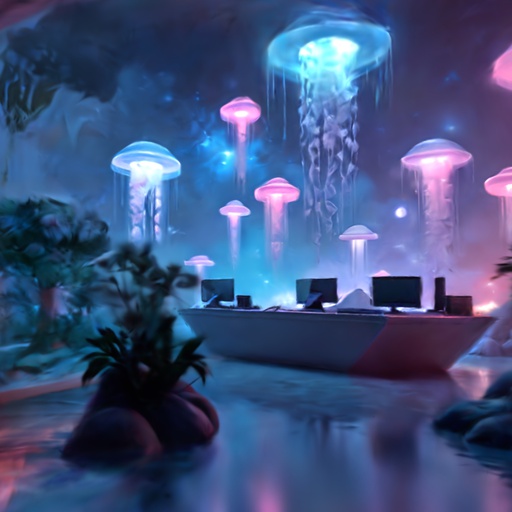} \\

\raisebox{1.5\height}{\rotatebox{90}{Ours}} &
\includegraphics[height=0.15\textwidth]{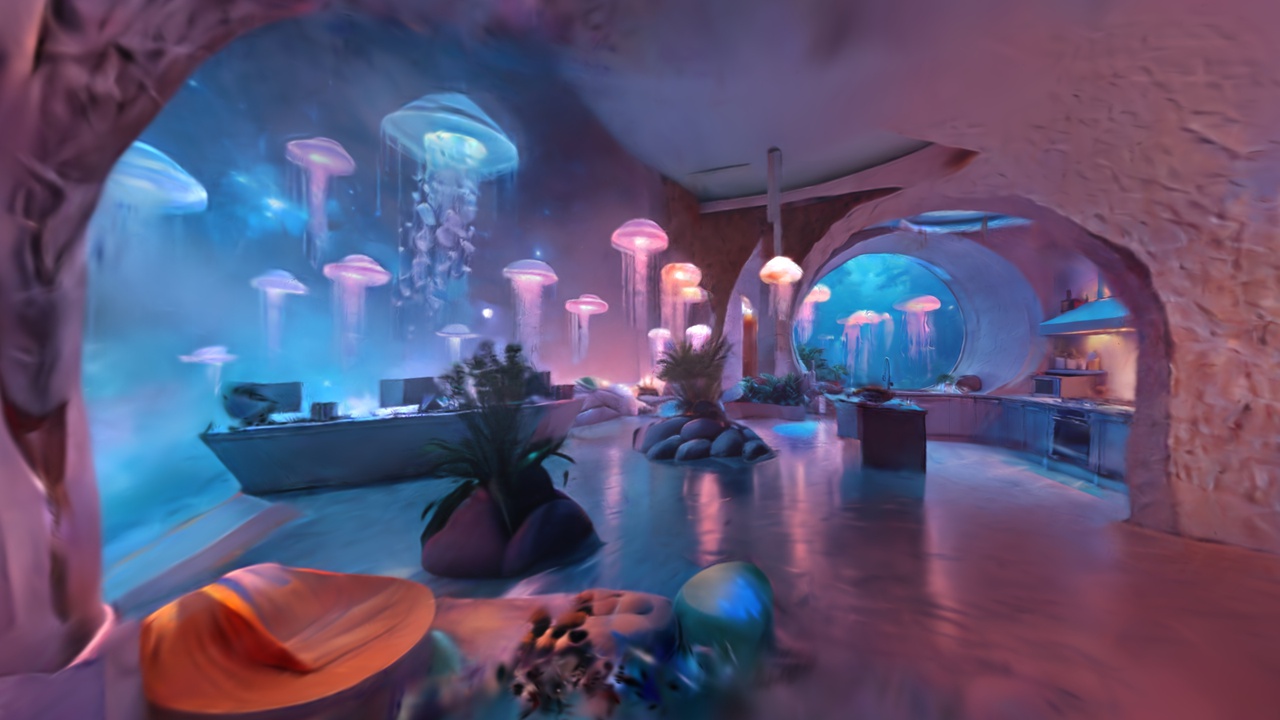} &
\includegraphics[height=0.15\textwidth]{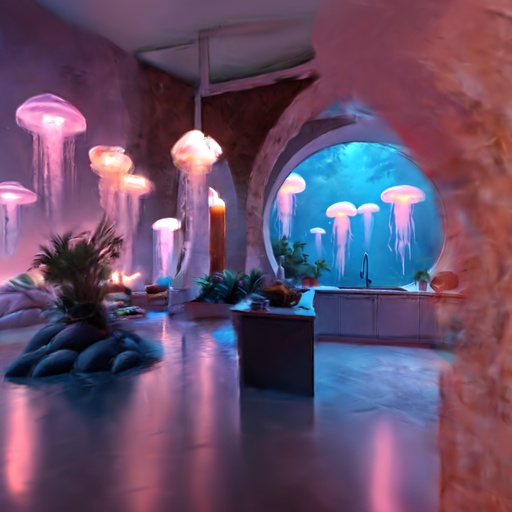} &
\includegraphics[height=0.15\textwidth]{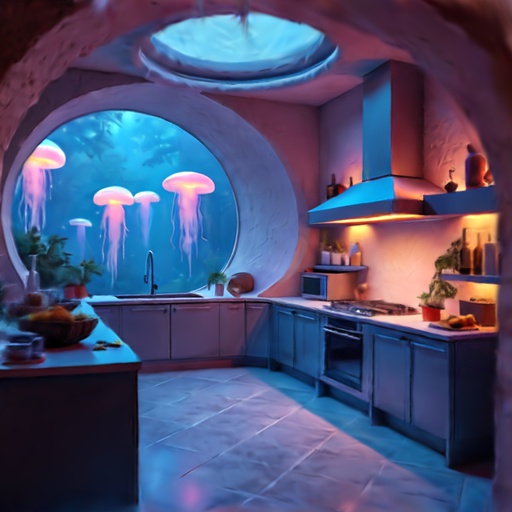} &
\includegraphics[height=0.15\textwidth]{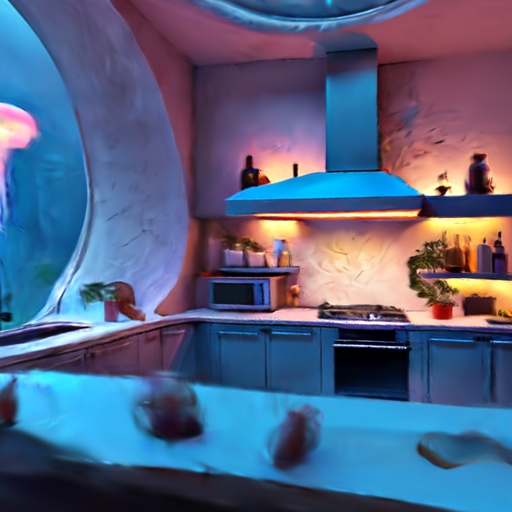} &
\includegraphics[height=0.15\textwidth]{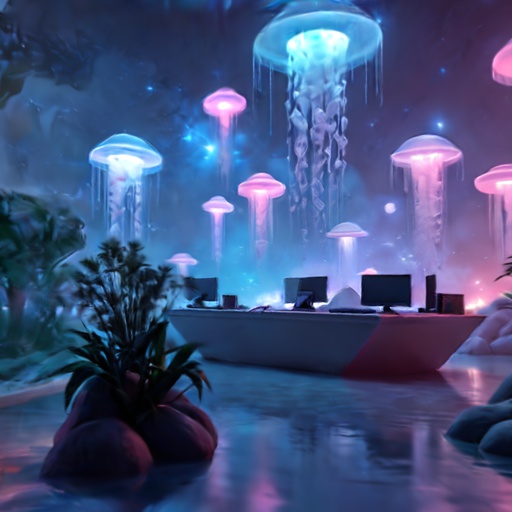} \\
\end{tabular}

\caption{
\textbf{Ablation study on the key components of our method.}
When conditioning video generation on only 1 initial image, generated scenes become smaller and less diverse: it is harder to generate novel areas from scratch than to inpaint around existing panoramic observations.
While 4 initial panoramic images enable the creation of larger scenes, the frames of trajectories without a starting image are less 3D-consistent.
This results in floating artifacts during rendering.
Our proposed scene memory mechanism is necessary, as each individual trajectory only covers a local area of the scene.
Thus, conditioning only within that trajectory similarly creates 3D-inconsistent scenes.
Our collision detection ensures, that the pre-defined trajectories do not intersect with generated scene content like the archway towards the kitchen area.
Without it, these areas become more fuzzy due to degenerate videos running into these structures.
}
\Description{
\textbf{Ablation study on the key components of our method.}
When conditioning video generation on only 1 initial image, generated scenes become smaller and less diverse: it is harder to generate novel areas from scratch than to inpaint around existing panoramic observations.
While 4 initial panoramic images enable the creation of larger scenes, the frames of trajectories without a starting image are less 3D-consistent.
This results in floating artifacts during rendering.
Our proposed scene memory mechanism is necessary, as each individual trajectory only covers a local area of the scene.
Thus, conditioning only within that trajectory similarly creates 3D-inconsistent scenes.
Our collision detection ensures, that the pre-defined trajectories do not intersect with generated scene content like the archway towards the kitchen area.
Without it, these areas become more fuzzy due to degenerate videos running into these structures.
}
\label{fig:ours-ablation}
\end{figure*}

%% file: tables/fig_ours_scenes_updated.tex
\begin{figure*}
\centering
\setlength\tabcolsep{1pt}         
\renewcommand{\arraystretch}{0.5}   

\begin{tabular}{cccccc}

& Generated scene overview & \multicolumn{3}{c}{Rendered novel views} \\
\multirow{2}{*}[0.07\textwidth]{\rotatebox{90}{\textit{"airy Hamptons beach house"}}} &
\includegraphics[height=0.15\textwidth]{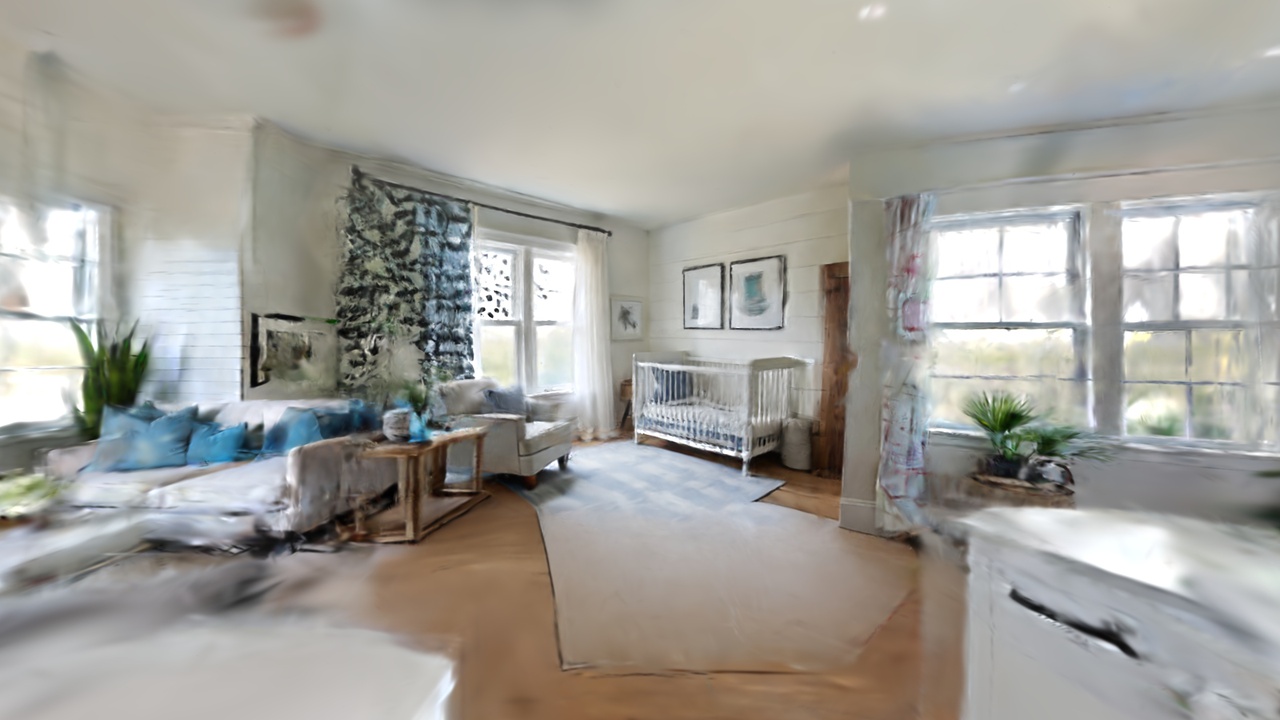} &
\includegraphics[height=0.15\textwidth]{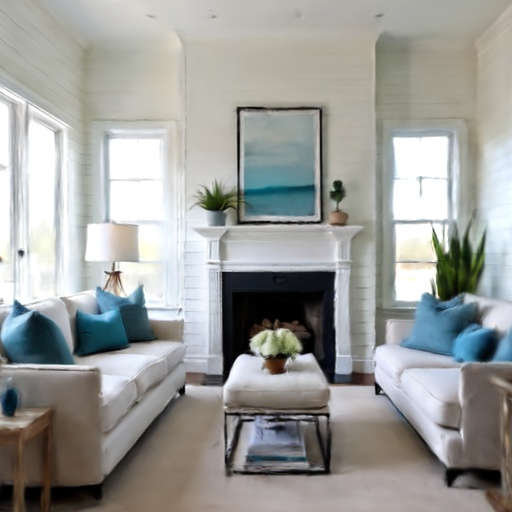} &
\includegraphics[height=0.15\textwidth]{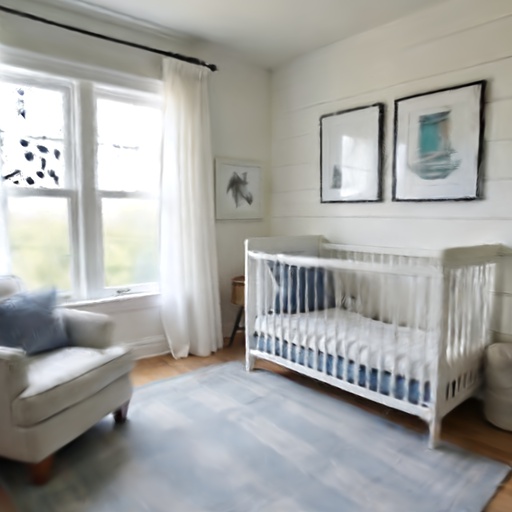} &
\includegraphics[height=0.15\textwidth]{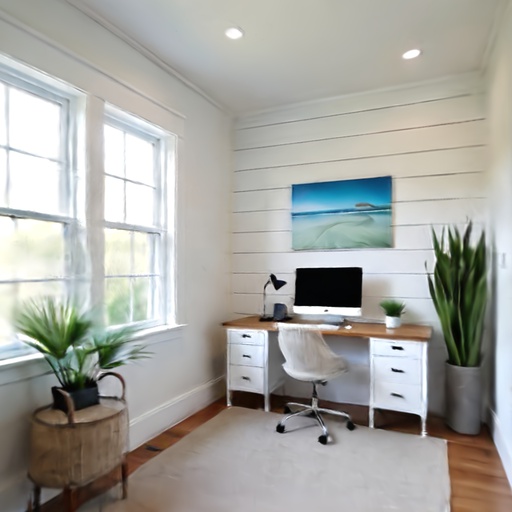} &
\includegraphics[height=0.15\textwidth]{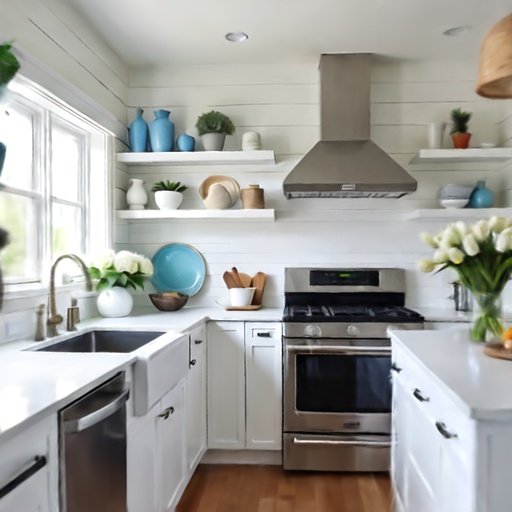} \\

&  %
\includegraphics[height=0.15\textwidth]{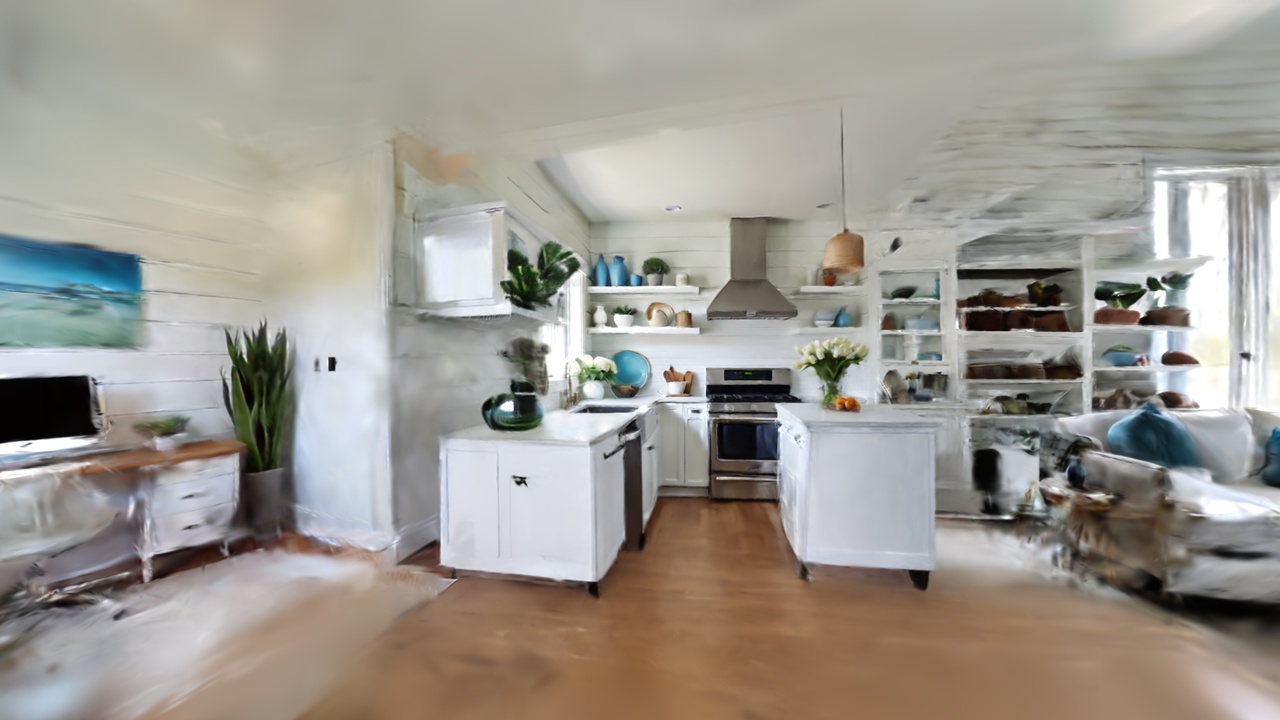} &
\includegraphics[height=0.15\textwidth]{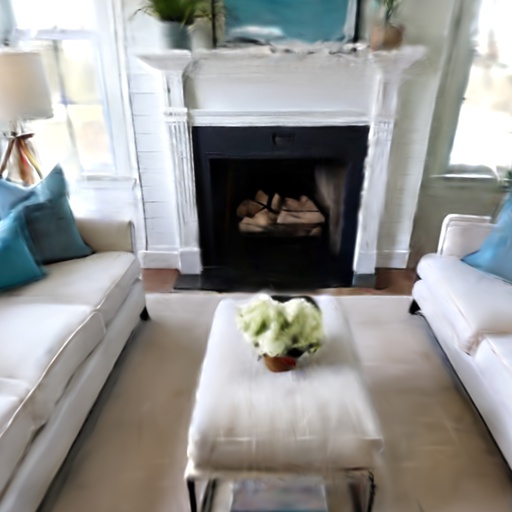} &
\includegraphics[height=0.15\textwidth]{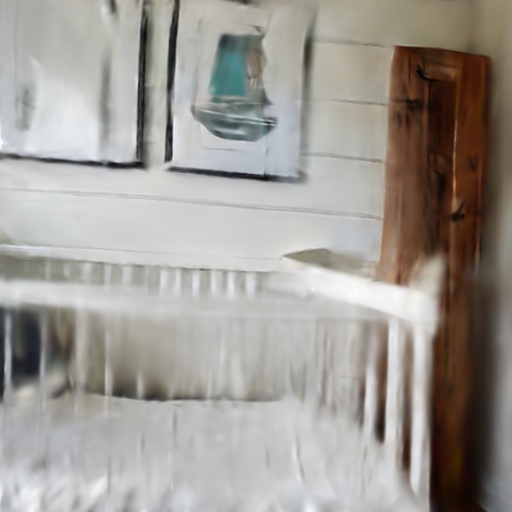} &
\includegraphics[height=0.15\textwidth]{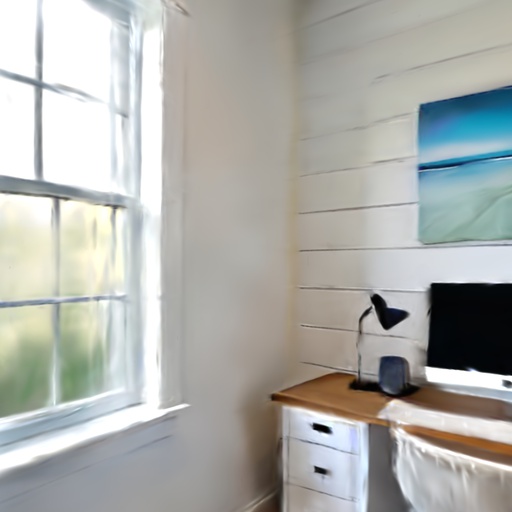} &
\includegraphics[height=0.15\textwidth]{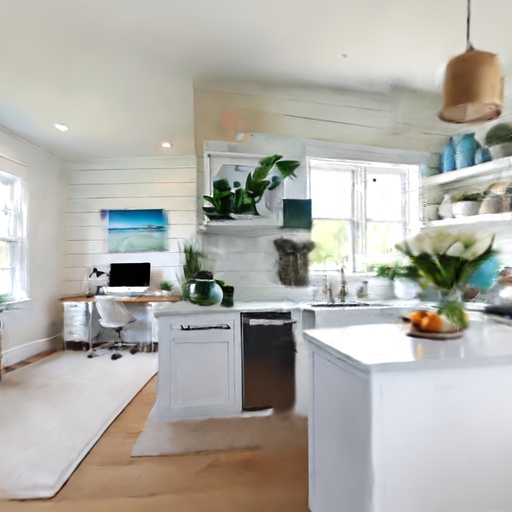} \\

\multirow{2}{*}[0.07\textwidth]{\rotatebox{90}{\textit{"melodic crystal music forest"}}} &
\includegraphics[height=0.15\textwidth]{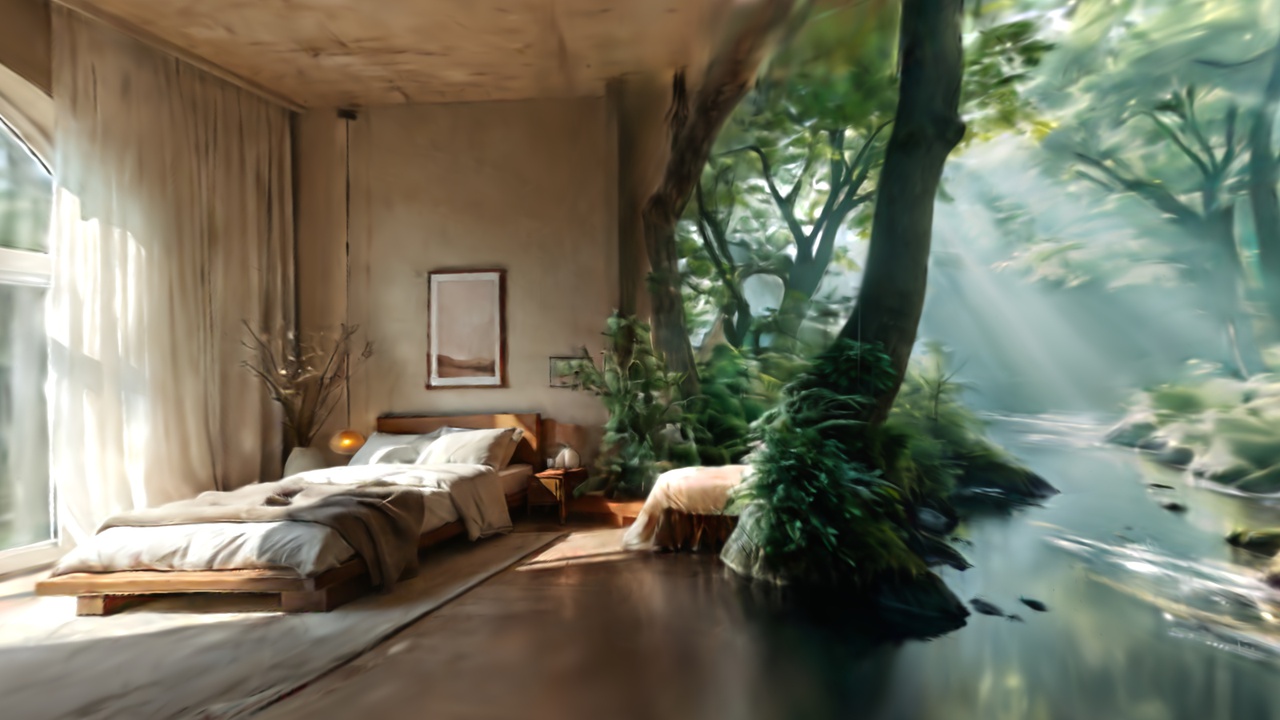} &
\includegraphics[height=0.15\textwidth]{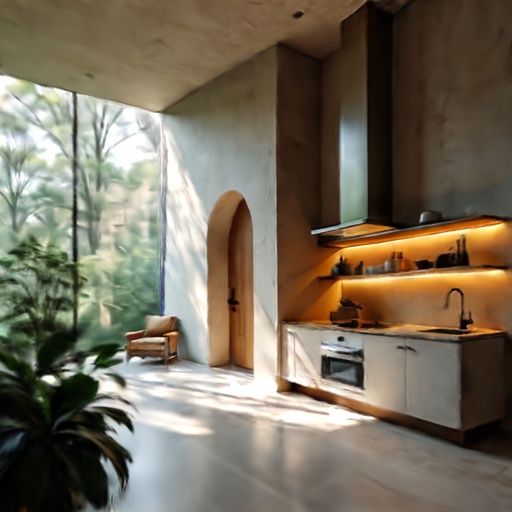} &
\includegraphics[height=0.15\textwidth]{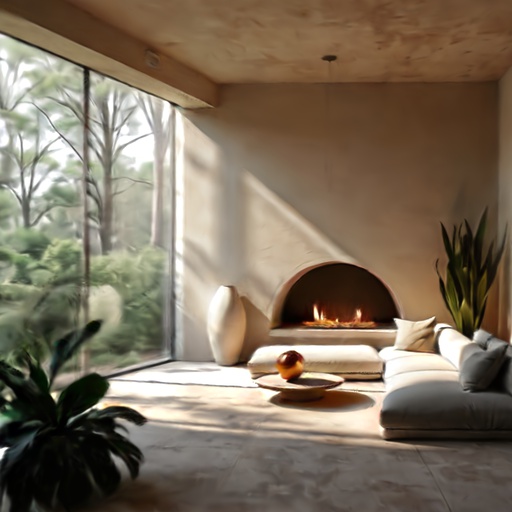} &
\includegraphics[height=0.15\textwidth]{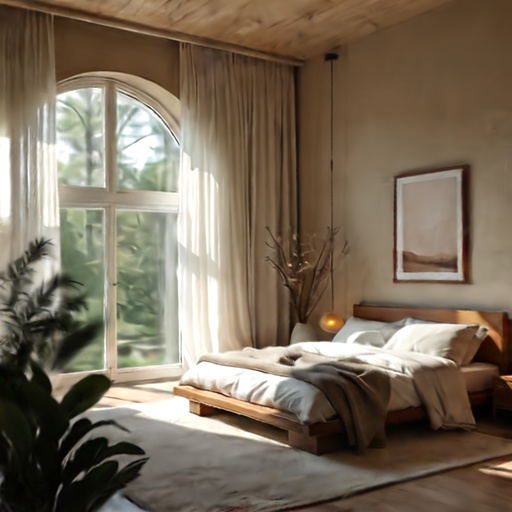} &
\includegraphics[height=0.15\textwidth]{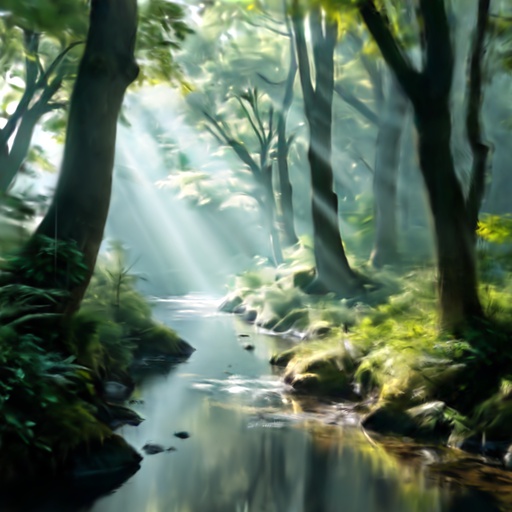} \\

&  %
\includegraphics[height=0.15\textwidth]{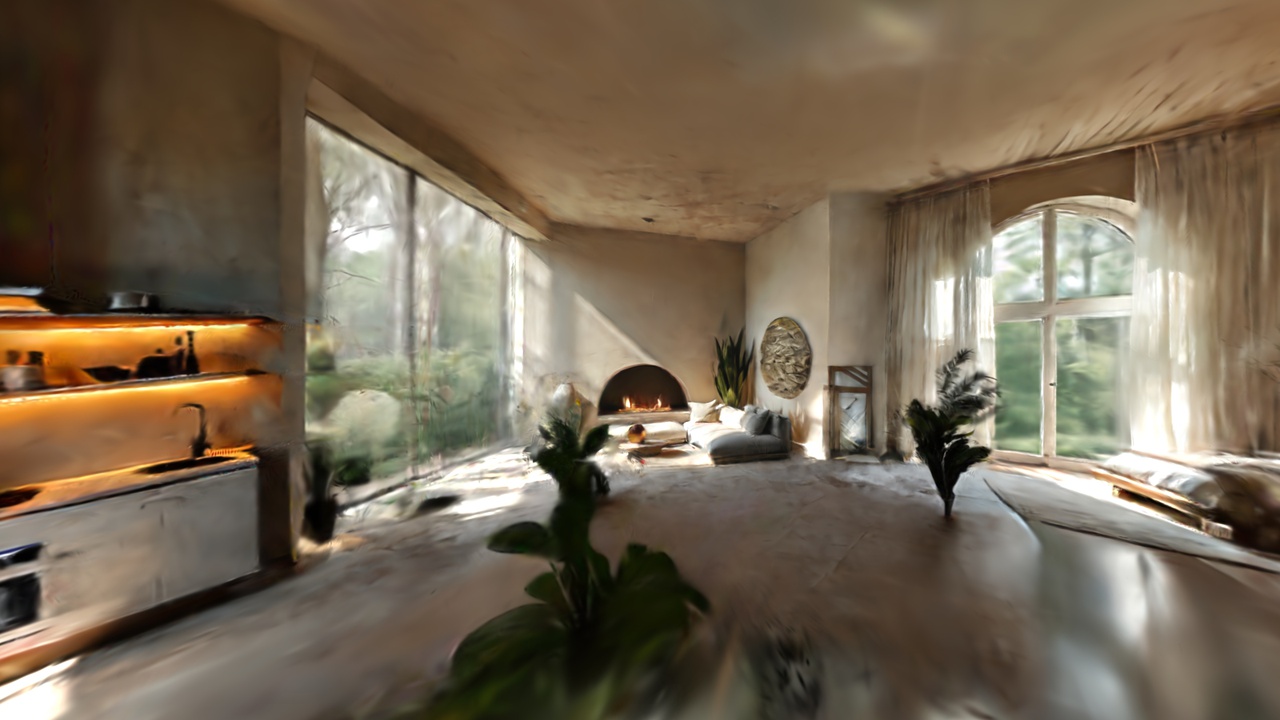} &
\includegraphics[height=0.15\textwidth]{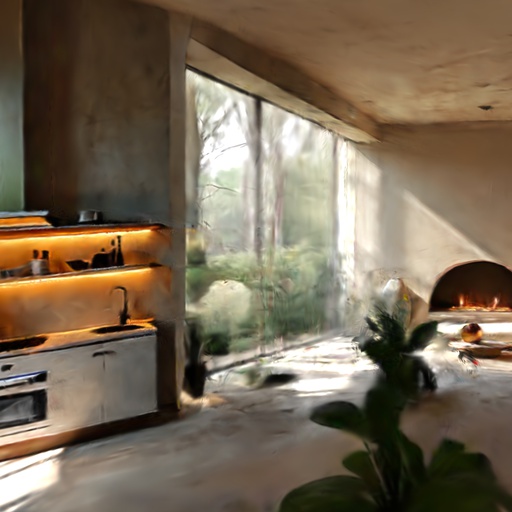} &
\includegraphics[height=0.15\textwidth]{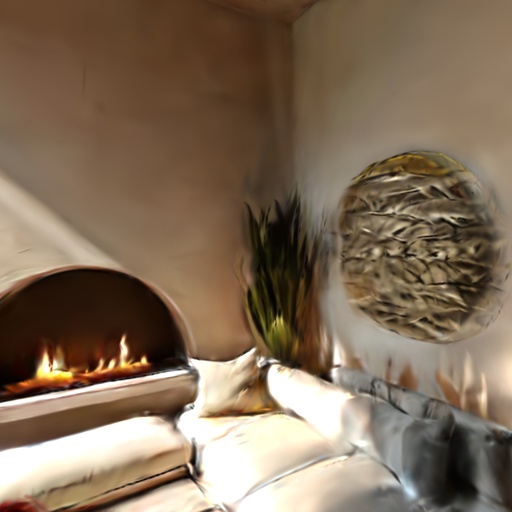} &
\includegraphics[height=0.15\textwidth]{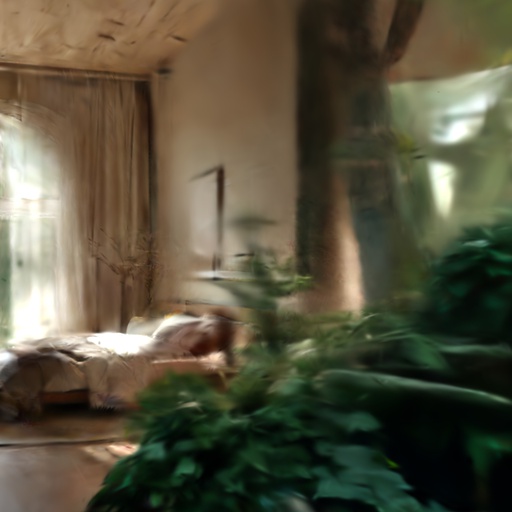} &
\includegraphics[height=0.15\textwidth]{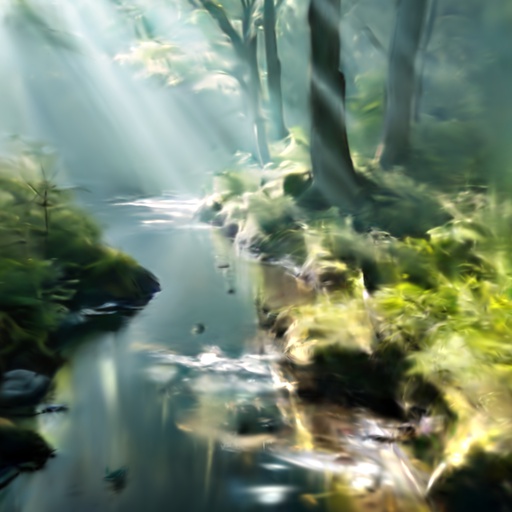} \\

\multirow{2}{*}[0.07\textwidth]{\rotatebox{90}{\textit{"liquid metal monastery"}}} &
\includegraphics[height=0.15\textwidth]{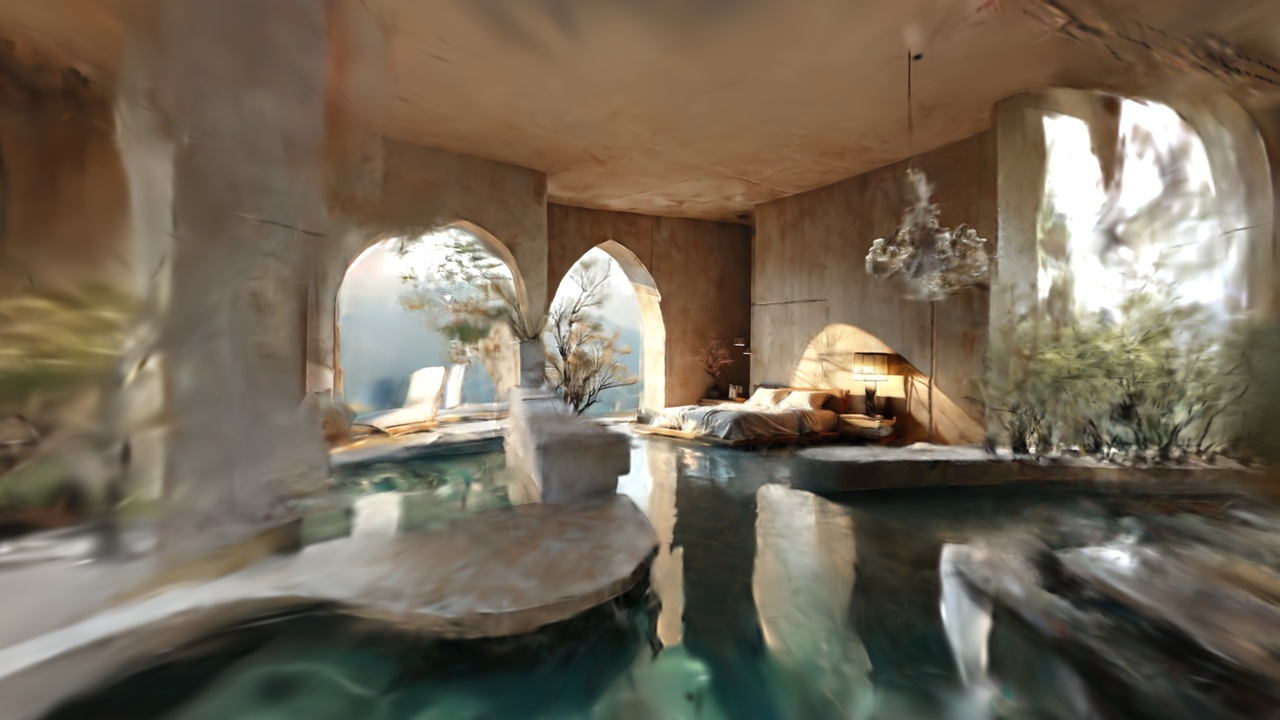} &
\includegraphics[height=0.15\textwidth]{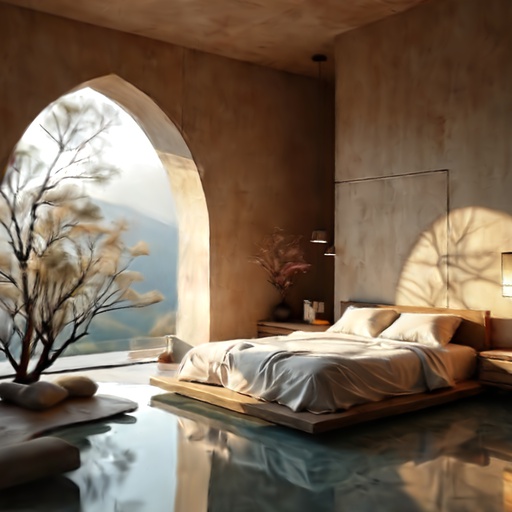} &
\includegraphics[height=0.15\textwidth]{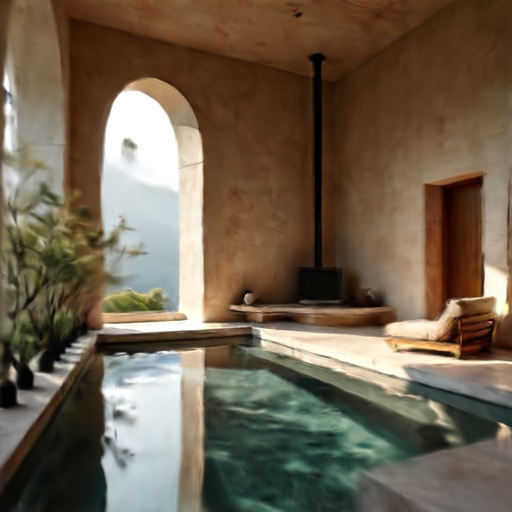} &
\includegraphics[height=0.15\textwidth]{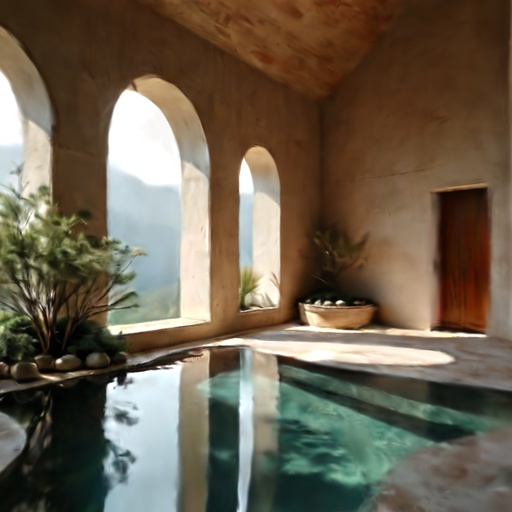} &
\includegraphics[height=0.15\textwidth]{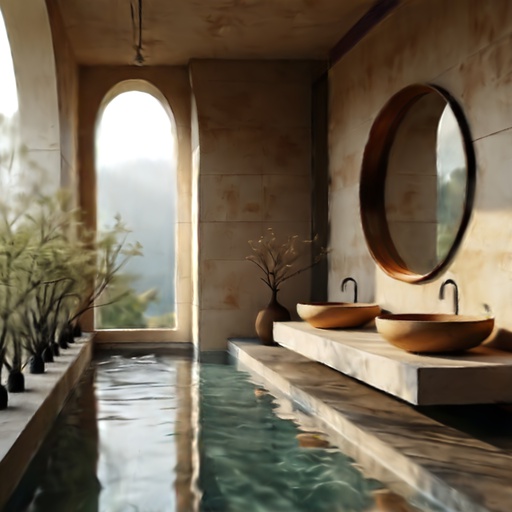} \\

&  %
\includegraphics[height=0.15\textwidth]{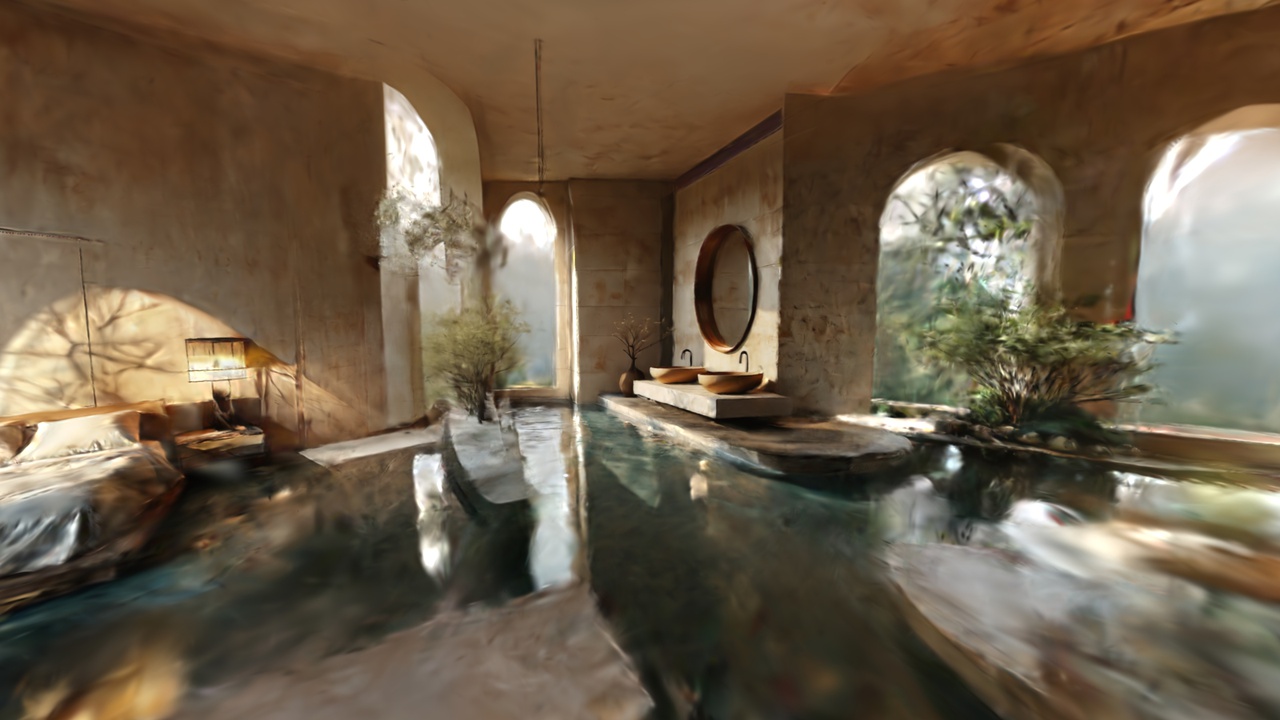} &
\includegraphics[height=0.15\textwidth]{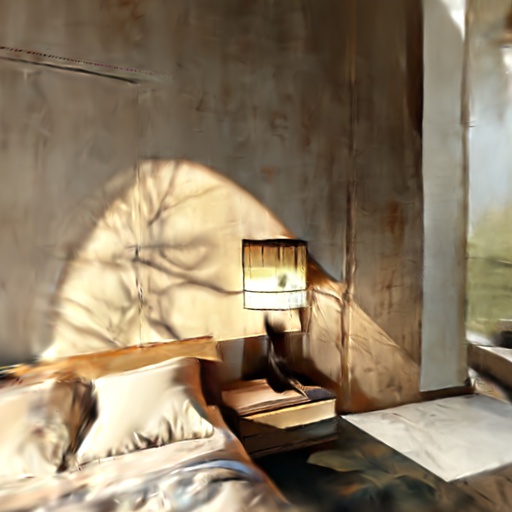} &
\includegraphics[height=0.15\textwidth]{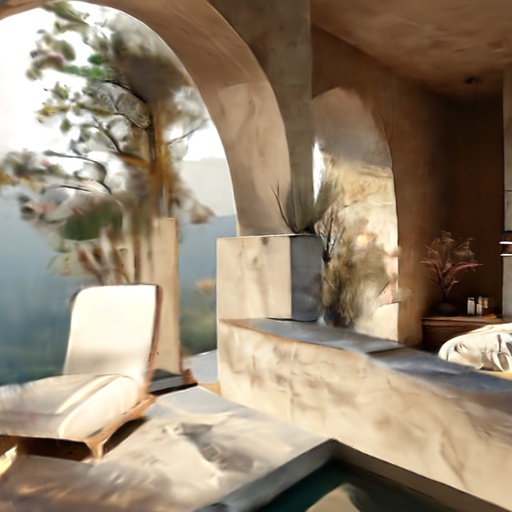} &
\includegraphics[height=0.15\textwidth]{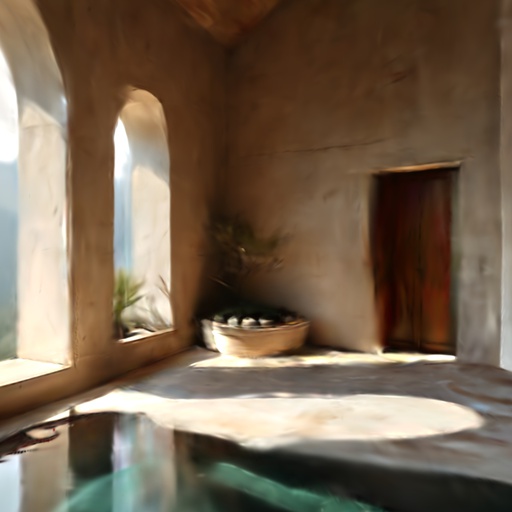} &
\includegraphics[height=0.15\textwidth]{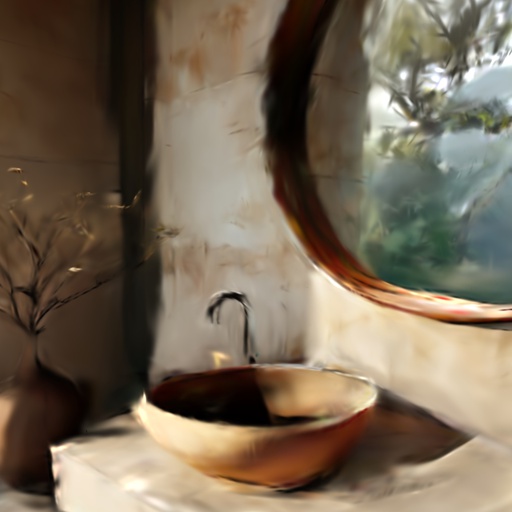}
\end{tabular}

\vspace{-4mm}
\caption{
\textbf{World generation results of our method.}
We visualize scene overview renderings close to the scene center facing in two opposite directions.
Additionally, we render novel views far beyond the scene center, that explore the different generated areas in more detail.
This showcases the possibility to interactively explore our worlds from arbitrary viewpoints.
We also demonstrate the diversity of our scenes by mixing indoor/outdoor elements in a single 3D scene.
Please see the supplementary material for animated flythroughs of our generated scenes.
}
\Description{
\textbf{World generation results of our method.}
We visualize scene overview renderings close to the scene center facing in two opposite directions.
Additionally, we render novel views far beyond the scene center, that explore the different generated areas in more detail.
This showcases the possibility to interactively explore our worlds from arbitrary viewpoints.
We also demonstrate the diversity of our scenes by mixing indoor/outdoor elements in a single 3D scene.
Please see the supplementary material for animated flythroughs of our generated scenes.
}
\label{fig:ours-scenes}
\end{figure*}

%% file: sections/5conclusion.tex
\section{Conclusion}
We have presented \OURS, a novel method for text-to-3D scene generation.
By utilizing video diffusion models in a scene generation scheme, we iteratively expand the generated scene from a panoramic scaffold towards fully navigable 3D environments.
The core insight of our method is a tailored scene memory mechanism for our pre-defined video trajectories.
It allows us to condition the video diffusion model on the most relevant previous frames and thus ensures 3D-consistent image generations.
We further demonstrate, our panorama stage as a helpful initializer for creating diverse environments.
Our output 3DGS scenes can be rendered from arbitrary novel viewpoints in real-time with classical rasterization pipelines.
We believe our method demonstrates an important step towards generating immersive 3D environments from only text input.

%% file: sections/Xsuppl.tex
\maketitlesupplementary
\appendix

\section{Supplemental Video}
Please watch our attached video~\footnote[1]{\url{https://youtu.be/N6NJsNyiv6I}} for a comprehensive evaluation of the proposed method.
We include rendered videos of multiple generated scenes from novel trajectories, that showcase the quality of our large-scale 3D scenes under various non-centered perspectives.
We also compare against baselines and ablations of our method by showing rendered videos.

\section{User Study Details}
\input{tables/fig_user_study}

We conduct a user study to determine Perceptual Quality (PQ) and 3D Consistency (3DC) of our 3D scenes.
In order to determine these scores, we asked participants to rate our scenes and baselines based on overall quality and 3D-consistency on a scale of $1{-}5$.
\Cref{fig:user-study} shows an example of how we asked the users to score the above two metrics.
We show the users short videos of trajectories rendered from our final scene, as well as from baselines, both displaying large camera motion. 
In total, we received 2432 datapoints from $n{=}64$ participants.

\section{Additional Baseline Details}
We compare against several different types of methods for 3D scene generation.
Here, we provide additional details about the chosen baselines.

\paragraph{Panorama-To-3D Generators}
\input{tables/fig_pano_baseline}

Since DreamScene360~\cite{zhou2024dreamscene360} and LayerPano3D~\cite{yang2024layerpano3d} can only be initialized from text input or 360 degree panorama input, we cannot directly provide our initial panoramic images to them.
To ensure a fair comparison, we instead provide these methods the identical text prompt.
Since their text-to-panorama backbone is built on a different T2I model than ours, we still obtain initialization with slightly different style.
For consistency between the baselines, we generate one panorama using LayerPano3D~\cite{yang2024layerpano3d} and use it for both baselines.
We visualize the panoramas for two of our comparison scenes in ~\Cref{fig:pano-baseline} and also compare them against our panoramic scene initialization.

\paragraph{Text2Room~\cite{hollein2023text2room}}
We provide one of our four initial images as input to the method.
We run it with their default hyperparameters.

\paragraph{FlexWorld~\cite{chen2025flexworld}}
We provide one of our four initial images as input to the method.
We run it with their default hyperparameters.

\paragraph{WonderWorld~\cite{yu2024wonderworld}}
We provide one of our four initial images as input to the method.
Their interactive scene generation scheme allows for fine-grained control over the next viewpoint for inpainting and its corresponding text prompt.
We carefully select viewpoints in a manner similar to the original presented results.

\paragraph{SEVA~\cite{zhou2025stable}}
\input{tables/fig_seva_video}

We follow the original codebase to generate an entire 3D scene starting from a single image.
Concretely, we manually define a single, continuous trajectory starting from one of our four initial panoramic images.
This trajectory covers the entire 360 degree viewpoints around that start image by a mostly horizontal and outward-looking trajectory.
However, coming up with a suitable trajectory a-priori from a single start image is hard, as the generated scene content is unknown.
We visualize the generated video output in ~\Cref{fig:seva-video}.
As can be seen, the video suffers from high-frequency flickering and inconsistencies across views, especially when moving far away from the initial image.
As a result, the optimized 3DGS~\cite{kerbl20233d} scene in our main results suffers from blurriness and duplicated content.
This highlights the necessity for an iterative scene generation scheme, that we propose on top of this camera-guided video diffusion model.
In other words, our pre-defined trajectories provide a robust way to split scene generation into multiple local parts and our panorama initialization provides the necessary global conditioning for this iterative generation.

\section{Limitations}
\input{tables/fig_limitations}

Our method generates high-quality 3D scenes from text as input, that can be fully explored, i.e., we can interactively move into scene parts and rotate around objects without quality degradation.
Nevertheless, there are remaining limitations (see ~\Cref{fig:limitations}).
As shown in \Cref{fig:limitations} top and mid, our method is still susceptible to floaters and blurriness in the final 3D reconstruction from certain perspectives. 
This is partially the case since synthetically generated images are typically never fully 3D consistent, especially across thousands of individual images.

Additionally, our pre-defined trajectory scheme may not cover all areas of a generated scene.
We visualize in \Cref{fig:limitations}bottom the front and back view of an armchair.
Since the backside of the armchair is very close to the wall behind it, our pre-defined trajectories did never observe that part of the object.
As a result, the 3D reconstruction remains inconsistent in these areas.

\section{Runtime/Memory Comparison}
\input{tables/tab_runtime}
Our method generates fully-explorable scenes, as opposed to baseline-scenes viewable from limited camera-poses.
We generate more observations in order to resolve occlusions and increase the 3D completeness. 
This yields higher-quality scenes, as measured by Perceptual-Quality (PQ) and 3D-Consistency (3DC) of our user-study in \Cref{tab:ours-baseline}.
Therefore, we report runtime amortized per generated frame on the “celestial library” scene in \Cref{tab:runtime}.
We achieve comparable per-frame generation rates and memory consumption as baselines approaches.
Our runtime is dominated by the second stage (\textasciitilde 10min/trajectory), while the first stage (\textasciitilde 5min) and third stage (\textasciitilde 11min) are comparatively faster.
We believe real-time video-diffusion will reduce runtime towards interactive generation.

\section{Scene Layout Diversity}
\input{tables/fig_l_shaped}
Our panoramic scaffold (\Cref{subsec:pano}) determines the global scene layout, i.e., which room types are combined.
In our main results, we create the scaffold by using the same text prompt, but four different room prefixes (e.g. \textit{``kitchen/office of XYZ''}).
This creates scenes with a ``cross-shaped'' layout, where each of the four rooms becomes an explorable area.

We showcase in \Cref{fig:l-shaped}, that our method can also generate other layouts (e.g., ``L-shaped'' scenes) by adjusting the panoramic scaffold as desired.
Concretely, we replace two room prefixes with \textit{``wall/window of XYZ''} to generate flat, non-explorable areas.
This demonstrates the diversity of our approach: it is possible to customize the generated scene layouts by adjusting the panoramic scene scaffold accordingly.

\section{Additional Qualitative Results}
\input{tables/fig_ours_scenes_suppl_1}
\input{tables/fig_ours_scenes_suppl_2}
\input{tables/fig_ours_scenes_suppl_3}

We show additional qualitative results of our method in ~\Cref{fig:ours-scenes-suppl-1}, ~\Cref{fig:ours-scenes-suppl-2}, and ~\Cref{fig:ours-scenes-suppl-3}.

%% file: tables/fig_user_study.tex
\begin{figure}
\centering
\includegraphics[width=0.5\textwidth]{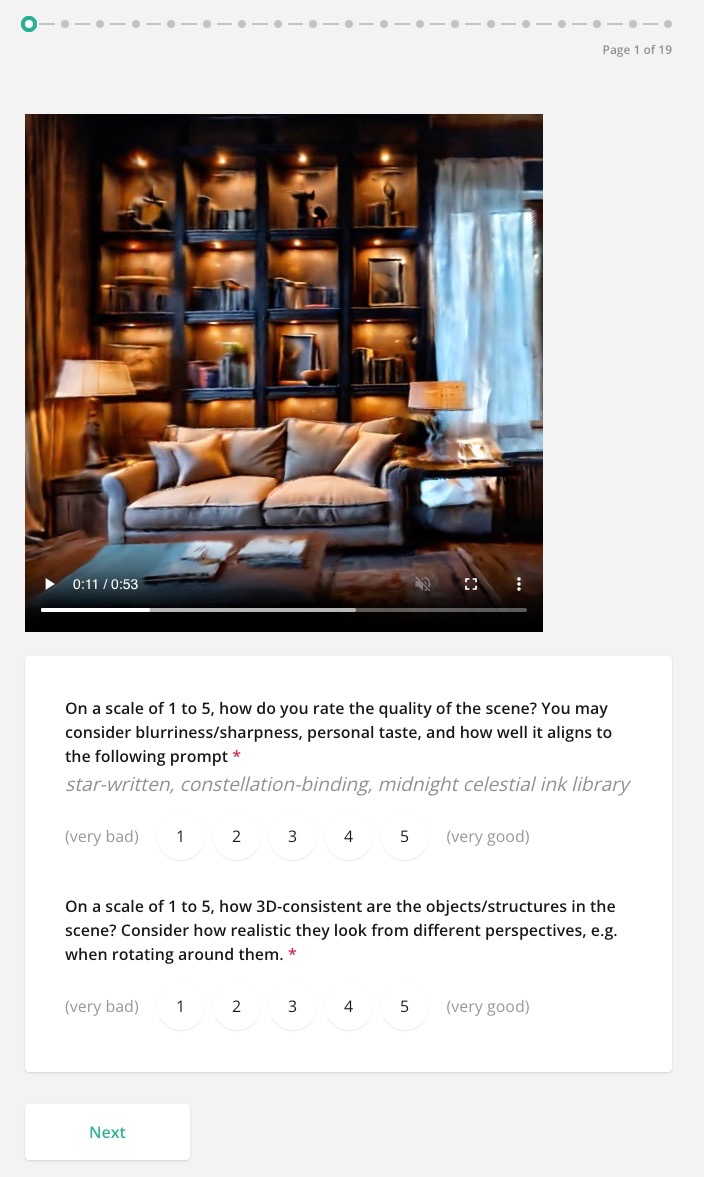}
\caption{\textbf{User study interface}. (Top) We provide users with videos along trajectories with large camera motion for our and baseline scenes. (Left) We ask users to scale the Perceptual Quality (PC) and 3D Consistency (3DC) on a scale from 1 to 5. 
}
\Description{\textbf{User study interface}. (Top) We provide users with videos along trajectories with large camera motion for our and baseline scenes. (Left) We ask users to scale the Perceptual Quality (PC) and 3D Consistency (3DC) on a scale from 1 to 5. 
}
\label{fig:user-study}
\end{figure}

%% file: tables/fig_pano_baseline.tex
\begin{figure*}
\centering
\setlength\tabcolsep{1pt}         
\renewcommand{\arraystretch}{1}   

\begin{tabular}{cc}
\includegraphics[width=0.45\linewidth]{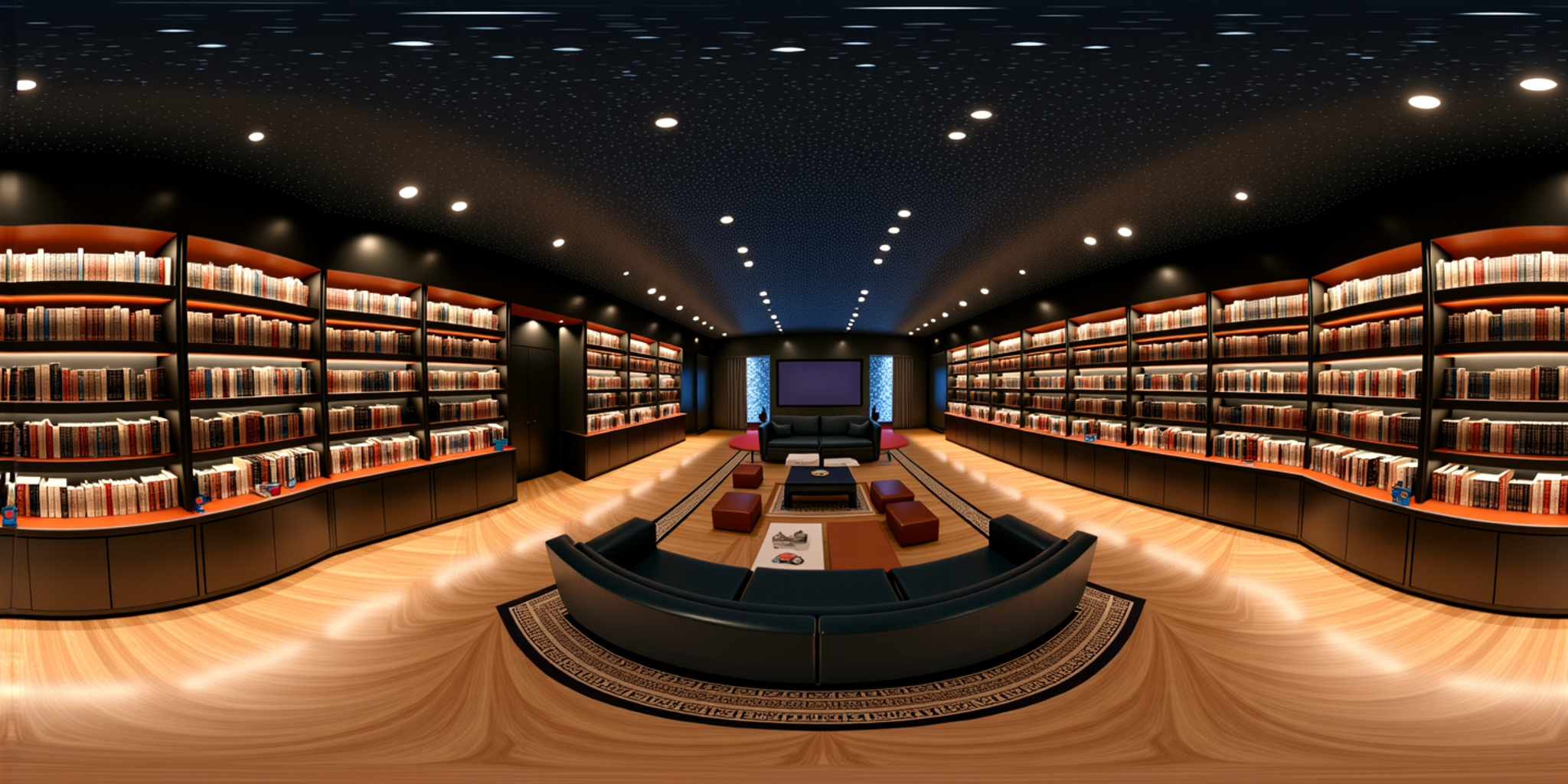} &
\includegraphics[width=0.45\linewidth]{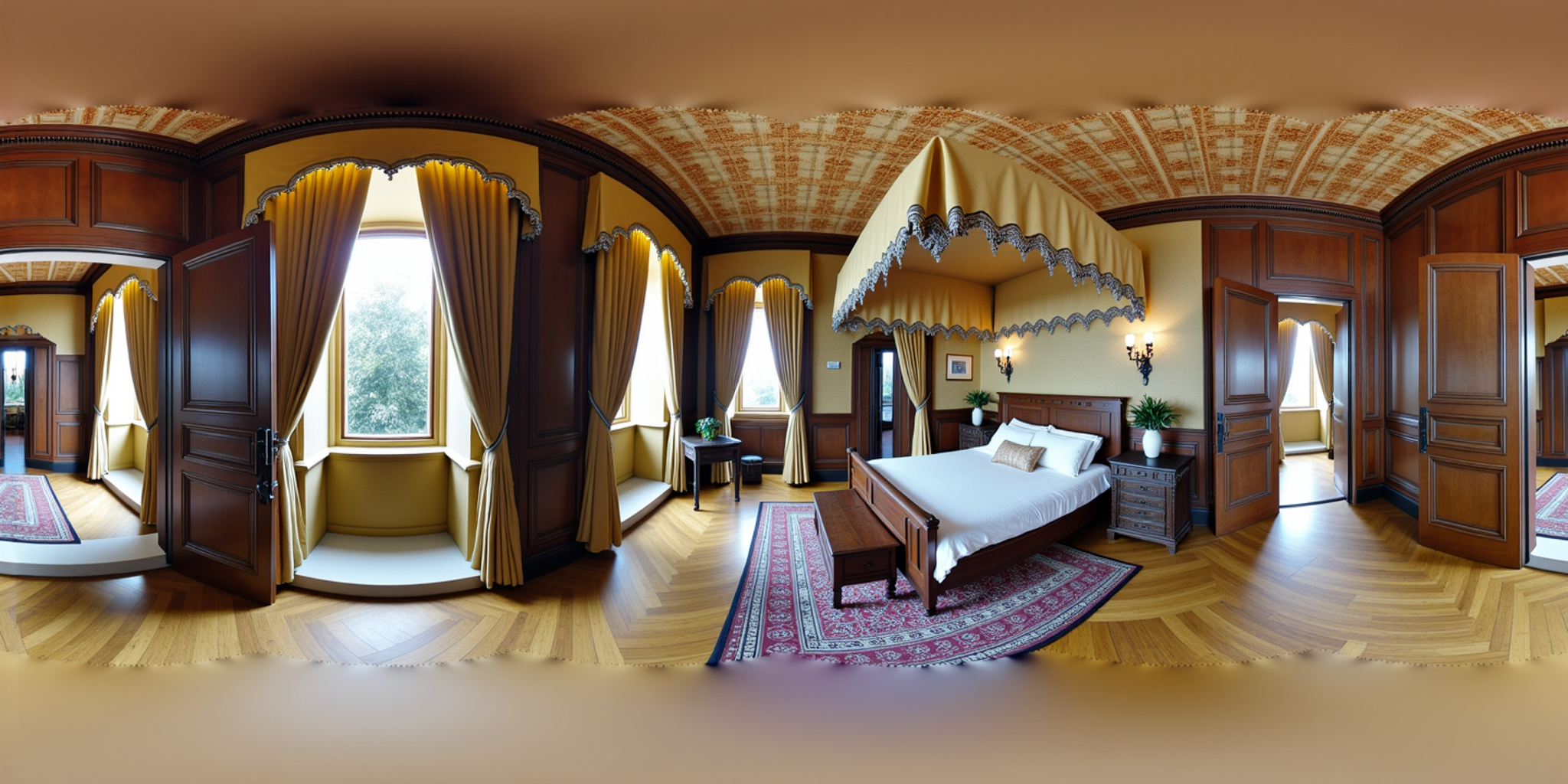} \\
(a) LayerPano3D~\cite{yang2024layerpano3d} for \textit{``celestial ink library''} & (b) LayerPano3D~\cite{yang2024layerpano3d} for \textit{``gothic revival mansion''} \\

\multicolumn{2}{c}{\includegraphics[width=0.95\linewidth]{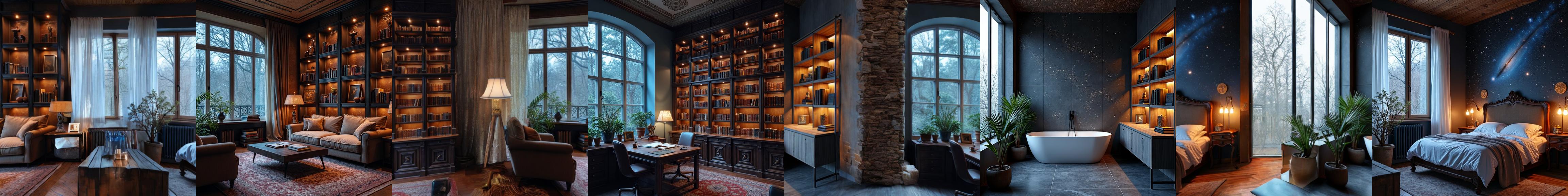}} \\
\multicolumn{2}{c}{The 8 initial panoramic images of our method for \textit{``celestial ink library''}} \\

\multicolumn{2}{c}{\includegraphics[width=0.95\linewidth]{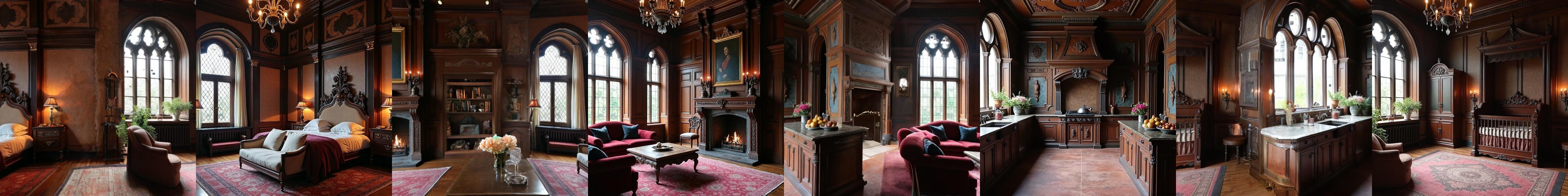}} \\
\multicolumn{2}{c}{The 8 initial panoramic images of our method for \textit{``gothic revival mansion''}} \\

\end{tabular}
\caption{
\textbf{Intermediate baseline outputs.} 
We compare the panorama generation of~\cite{yang2024layerpano3d} against our first stage panorama initialization (\Cref{subsec:pano}).
Since we generate four images independently from each other, we can obtain environments with higher diversity than the baselines (e.g., it is possible to specify certain room types like ``living room'' or ``kitchen'' separately for each of the four images).
}
\Description{
\textbf{Intermediate baseline outputs.} 
We compare the panorama generation of~\cite{yang2024layerpano3d} against our first stage panorama initialization (\Cref{subsec:pano}).
Since we generate four images independently from each other, we can obtain environments with higher diversity than the baselines (e.g., it is possible to specify certain room types like ``living room'' or ``kitchen'' separately for each of the four images).
}
\label{fig:pano-baseline}
\end{figure*}

%% file: tables/fig_seva_video.tex
\begin{figure*}
\centering
\setlength\tabcolsep{1pt}         
\renewcommand{\arraystretch}{1}

\begin{center}
  \begin{tabular}{*{6}{c}}
    \includegraphics[width=0.15\textwidth,height=0.15\textwidth,keepaspectratio]{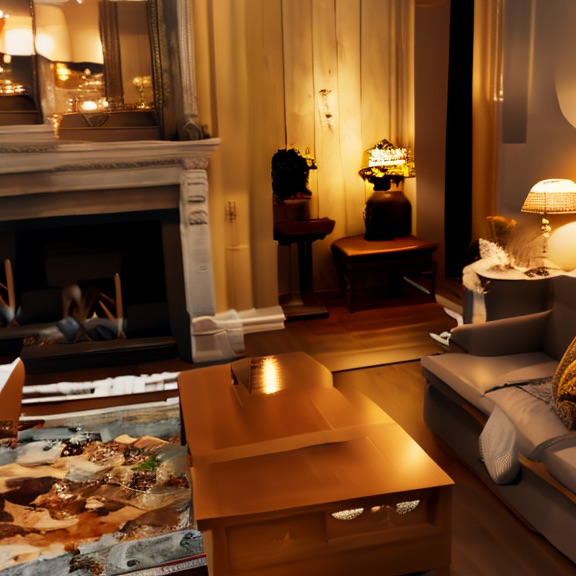}  &
    \includegraphics[width=0.15\textwidth,height=0.15\textwidth,keepaspectratio]{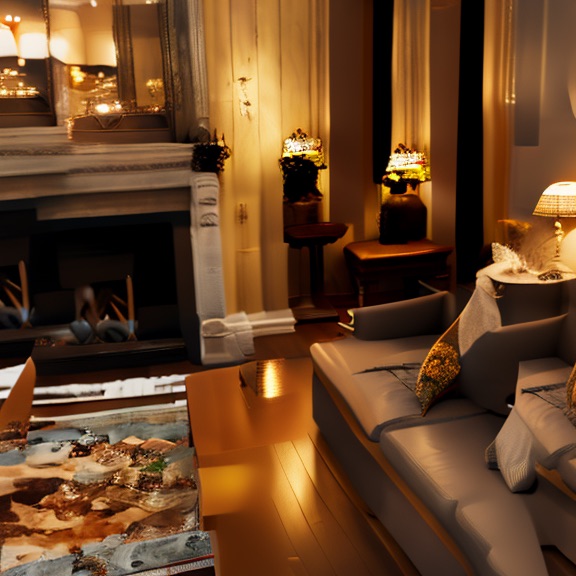}  &
    \includegraphics[width=0.15\textwidth,height=0.15\textwidth,keepaspectratio]{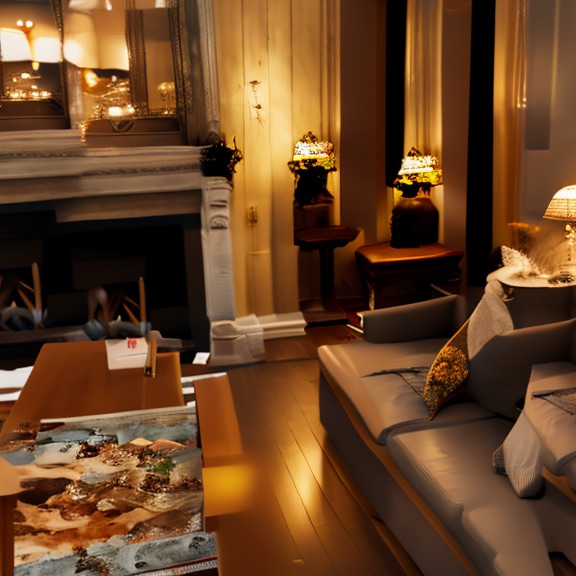}  &
    \includegraphics[width=0.15\textwidth,height=0.15\textwidth,keepaspectratio]{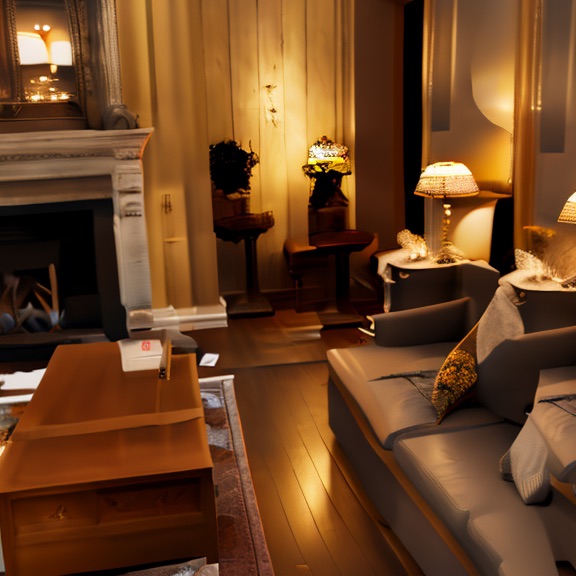}  &
    \includegraphics[width=0.15\textwidth,height=0.15\textwidth,keepaspectratio]{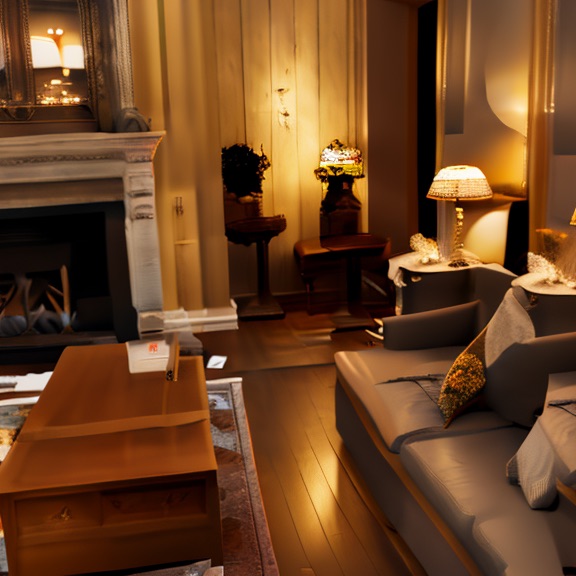}  &
    \includegraphics[width=0.15\textwidth,height=0.15\textwidth,keepaspectratio]{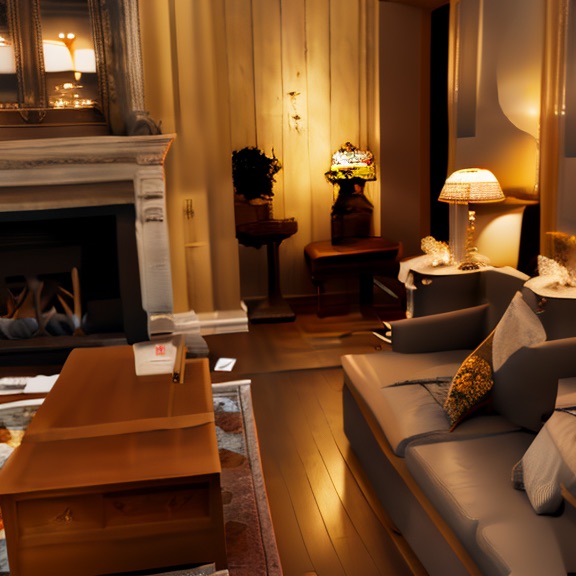}  \\
    \includegraphics[width=0.15\textwidth,height=0.15\textwidth,keepaspectratio]{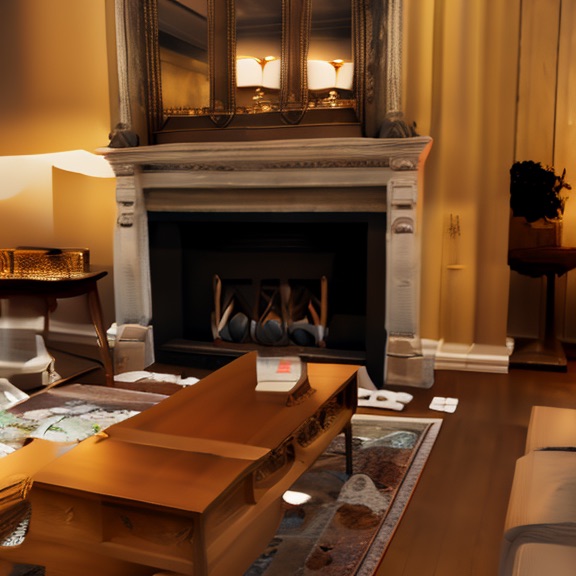}  &
    \includegraphics[width=0.15\textwidth,height=0.15\textwidth,keepaspectratio]{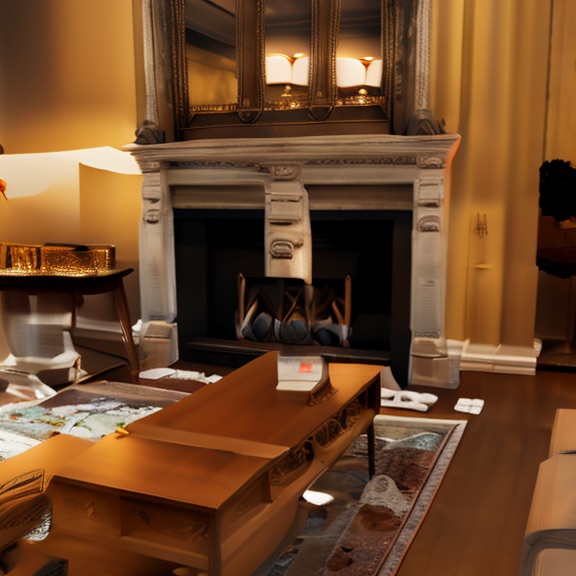}  &
    \includegraphics[width=0.15\textwidth,height=0.15\textwidth,keepaspectratio]{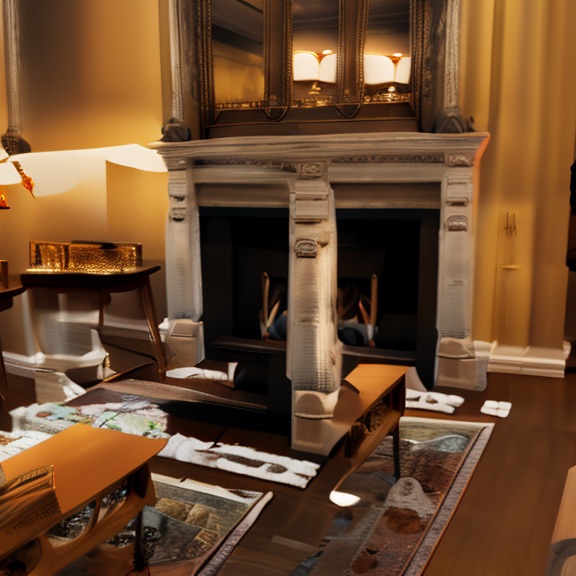}  &
    \includegraphics[width=0.15\textwidth,height=0.15\textwidth,keepaspectratio]{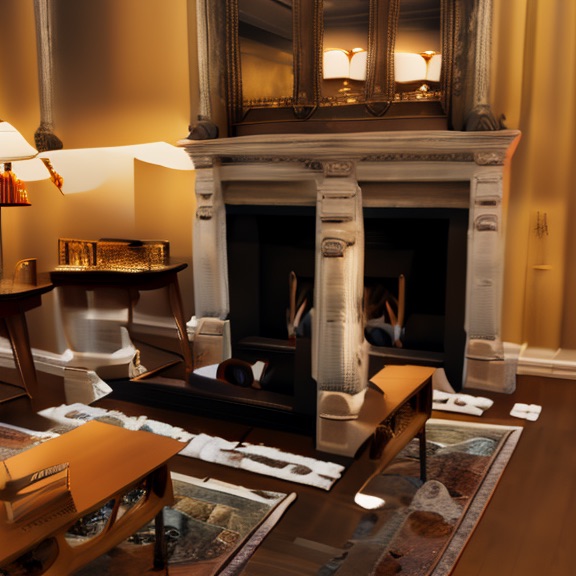}  &
    \includegraphics[width=0.15\textwidth,height=0.15\textwidth,keepaspectratio]{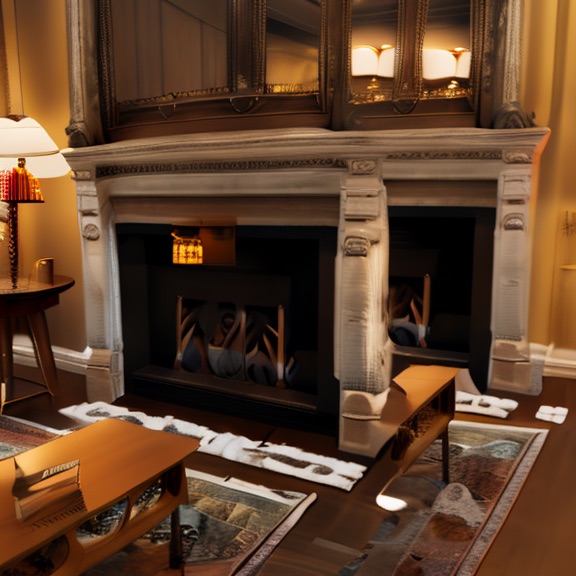}  &
    \includegraphics[width=0.15\textwidth,height=0.15\textwidth,keepaspectratio]{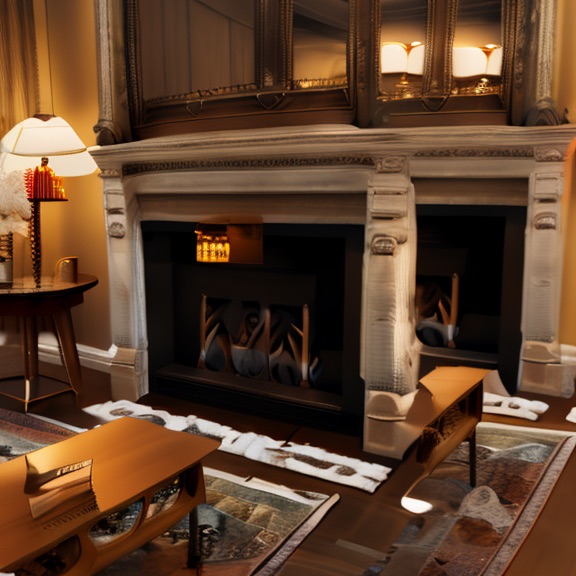}  \\
  \end{tabular}
\end{center}
\caption{
\textbf{Intermediate video generation results of ~\cite{zhou2025stable}.}
The above frames represents the video output generated along a single, continuous trajectory from the SEVA baseline~\cite{zhou2025stable} initialized from a single input image on the \textit{``celestial ink library''} scene.
Even though we have chosen a high guidance scale of 6.0 for the generation, we still obtain inconsistent results (e.g. multiple different tables are generated in the bottom row).
This leads to visible blurriness and duplicated object results when optimizing a 3DGS ~\cite{kerbl20233d} representation from these images.
In contrast, our full pipeline divides scene generation into multiple smaller videos that are further conditioned on 8 initial panoramic images.
The resulting scenes from our approach exhibit 3D-consistent and high quality renderings.
}
\Description{
\textbf{Intermediate video generation results of ~\cite{zhou2025stable}.}
The above frames represents the video output generated along a single, continuous trajectory from the SEVA baseline~\cite{zhou2025stable} initialized from a single input image on the \textit{``celestial ink library''} scene.
Even though we have chosen a high guidance scale of 6.0 for the generation, we still obtain inconsistent results (e.g. multiple different tables are generated in the bottom row).
This leads to visible blurriness and duplicated object results when optimizing a 3DGS ~\cite{kerbl20233d} representation from these images.
In contrast, our full pipeline divides scene generation into multiple smaller videos that are further conditioned on 8 initial panoramic images.
The resulting scenes from our approach exhibit 3D-consistent and high quality renderings.
}
\label{fig:seva-video}
\end{figure*}

%% file: tables/fig_limitations.tex
\begin{figure}
\centering
\setlength\tabcolsep{1pt}         
\begin{tabular}{cc}
\includegraphics[width=0.23\linewidth]{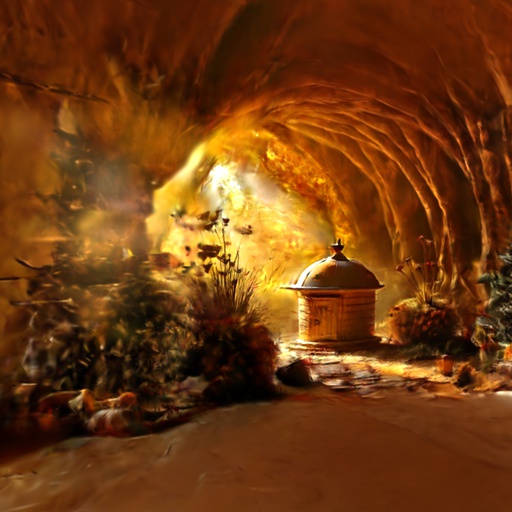} &
\includegraphics[width=0.23\linewidth]{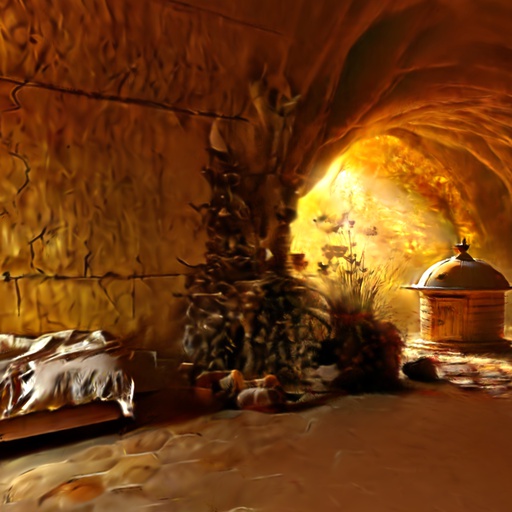} \\
\includegraphics[width=0.23\linewidth]{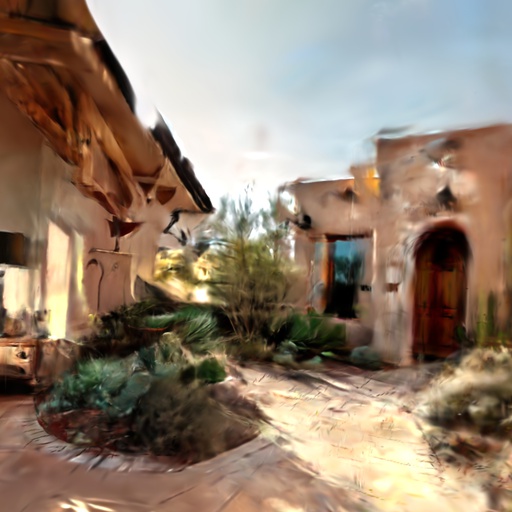} &
\includegraphics[width=0.23\linewidth]{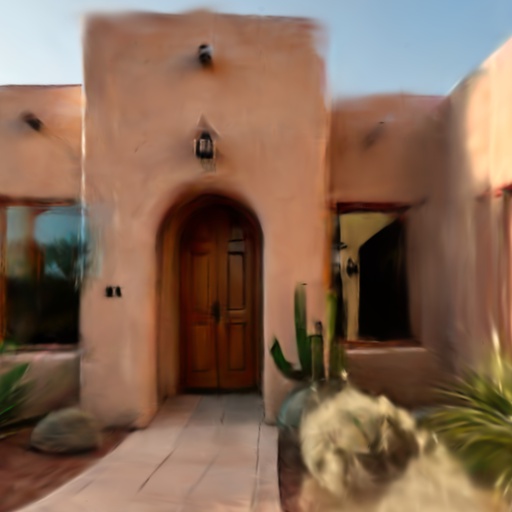} \\
\includegraphics[width=0.23\linewidth]{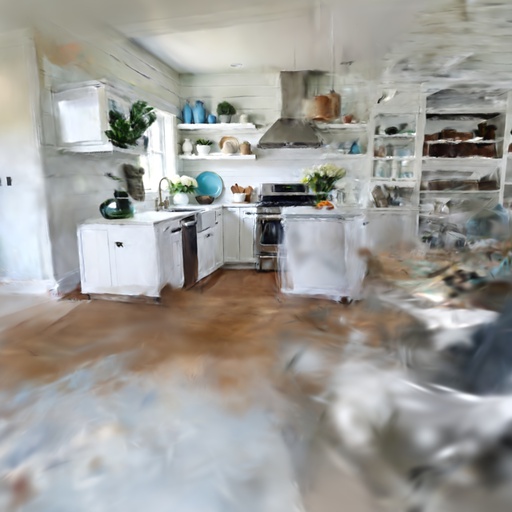} &
\includegraphics[width=0.23\linewidth]{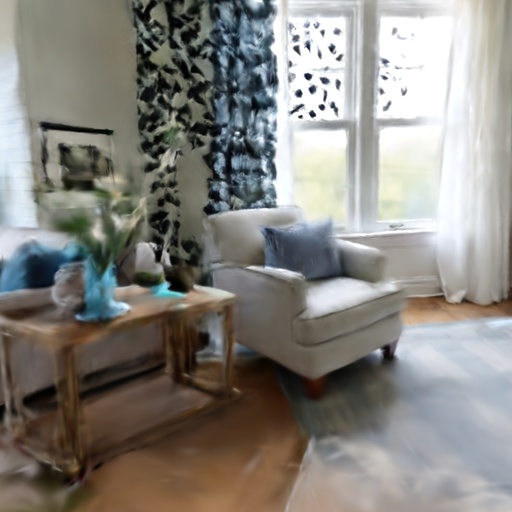}
\end{tabular}
\caption{
\textbf{Limitations of our method.} 
Top and mid: high-frequency flickering in generated video frames lead to floating artifacts and blurry textures for certain parts of our 3D reconstruction.
Bottom: objects placed at the very edge of our generated scene are rarely observed from all sides during video trajectory generation.
Thus, the reconstructed scene is underconstrained at these areas, which results in degenerate rendering results.
}
\Description{
\textbf{Limitations of our method.} 
Top and mid: high-frequency flickering in generated video frames lead to floating artifacts and blurry textures for certain parts of our 3D reconstruction.
Bottom: objects placed at the very edge of our generated scene are rarely observed from all sides during video trajectory generation.
Thus, the reconstructed scene is underconstrained at these areas, which results in degenerate rendering results.
}
\label{fig:limitations}
\end{figure}

%% file: tables/tab_runtime.tex
\begin{table*}
  \centering
    \caption{
  \textbf{Runtime/Memory comparison.}
    Our method generates fully-explorable scenes, as opposed to baseline-scenes viewable from limited camera-poses.
    We generate more observations in order to resolve occlusions and increase the 3D completeness.
    The runtime amortized per generated frame on the “celestial library” scene is comparable to baselines.
  }
  \begin{tabular}{l cc cc}
    \toprule
    Method & Total runtime & Generated frames & Runtime per frame & Memory \\
    \midrule
    DreamScene360~\cite{zhou2024dreamscene360} & 0.5h & one panorama & N/A & 16GB \\
    LayerPano3D~\cite{yang2024layerpano3d} & 0.2h & one panorama & N/A & 5GB \\
    \midrule
    Text2Room~\cite{hollein2023text2room} & 0.9h & 200 & 16.2s & 15GB \\
    WonderWorld~\cite{yu2024wonderworld} & \multicolumn{2}{c}{user-specific interactive generation} & \textasciitilde 10s & 45GB \\
    \midrule
    FlexWorld~\cite{chen2025flexworld} & 0.5h & 144 & 12.9s & 48GB \\
    SEVA~\cite{zhou2025stable} & 0.9h & 270 & 13.5s & 22GB \\
    \midrule
    \midrule
    Ours & 7h & 2251 & 14.1s & 22GB \\
    \bottomrule
  \end{tabular}
  \label{tab:runtime}
\end{table*}

%% file: tables/fig_l_shaped.tex
\begin{figure*}
\centering
\setlength\tabcolsep{1pt}
\renewcommand{\arraystretch}{1}

\begin{tabular}{cccccccc}
\includegraphics[width=0.12\linewidth]{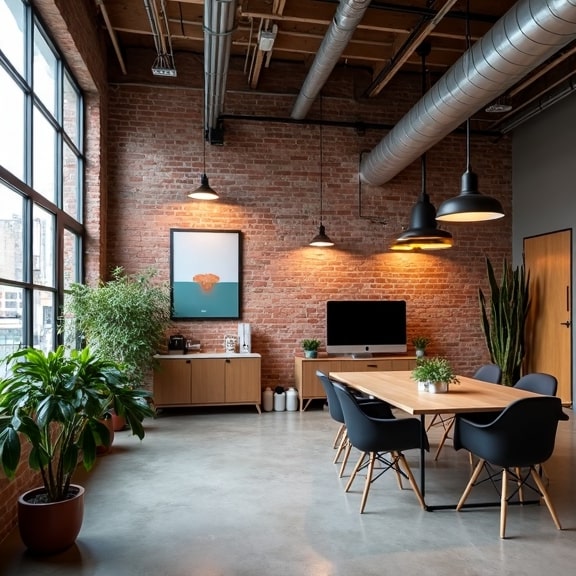} &
\includegraphics[width=0.12\linewidth]{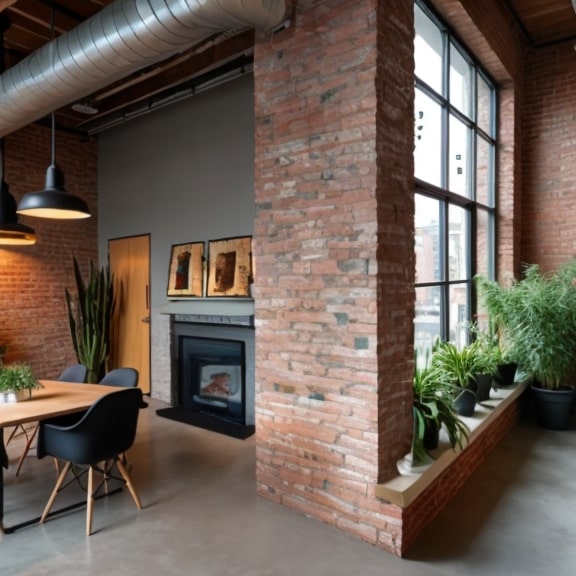} &
\includegraphics[width=0.12\linewidth]{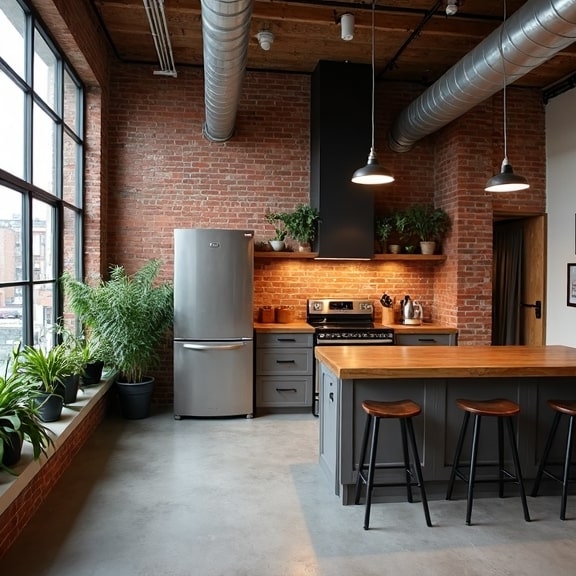} &
\includegraphics[width=0.12\linewidth]{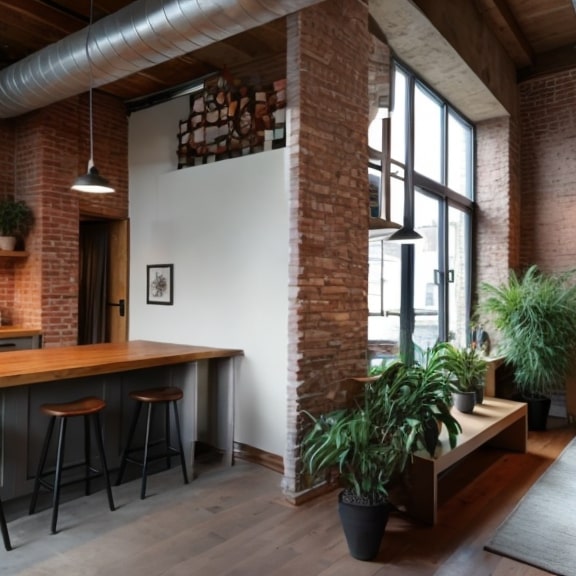} &
\includegraphics[width=0.12\linewidth]{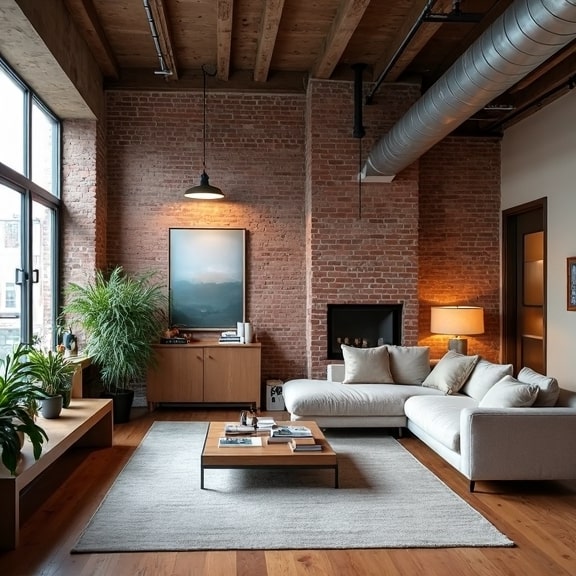} &
\includegraphics[width=0.12\linewidth]{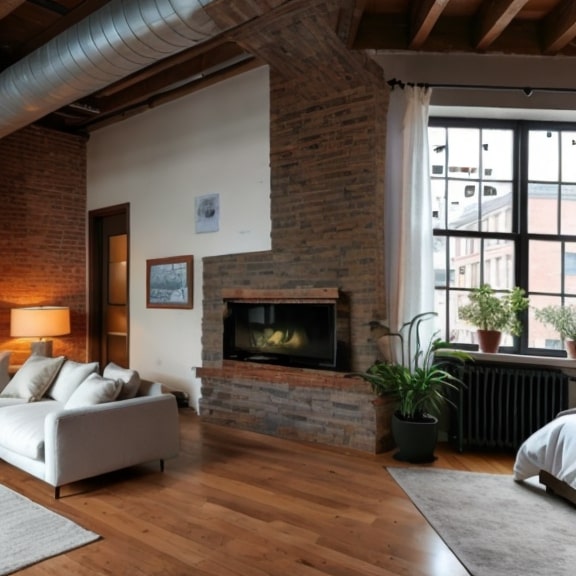} &
\includegraphics[width=0.12\linewidth]{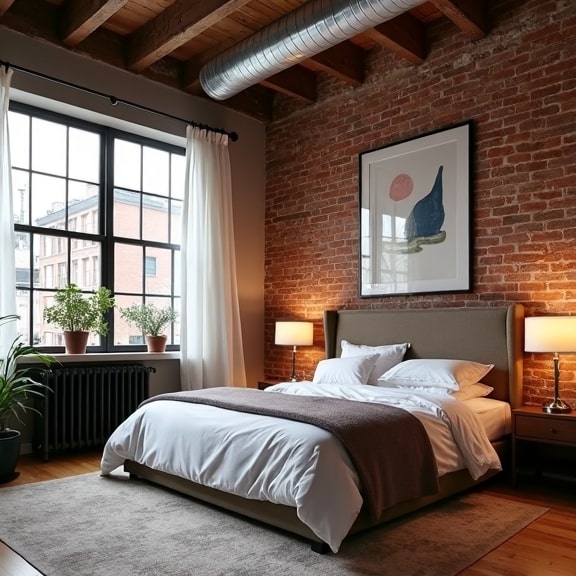} &
\includegraphics[width=0.12\linewidth]{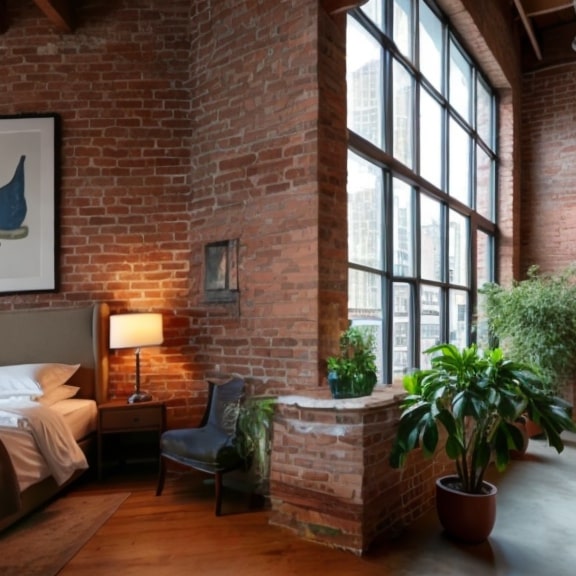} \\
\multicolumn{8}{c}{The 8 initial panoramic images corresponding to the prefixes: dining room $\rightarrow$ kitchen $\rightarrow$ living room $\rightarrow$ bedroom} \\
\end{tabular}

\begin{tabular}{cccccccc}
\includegraphics[width=0.12\linewidth]{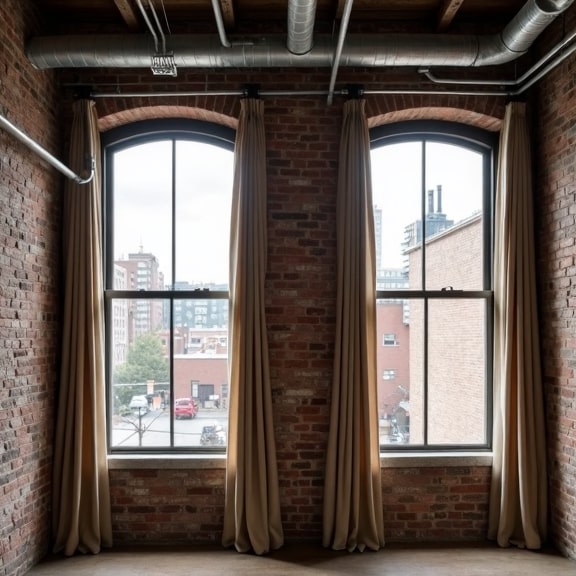} &
\includegraphics[width=0.12\linewidth]{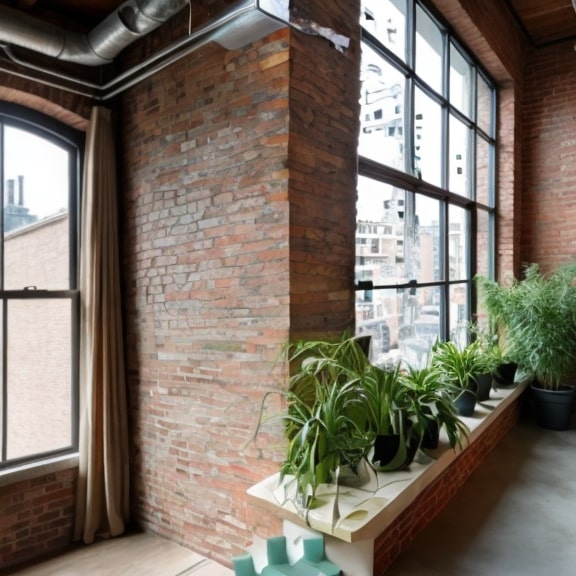} &
\includegraphics[width=0.12\linewidth]{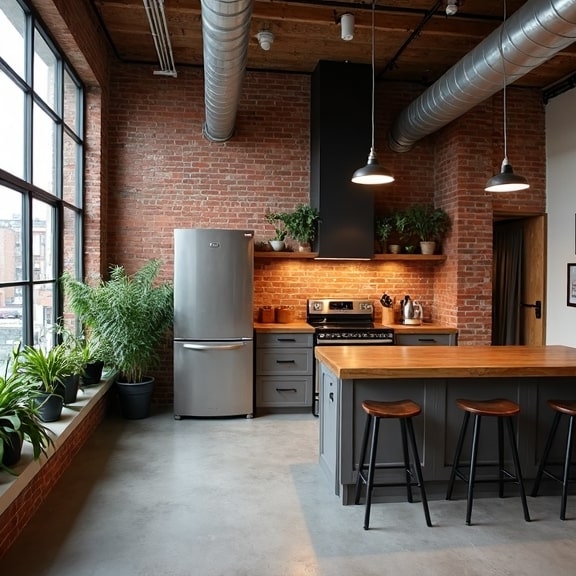} &
\includegraphics[width=0.12\linewidth]{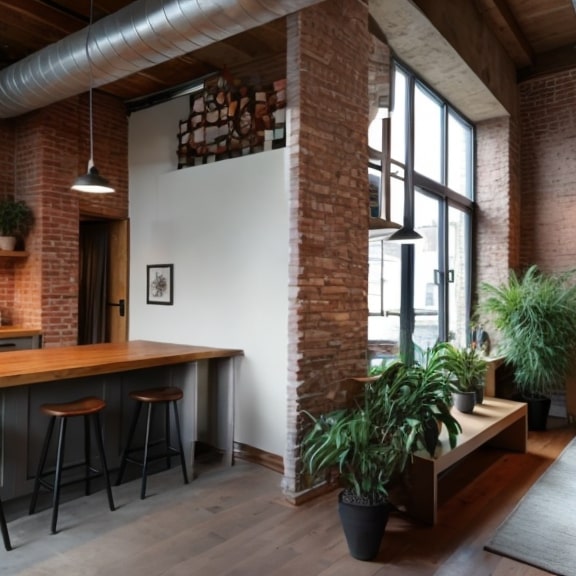} &
\includegraphics[width=0.12\linewidth]{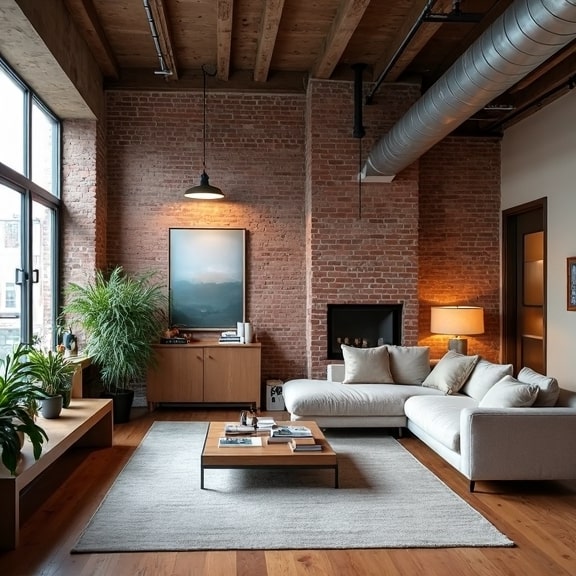} &
\includegraphics[width=0.12\linewidth]{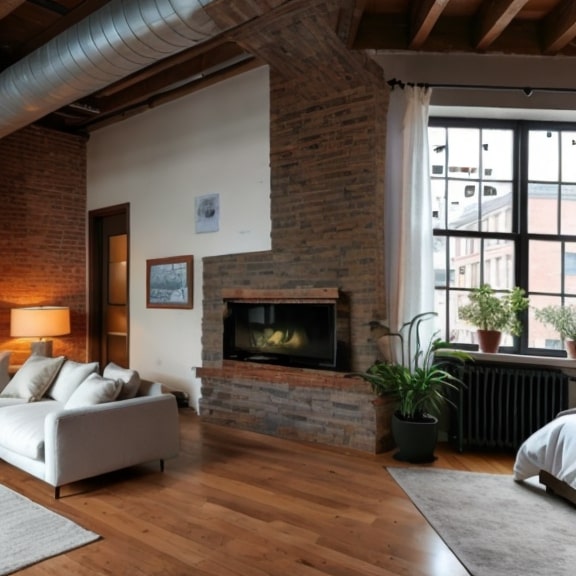} &
\includegraphics[width=0.12\linewidth]{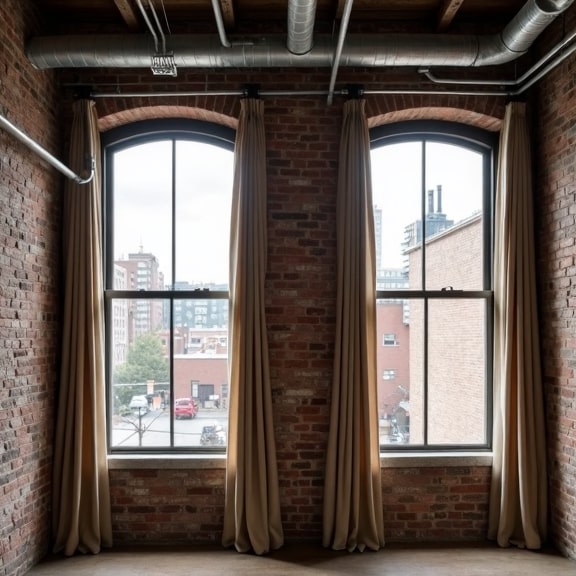} &
\includegraphics[width=0.12\linewidth]{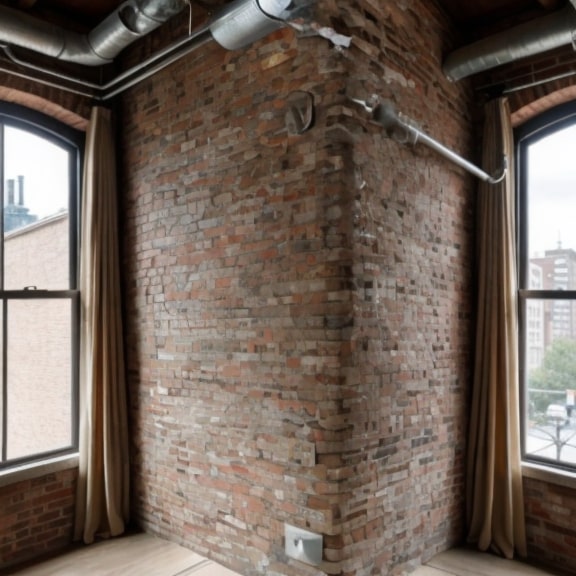} \\
\multicolumn{8}{c}{The 8 initial panoramic images corresponding to the prefixes: wall $\rightarrow$ kitchen $\rightarrow$ living room $\rightarrow$ wall} \\
\end{tabular}

\begin{tabular}{ccc}
\includegraphics[width=0.32\linewidth]{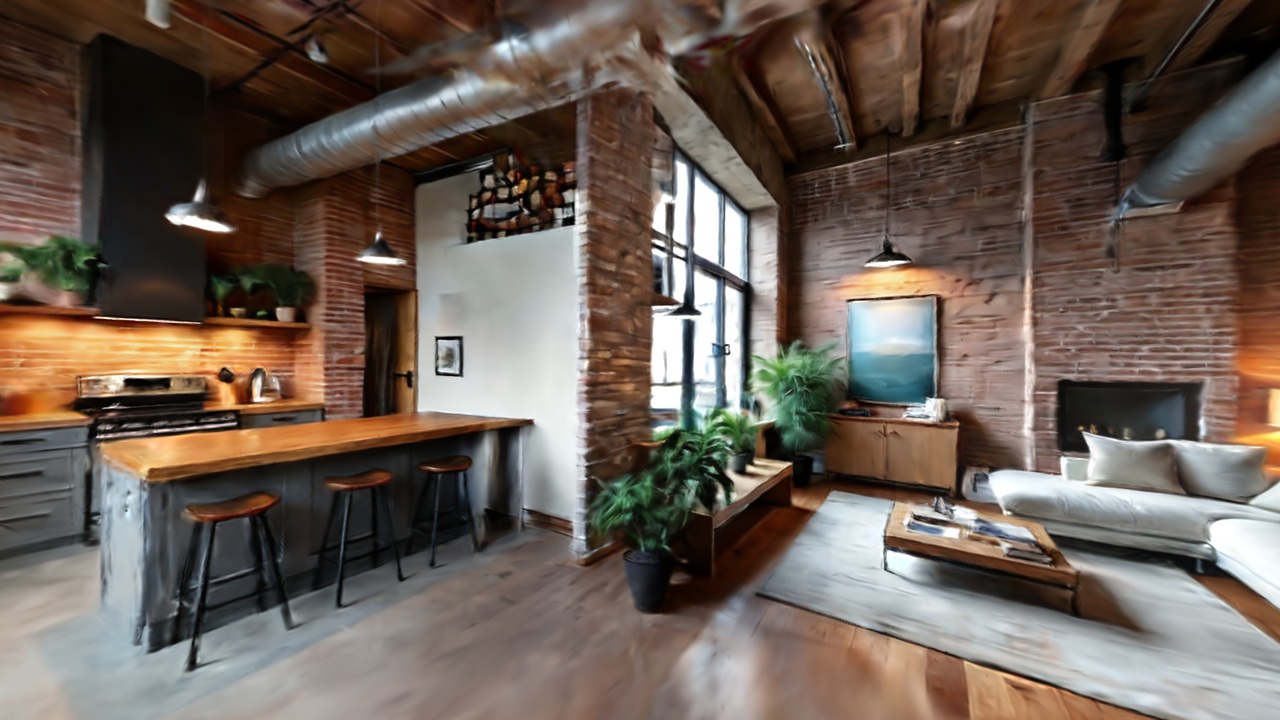} &
\includegraphics[width=0.32\linewidth]{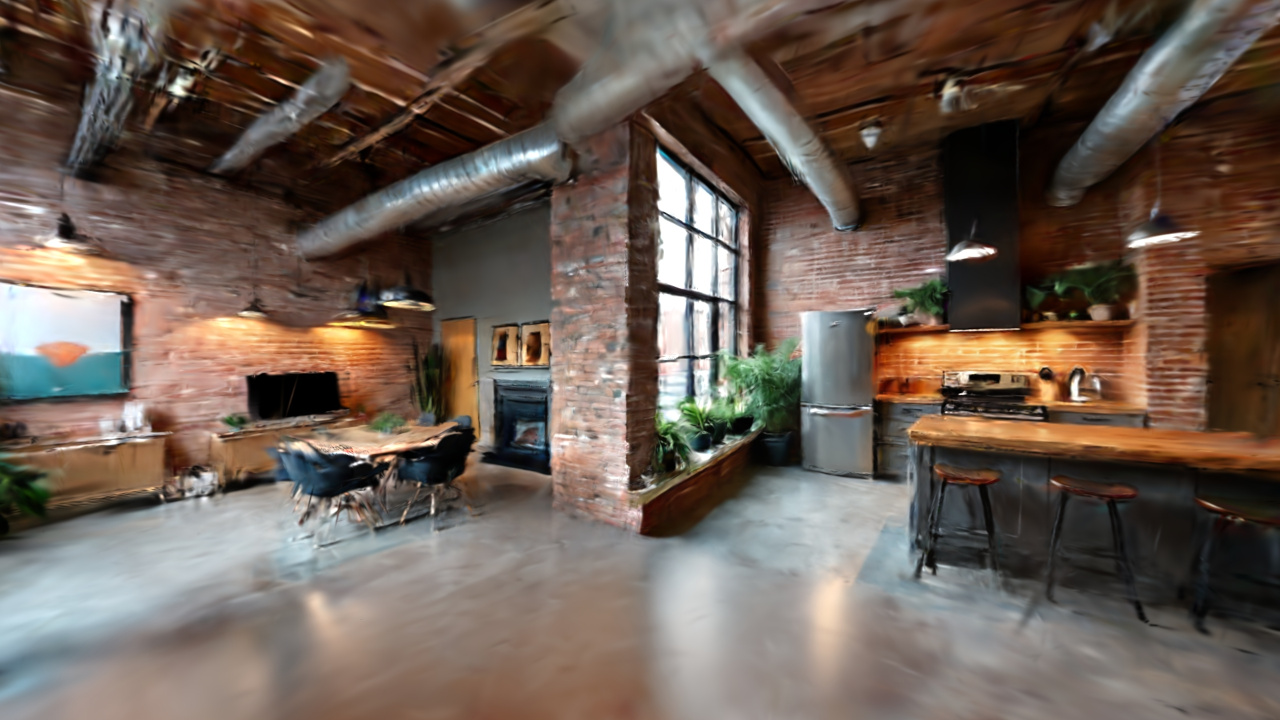} &
\includegraphics[width=0.32\linewidth]{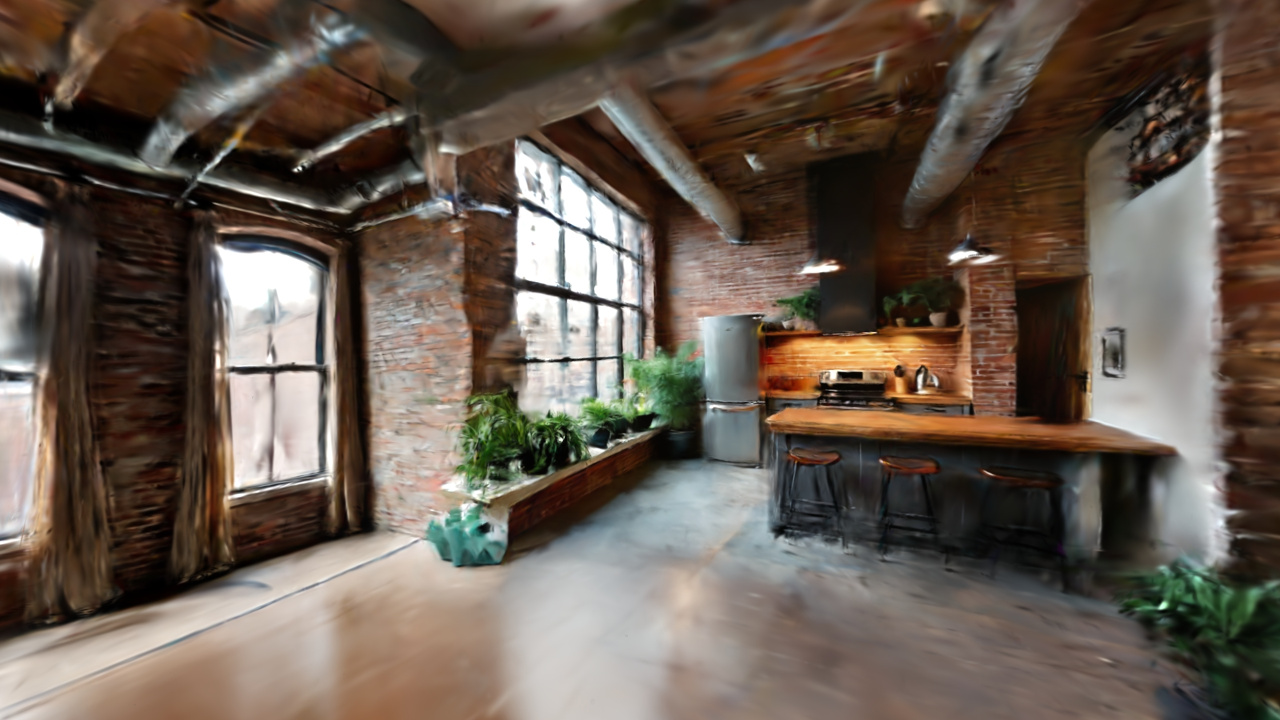} \\
Scene rendering: kitchen $\rightarrow$ living room &
Scene rendering: dining room $\rightarrow$ kitchen &
Scene rendering: wall $\rightarrow$ kitchen \\
\end{tabular}

\caption{
\textbf{Scene layout diversity.} 
We generate an \textit{"urban industrial loft"} with two different layouts by modifying the text prompts in the first stage.
Top: generated panoramic scene scaffold with the room prefixes ``dining room, kitchen, living room, bedroom''.
Mid: replacing ``dining room'' and ``bedroom'' prefixes with ``wall''.
Bottom left: in both cases, we obtain scenes that transition from a kitchen to a living room.
Bottom mid: in the first case (top) the generated scene contains another transition from a dining room area to the kitchen.
Bottom right: in the second case (mid) we instead create a smaller scene without another room, but instead a flat wall transitioning to the kitchen.
}
\Description{Scene layout diversity. We generate an \textit{"urban industrial loft"} with two different layouts by modifying the text prompts in the first stage.
Top: generated panoramic scene scaffold with the room prefixes ``dining room, kitchen, living room, bedroom''.
Mid: replacing ``dining room'' and ``bedroom'' prefixes with ``wall''.
Bottom left: in both cases, we obtain scenes that transition from a kitchen to a living room.
Bottom mid: in the first case (top) the generated scene contains another transition from a dining room area to the kitchen.
Bottom right: in the second case (mid) we instead create a smaller scene without another room, but instead a flat wall transitioning to the kitchen.}
\label{fig:l-shaped}
\end{figure*}

%% file: tables/fig_ours_scenes_suppl_1.tex
\begin{figure*}
\centering
\setlength\tabcolsep{1pt}         
\renewcommand{\arraystretch}{1}   

\begin{tabular}{cccccc}

& Generated scene overview & \multicolumn{3}{c}{Rendered novel views} \\

\multirow{2}{*}[0.04\textwidth]{\rotatebox{90}{\textit{"lush desert oasis"}}} &
\includegraphics[height=0.15\textwidth]{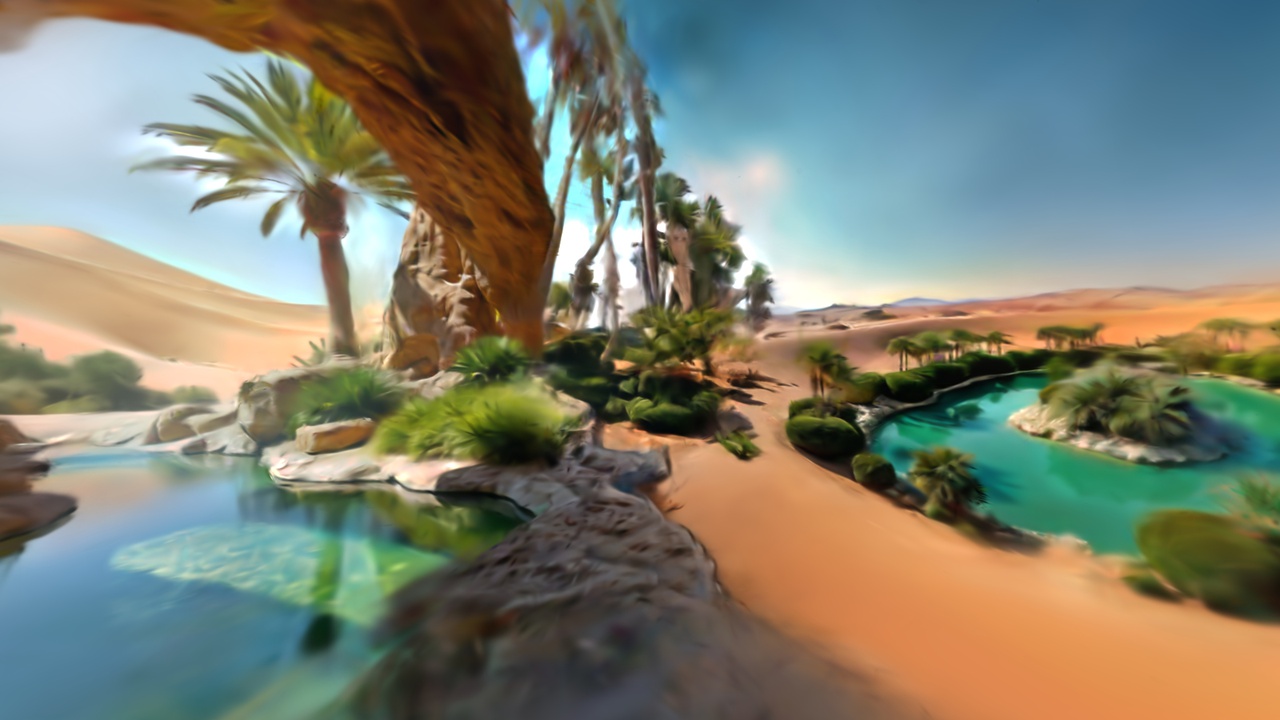} &
\includegraphics[height=0.15\textwidth]{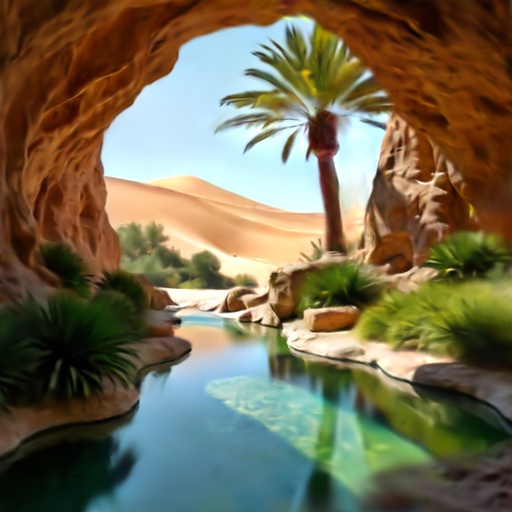} &
\includegraphics[height=0.15\textwidth]{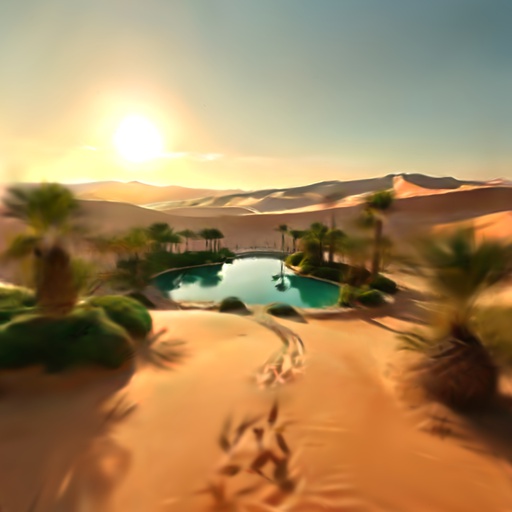} &
\includegraphics[height=0.15\textwidth]{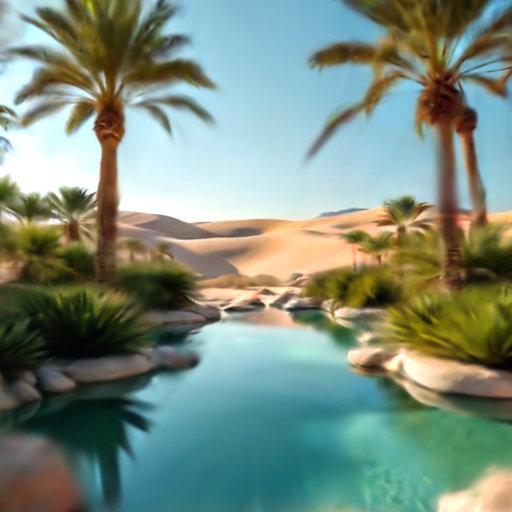} &
\includegraphics[height=0.15\textwidth]{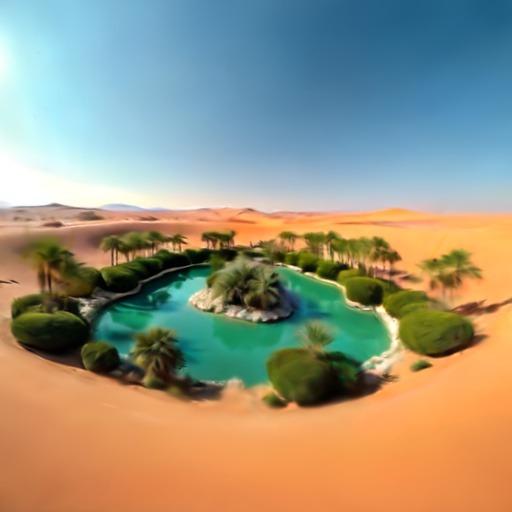} \\

&  %
\includegraphics[height=0.15\textwidth]{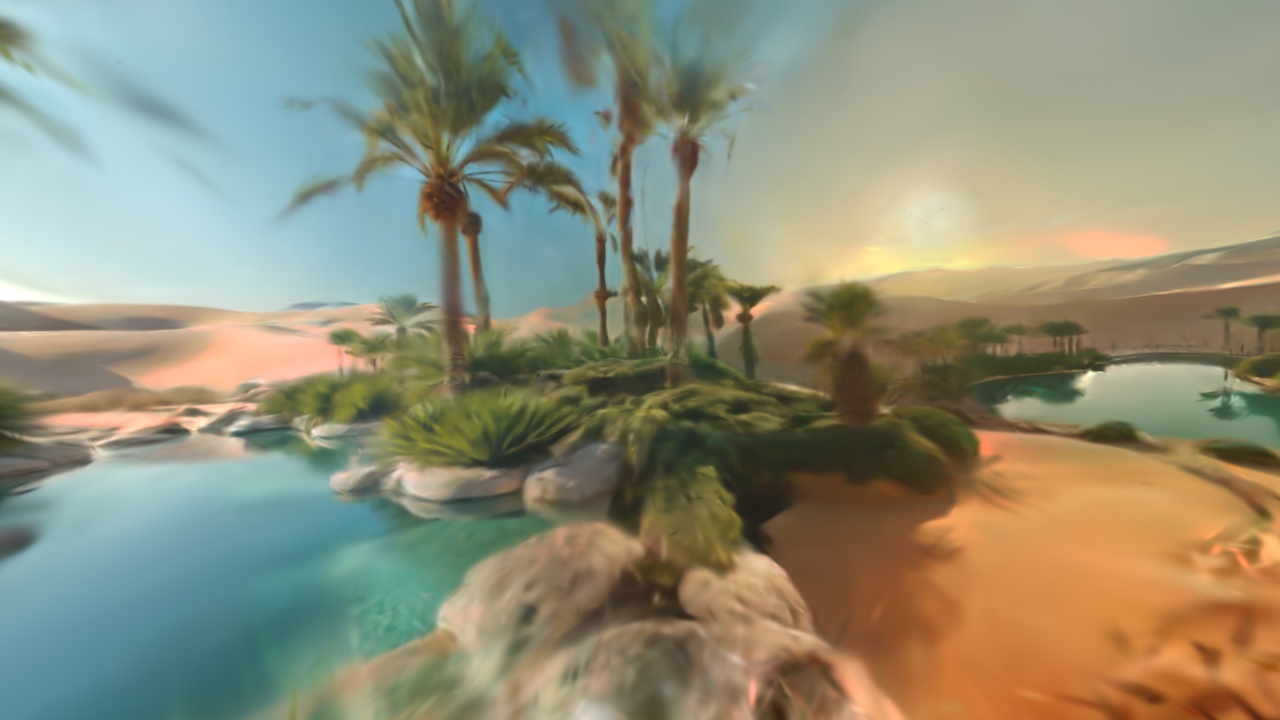} &
\includegraphics[height=0.15\textwidth]{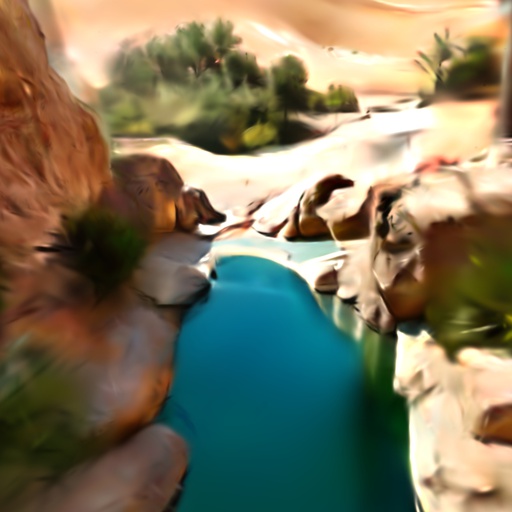} &
\includegraphics[height=0.15\textwidth]{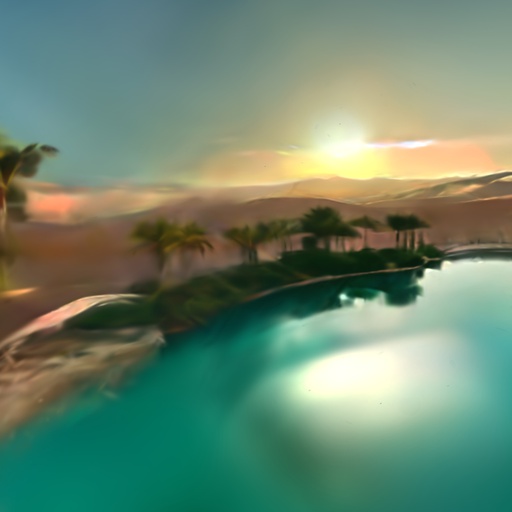} &
\includegraphics[height=0.15\textwidth]{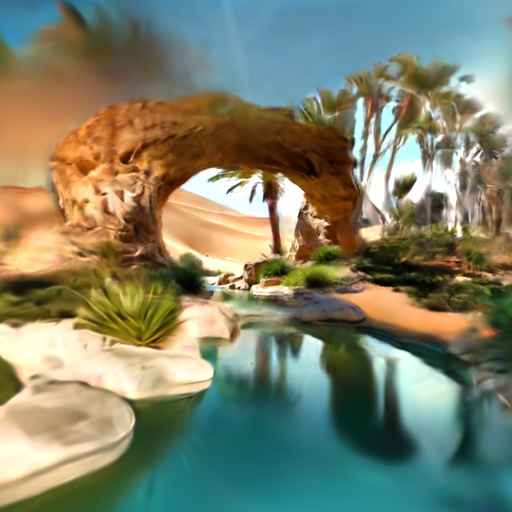} &
\includegraphics[height=0.15\textwidth]{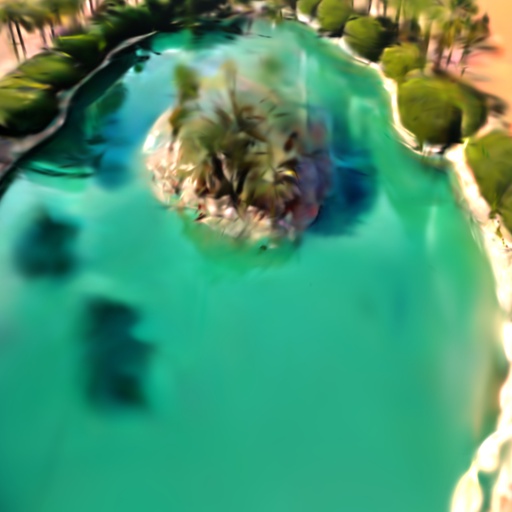} \\

\multirow{2}{*}[0.08\textwidth]{\rotatebox{90}{\textit{"elegant French country estate"}}} &
\includegraphics[height=0.15\textwidth]{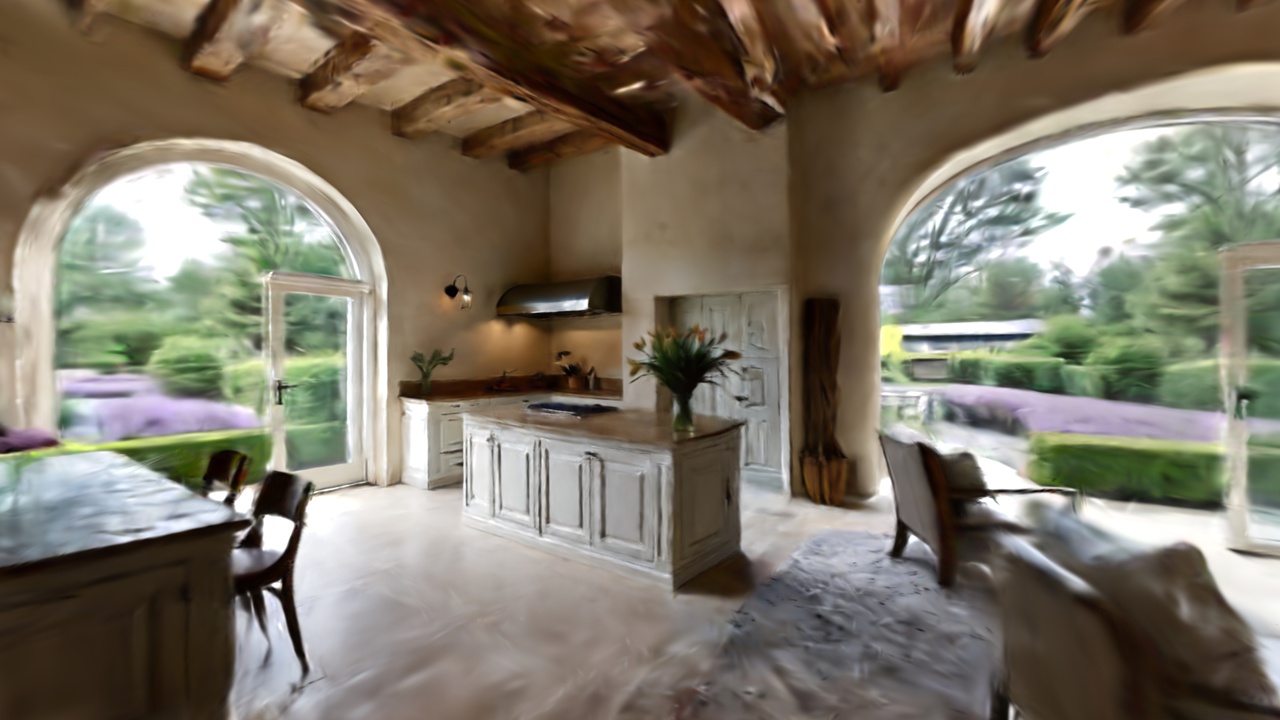} &
\includegraphics[height=0.15\textwidth]{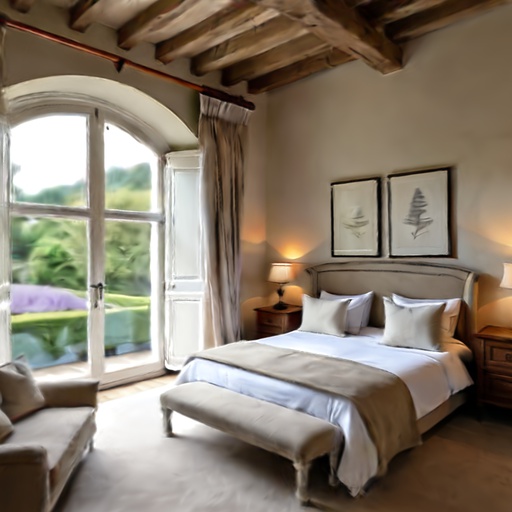} &
\includegraphics[height=0.15\textwidth]{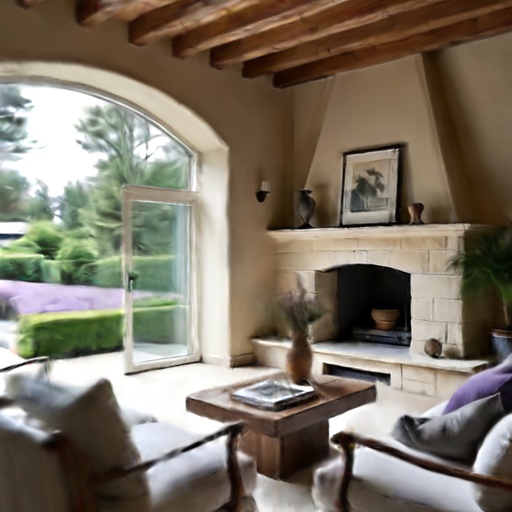} &
\includegraphics[height=0.15\textwidth]{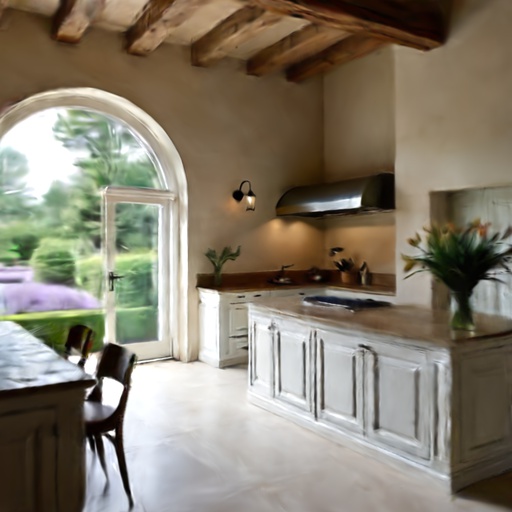} &
\includegraphics[height=0.15\textwidth]{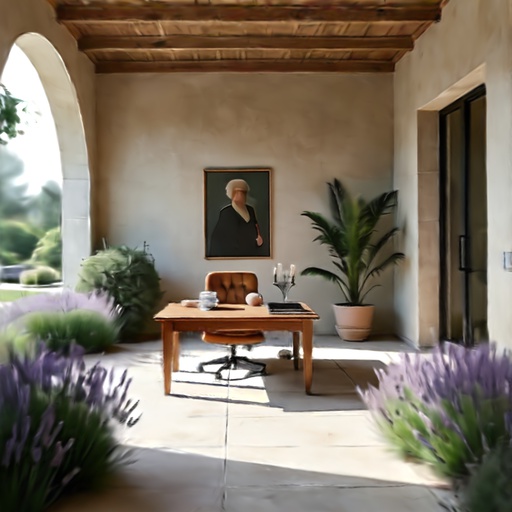} \\

&  %
\includegraphics[height=0.15\textwidth]{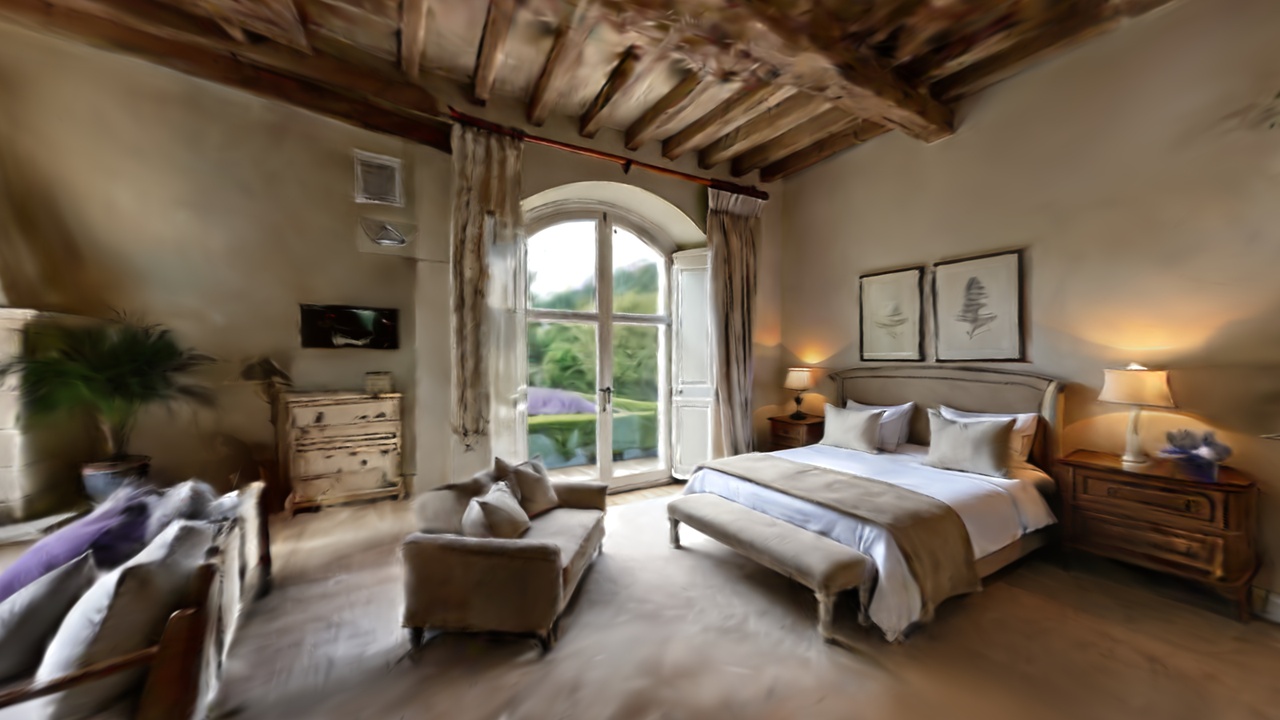} &
\includegraphics[height=0.15\textwidth]{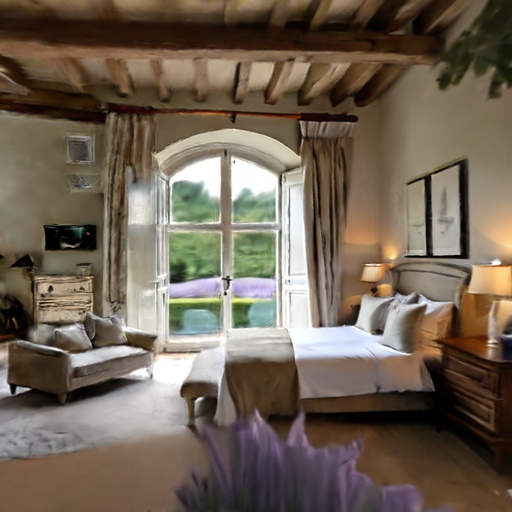} &
\includegraphics[height=0.15\textwidth]{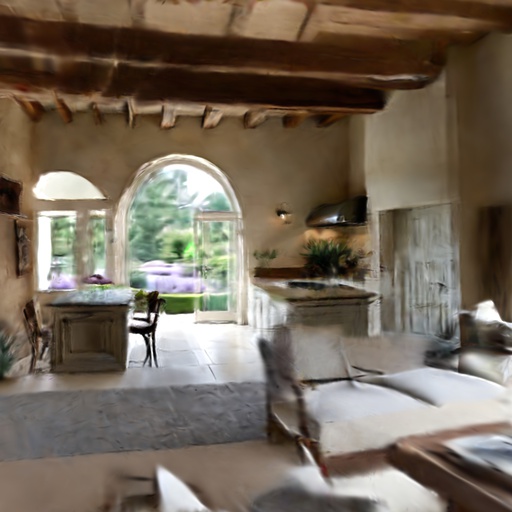} &
\includegraphics[height=0.15\textwidth]{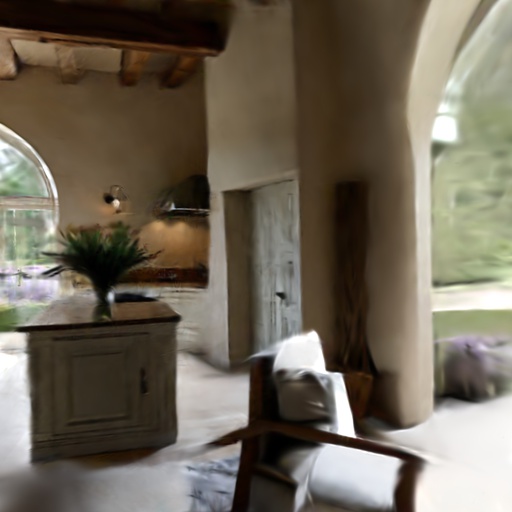} &
\includegraphics[height=0.15\textwidth]{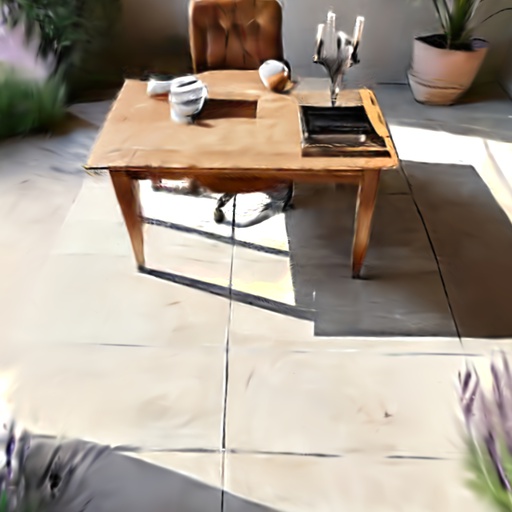} \\

\multirow{2}{*}[0.07\textwidth]{\rotatebox{90}{\textit{"bubble gum coral reef"}}} &
\includegraphics[height=0.15\textwidth]{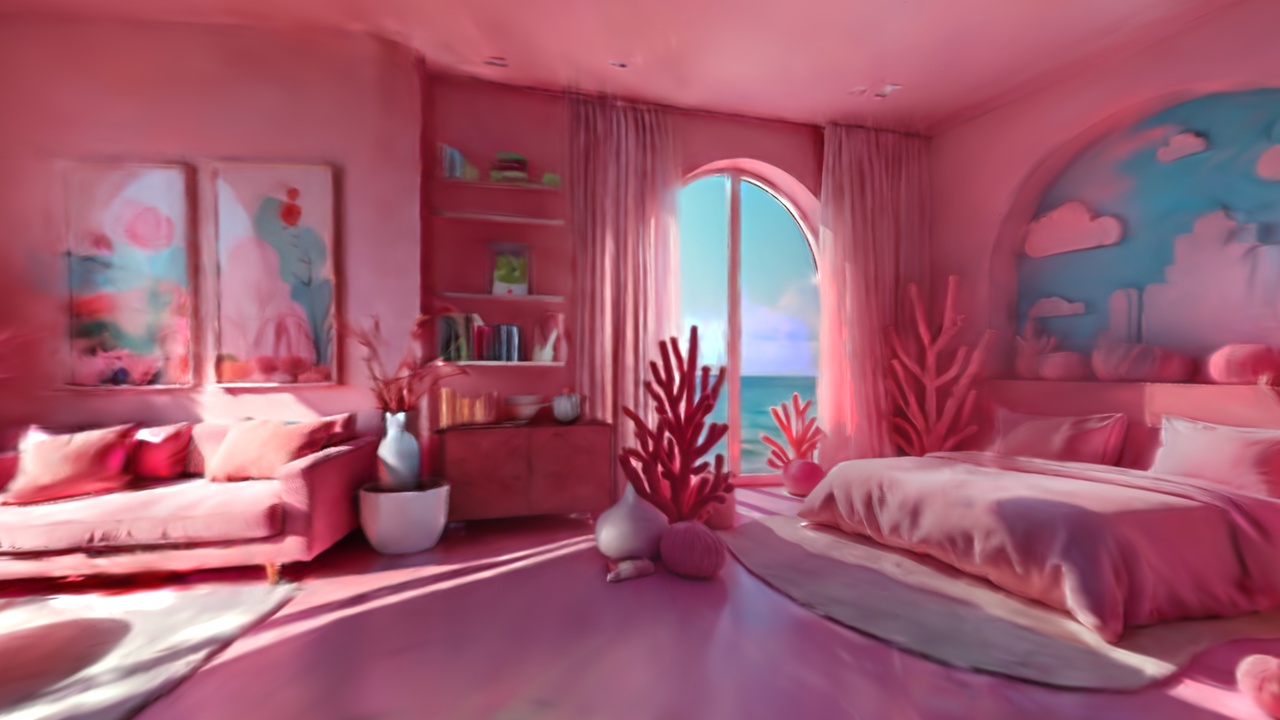} &
\includegraphics[height=0.15\textwidth]{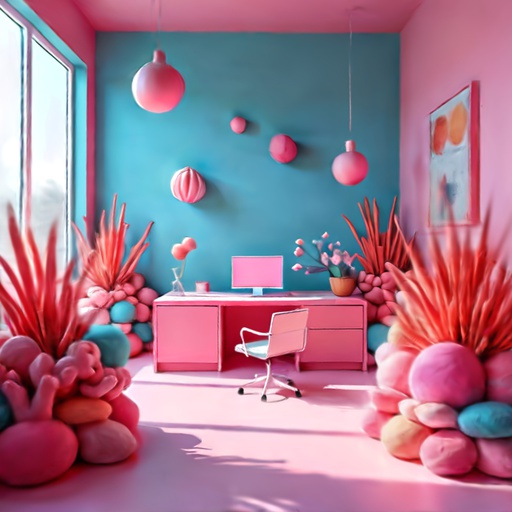} &
\includegraphics[height=0.15\textwidth]{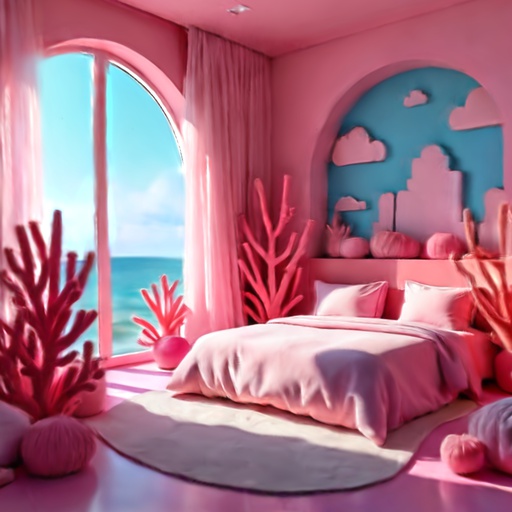} &
\includegraphics[height=0.15\textwidth]{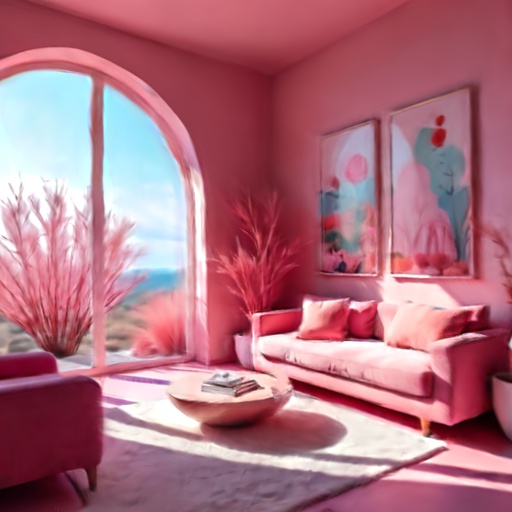} &
\includegraphics[height=0.15\textwidth]{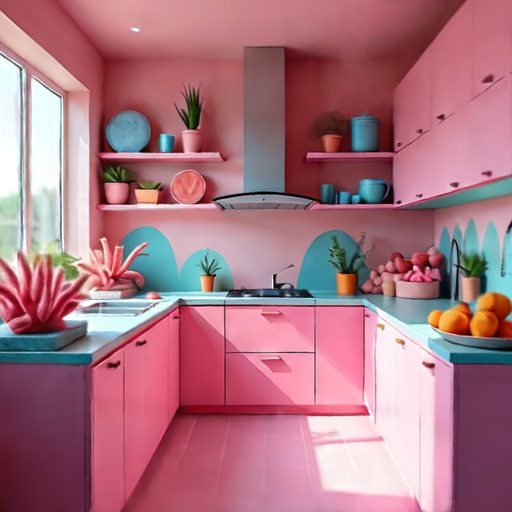} \\

&  %
\includegraphics[height=0.15\textwidth]{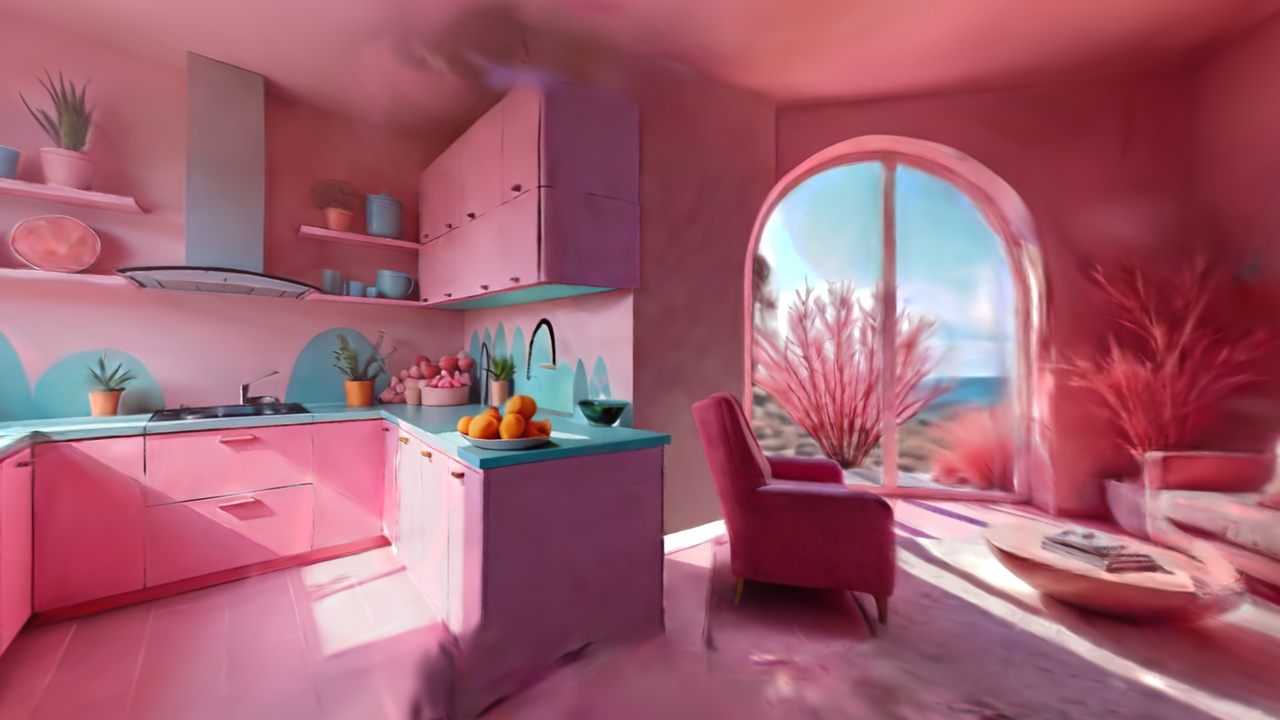} &
\includegraphics[height=0.15\textwidth]{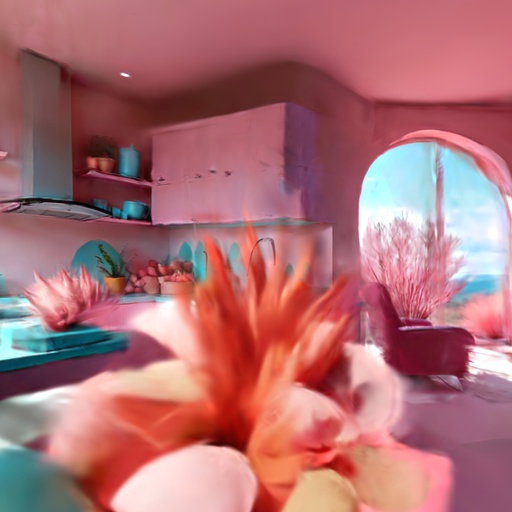} &
\includegraphics[height=0.15\textwidth]{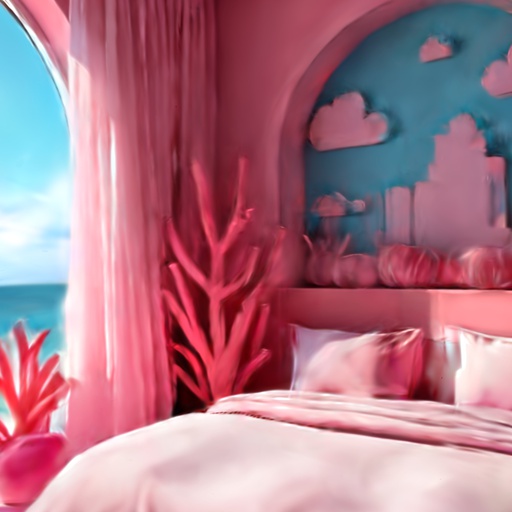} &
\includegraphics[height=0.15\textwidth]{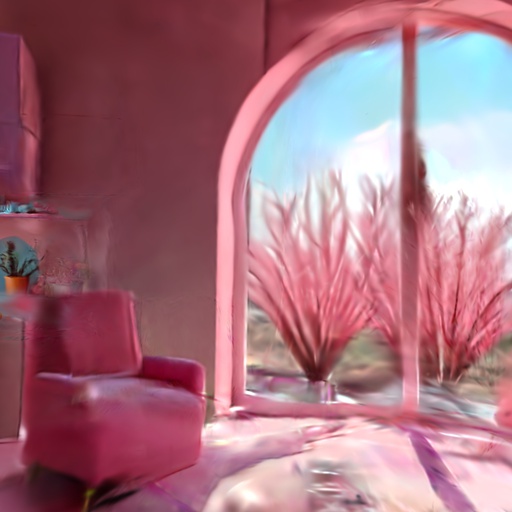} &
\includegraphics[height=0.15\textwidth]{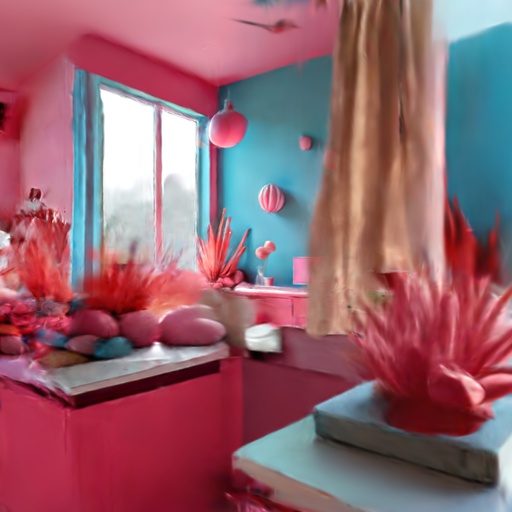} \\
\end{tabular}

\caption{
\textbf{Additional world generation results of our method.}
We visualize scene overview renderings close to the scene center facing in two opposite directions.
Additionally, we render novel views far beyond the scene center, that explore the different generated areas in more detail.
This showcases the possibility to interactively explore our worlds from arbitrary viewpoints.
Please see the supplementary video for animated flythroughs of our generated scenes.
}
\Description{
\textbf{Additional world generation results of our method.}
We visualize scene overview renderings close to the scene center facing in two opposite directions.
Additionally, we render novel views far beyond the scene center, that explore the different generated areas in more detail.
This showcases the possibility to interactively explore our worlds from arbitrary viewpoints.
Please see the supplementary video for animated flythroughs of our generated scenes.
}
\label{fig:ours-scenes-suppl-1}
\end{figure*}

%% file: tables/fig_ours_scenes_suppl_2.tex
\begin{figure*}
\centering
\setlength\tabcolsep{1pt}         
\renewcommand{\arraystretch}{1}   

\begin{tabular}{cccccc}

& Generated scene overview & \multicolumn{3}{c}{Rendered novel views} \\

\multirow{2}{*}[0.07\textwidth]{\rotatebox{90}{\textit{"celestial ink library"}}} &
\includegraphics[height=0.15\textwidth]{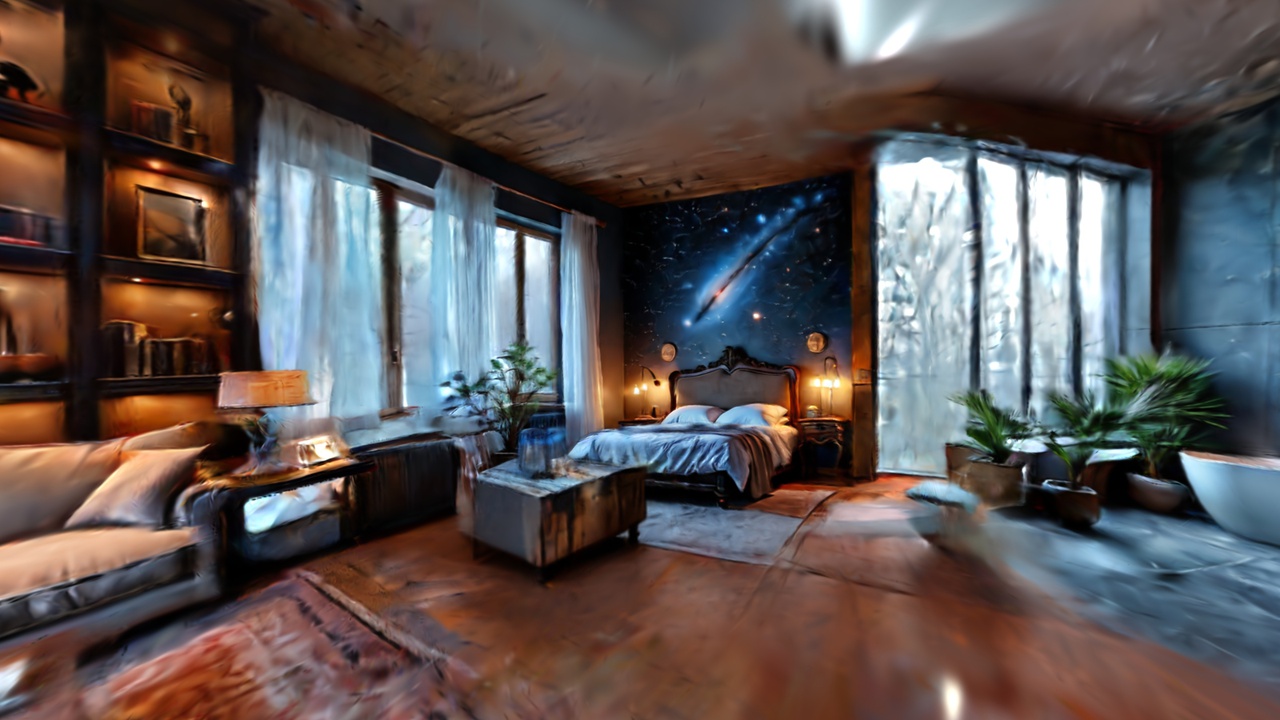} &
\includegraphics[height=0.15\textwidth]{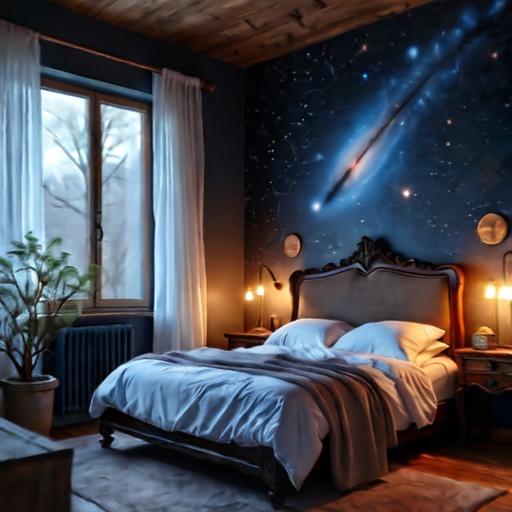} &
\includegraphics[height=0.15\textwidth]{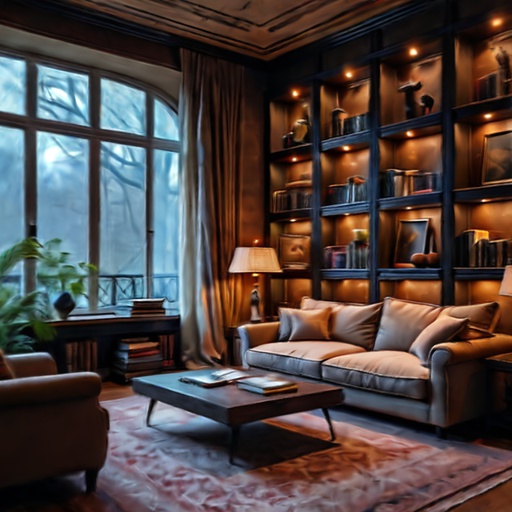} &
\includegraphics[height=0.15\textwidth]{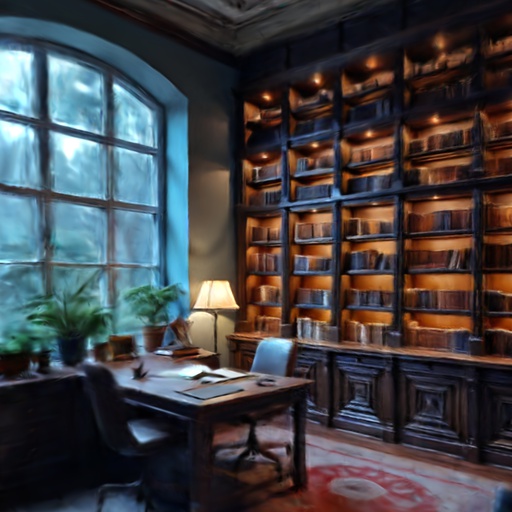} &
\includegraphics[height=0.15\textwidth]{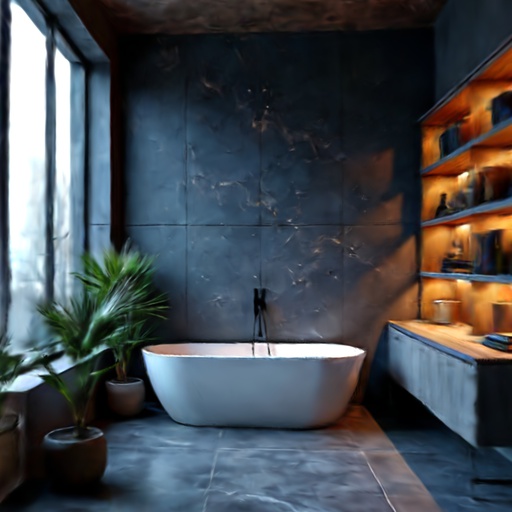} \\

&  %
\includegraphics[height=0.15\textwidth]{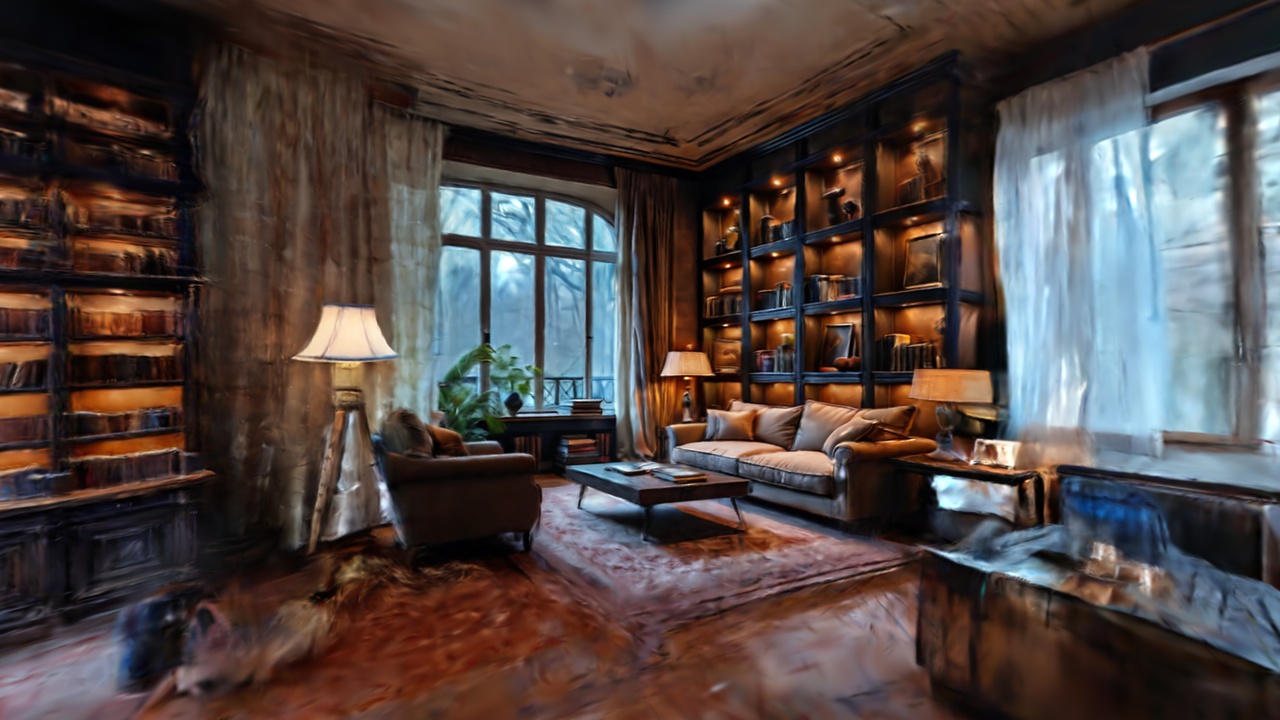} &
\includegraphics[height=0.15\textwidth]{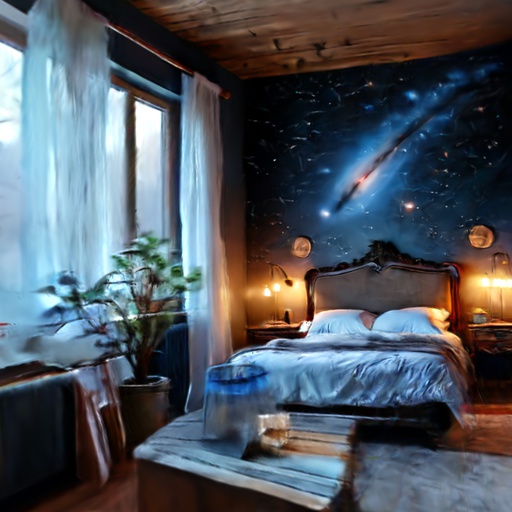} &
\includegraphics[height=0.15\textwidth]{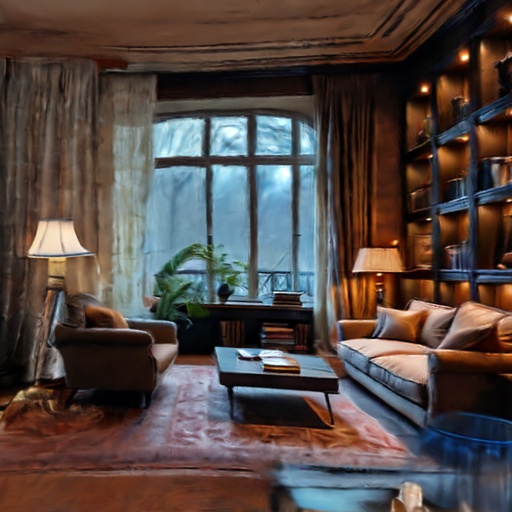} &
\includegraphics[height=0.15\textwidth]{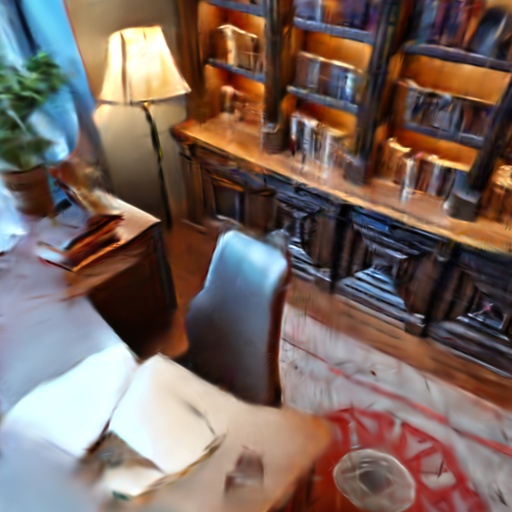} &
\includegraphics[height=0.15\textwidth]{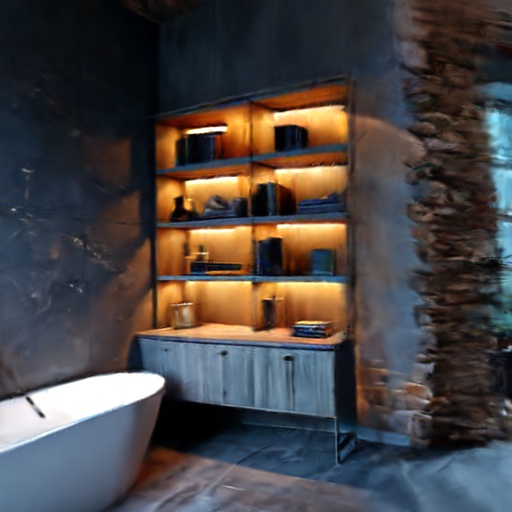} \\

\multirow{2}{*}[0.07\textwidth]{\rotatebox{90}{\textit{"whimsical chocolate factory"}}} &
\includegraphics[height=0.15\textwidth]{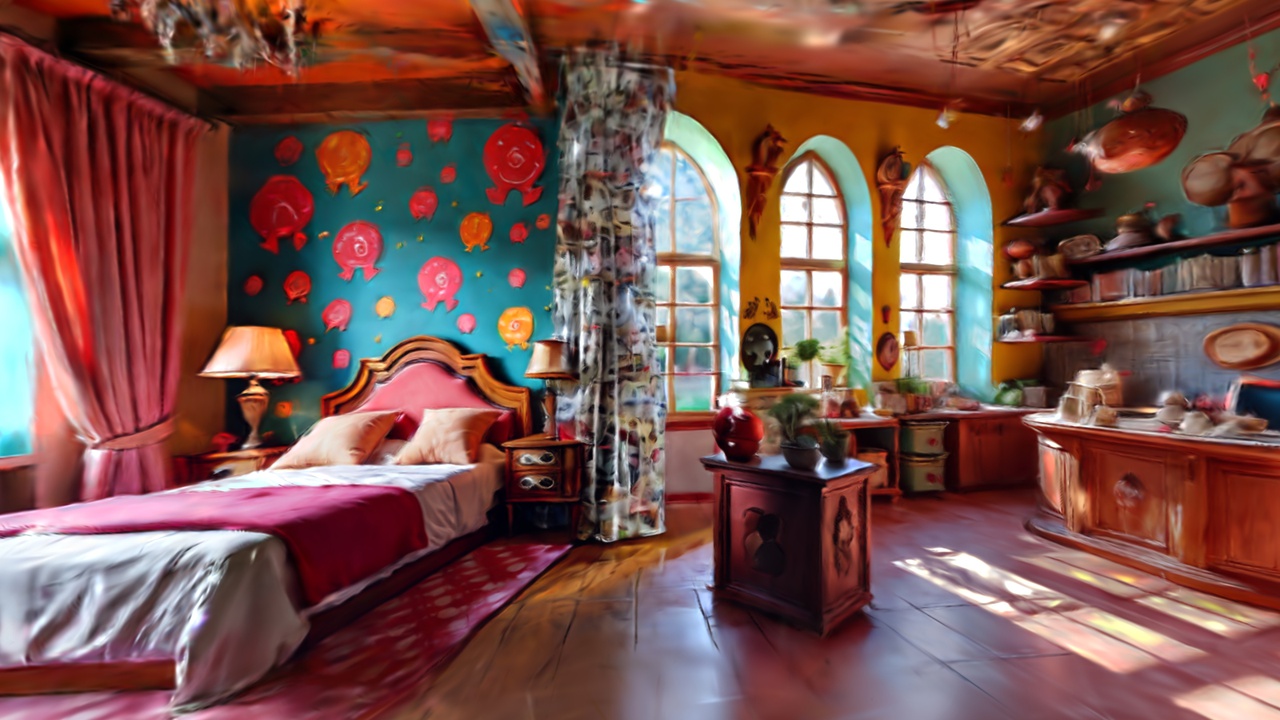} &
\includegraphics[height=0.15\textwidth]{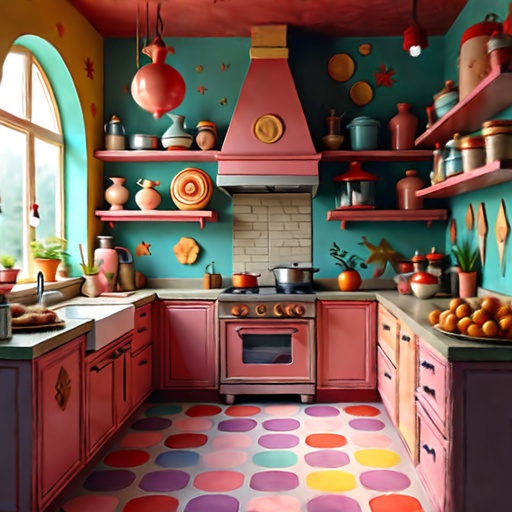} &
\includegraphics[height=0.15\textwidth]{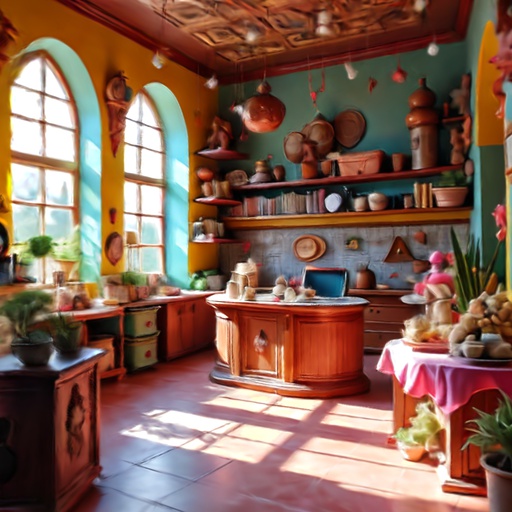} &
\includegraphics[height=0.15\textwidth]{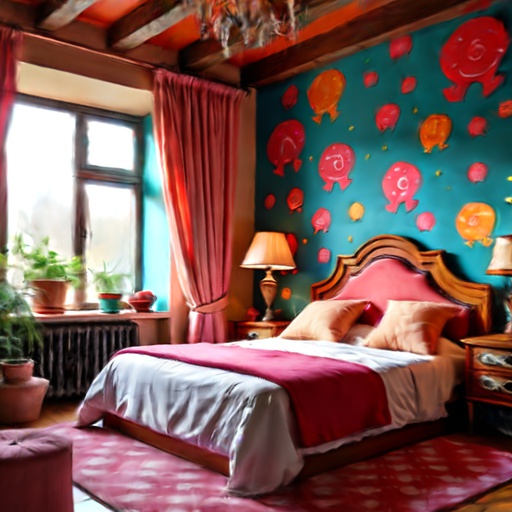} &
\includegraphics[height=0.15\textwidth]{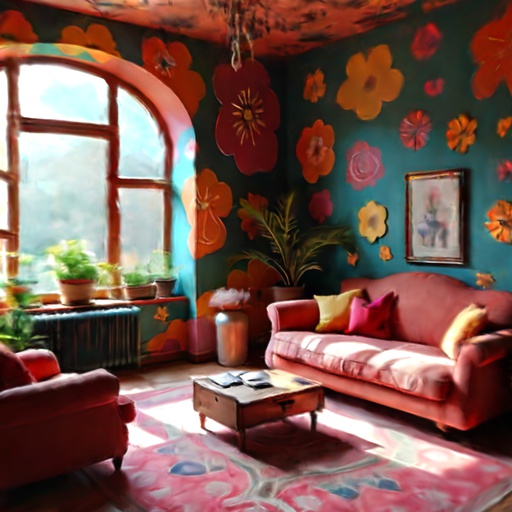} \\

&  %
\includegraphics[height=0.15\textwidth]{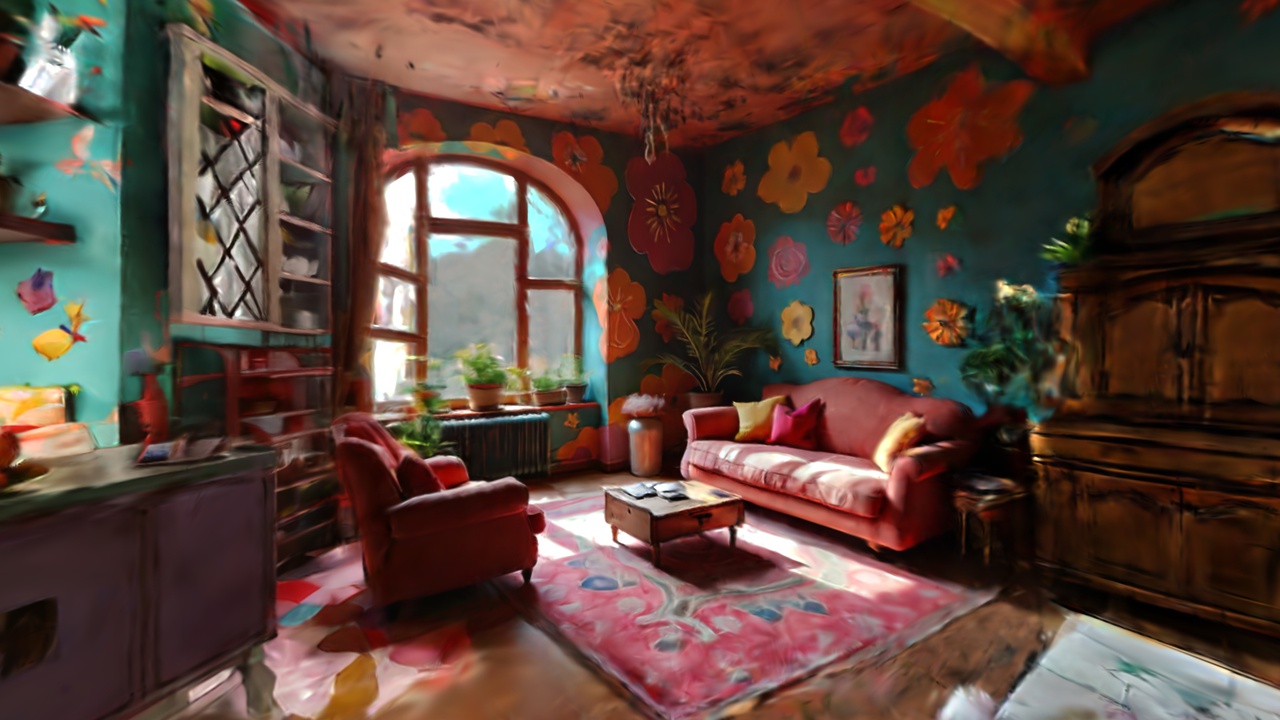} &
\includegraphics[height=0.15\textwidth]{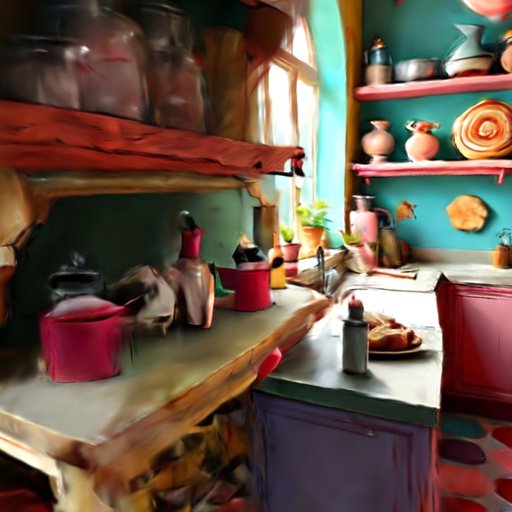} &
\includegraphics[height=0.15\textwidth]{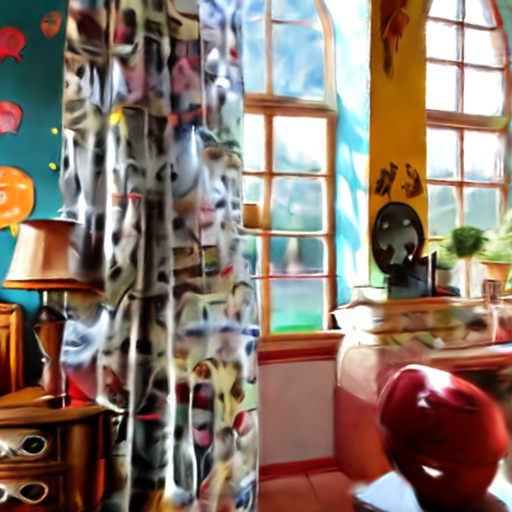} &
\includegraphics[height=0.15\textwidth]{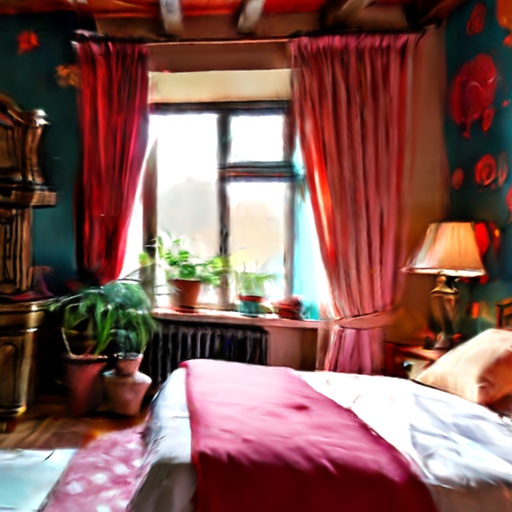} &
\includegraphics[height=0.15\textwidth]{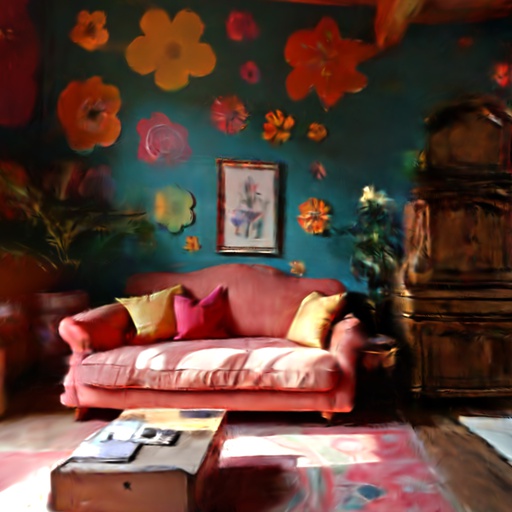} \\

\multirow{2}{*}[0.07\textwidth]{\rotatebox{90}{\textit{"honeycomb observatory"}}} &
\includegraphics[height=0.15\textwidth]{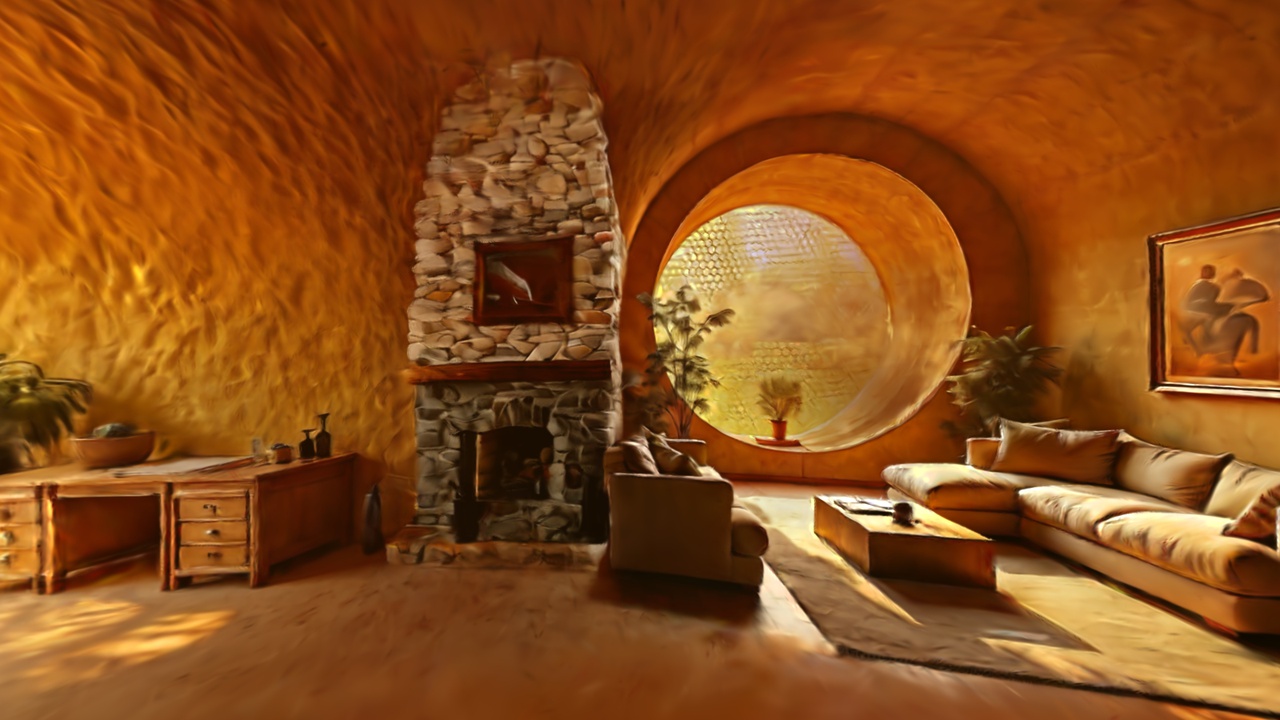} &
\includegraphics[height=0.15\textwidth]{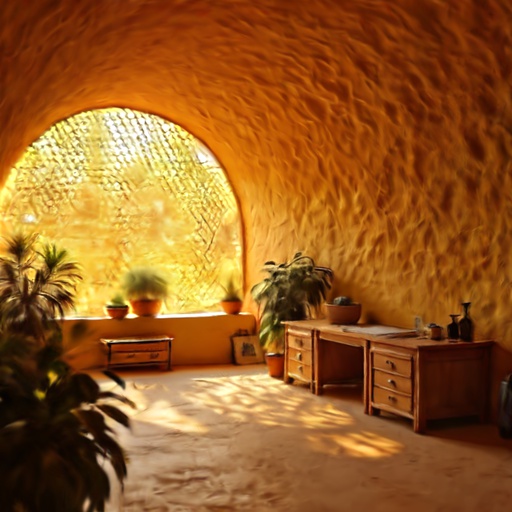} &
\includegraphics[height=0.15\textwidth]{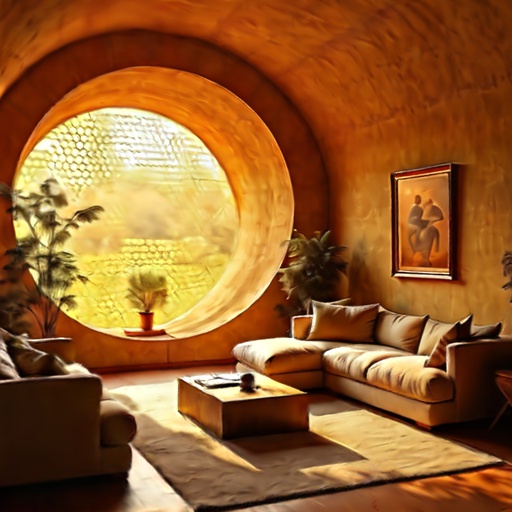} &
\includegraphics[height=0.15\textwidth]{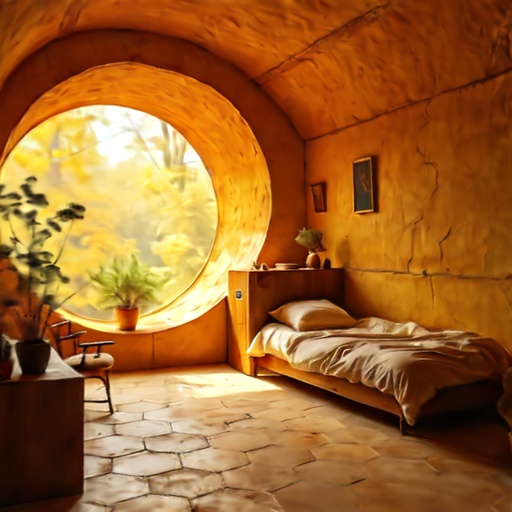} &
\includegraphics[height=0.15\textwidth]{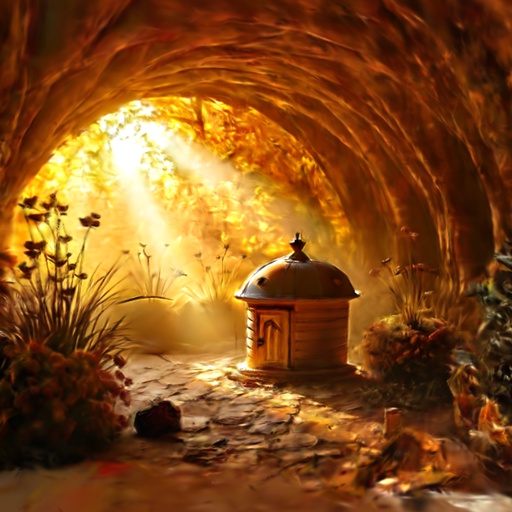} \\

&  %
\includegraphics[height=0.15\textwidth]{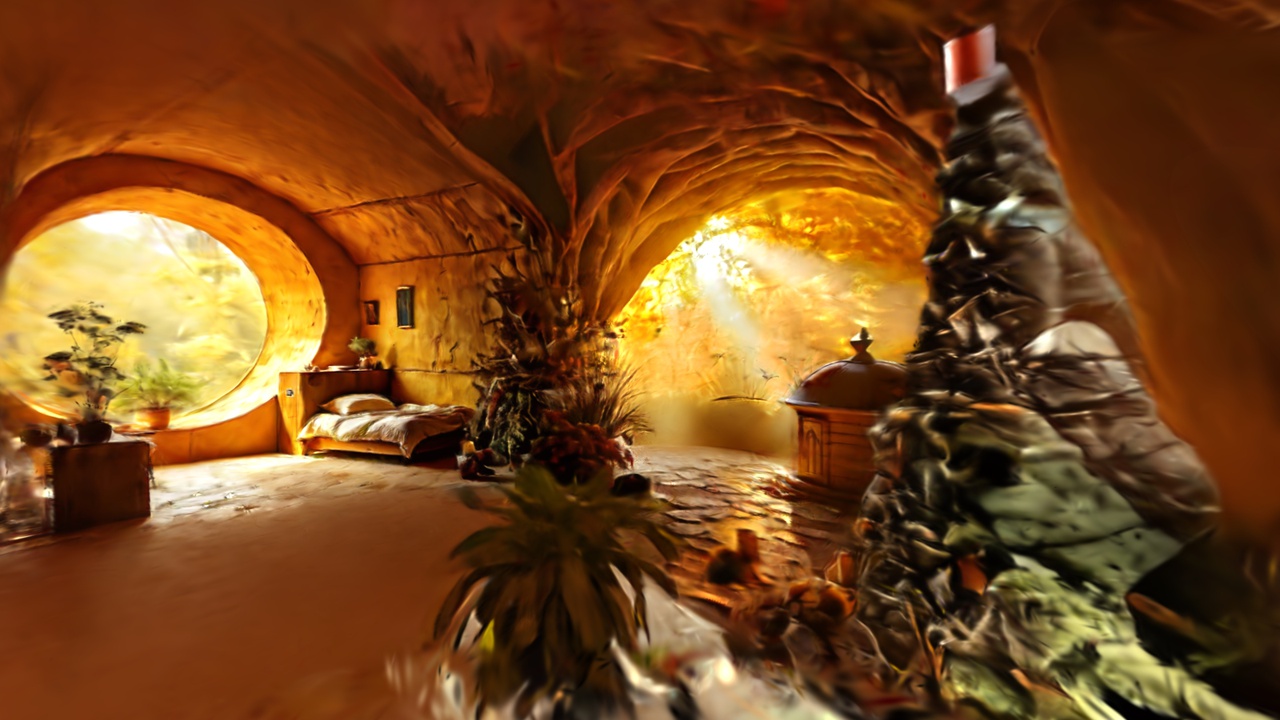} &
\includegraphics[height=0.15\textwidth]{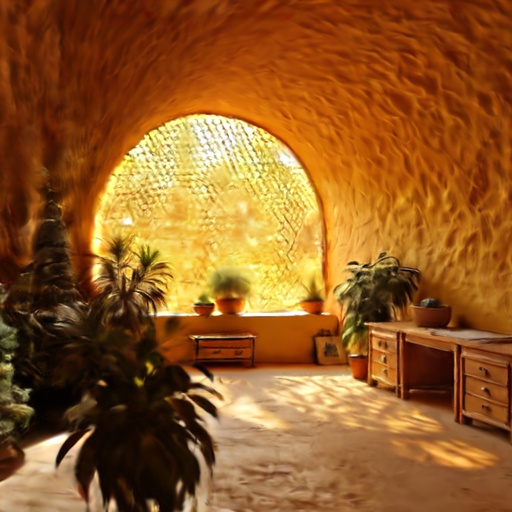} &
\includegraphics[height=0.15\textwidth]{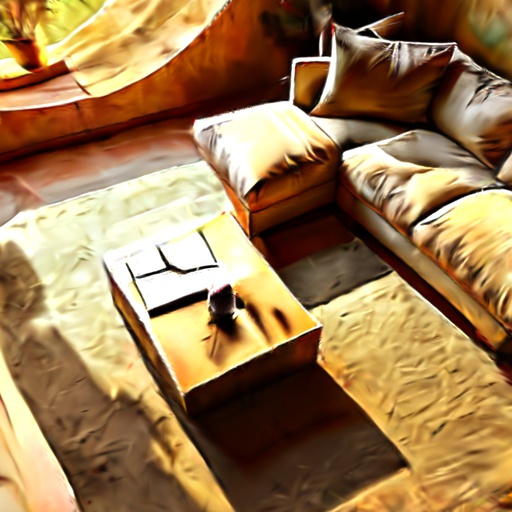} &
\includegraphics[height=0.15\textwidth]{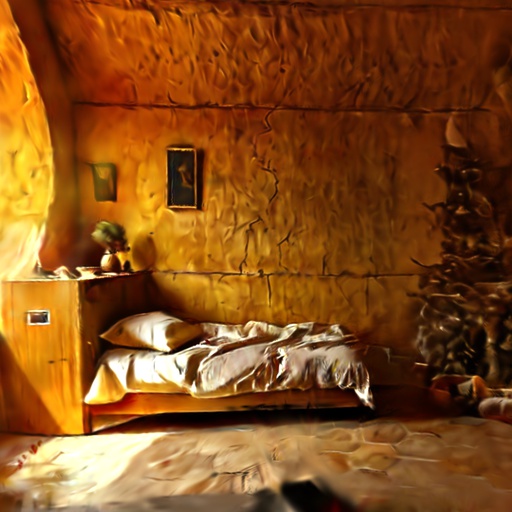} &
\includegraphics[height=0.15\textwidth]{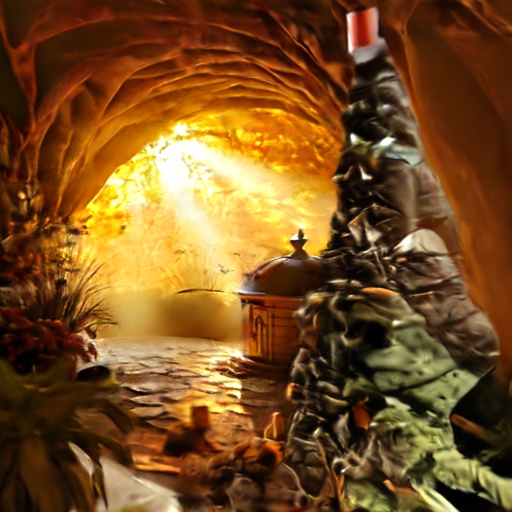} \\
\end{tabular}

\caption{
\textbf{Additional world generation results of our method.}
We visualize scene overview renderings close to the scene center facing in two opposite directions.
Additionally, we render novel views far beyond the scene center, that explore the different generated areas in more detail.
This showcases the possibility to interactively explore our worlds from arbitrary viewpoints.
Please see the supplementary video for animated flythroughs of our generated scenes.
}
\Description{
\textbf{Additional world generation results of our method.}
We visualize scene overview renderings close to the scene center facing in two opposite directions.
Additionally, we render novel views far beyond the scene center, that explore the different generated areas in more detail.
This showcases the possibility to interactively explore our worlds from arbitrary viewpoints.
Please see the supplementary video for animated flythroughs of our generated scenes.
}
\label{fig:ours-scenes-suppl-2}
\end{figure*}

%% file: tables/fig_ours_scenes_suppl_3.tex
\begin{figure*}
\centering
\setlength\tabcolsep{1pt}         
\renewcommand{\arraystretch}{1}   

\begin{tabular}{cccccc}

& Generated scene overview & \multicolumn{3}{c}{Rendered novel views} \\

\multirow{2}{*}[0.07\textwidth]{\rotatebox{90}{\textit{"Urban industrial loft"}}} &
\includegraphics[height=0.15\textwidth]{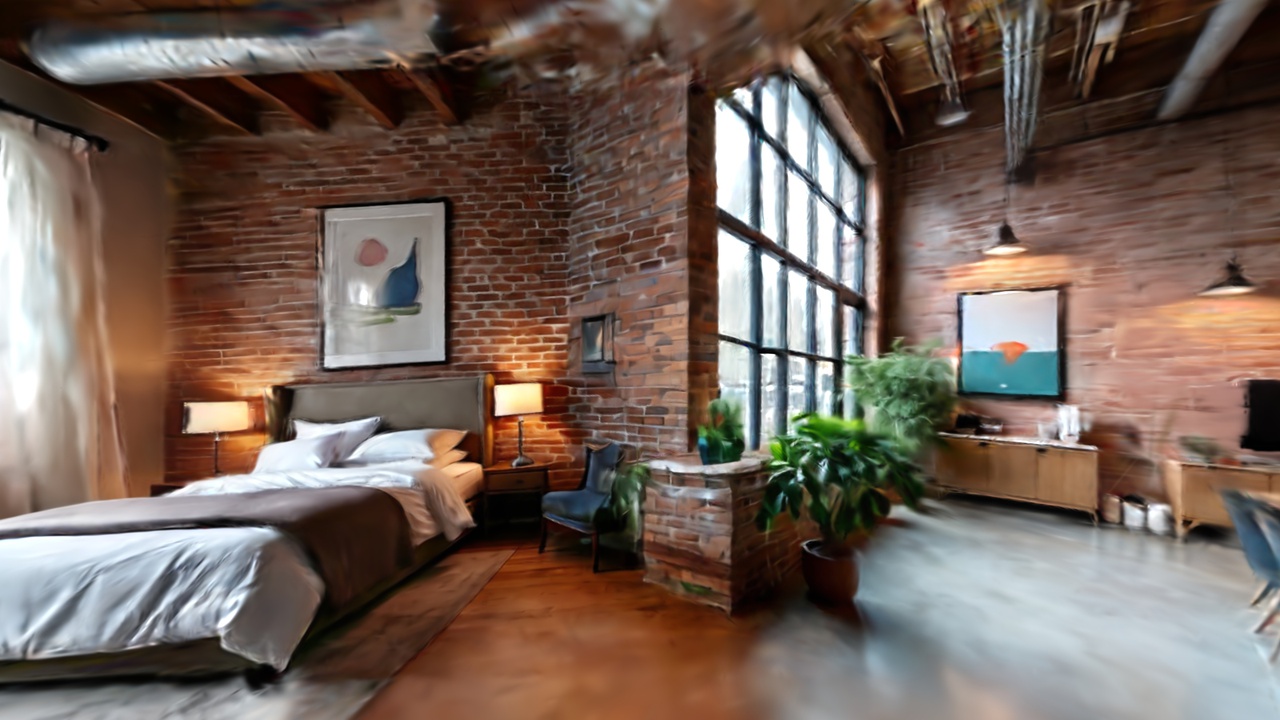} &
\includegraphics[height=0.15\textwidth]{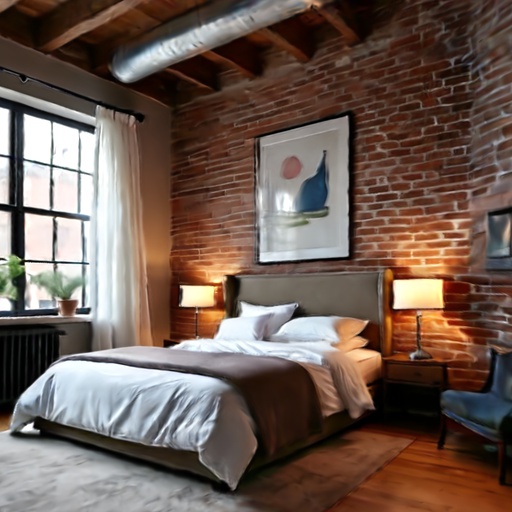} &
\includegraphics[height=0.15\textwidth]{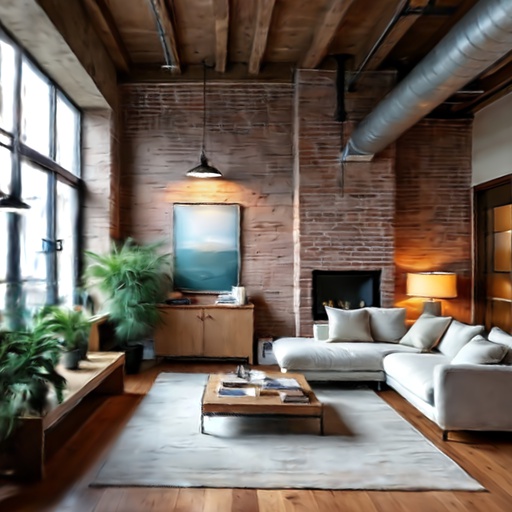} &
\includegraphics[height=0.15\textwidth]{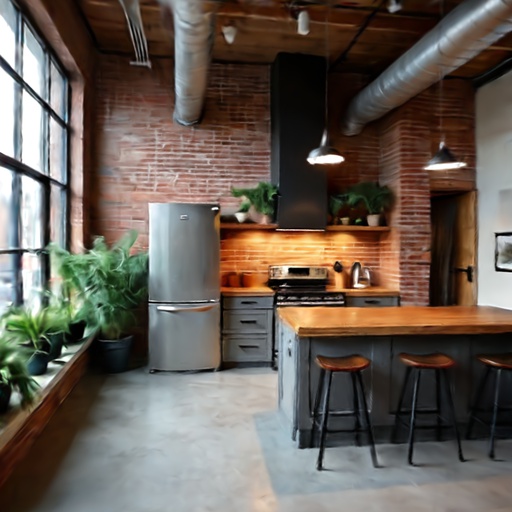} &
\includegraphics[height=0.15\textwidth]{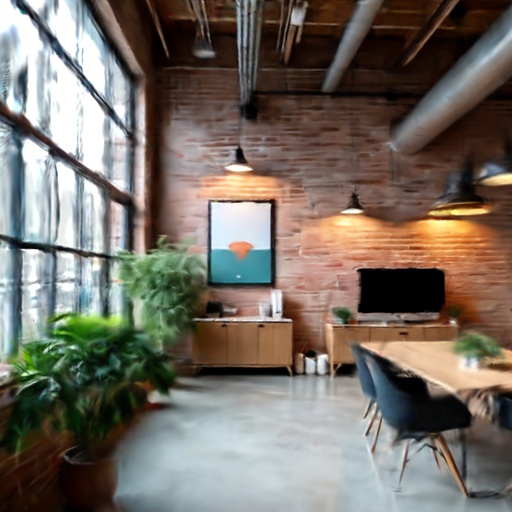} \\

&  %
\includegraphics[height=0.15\textwidth]{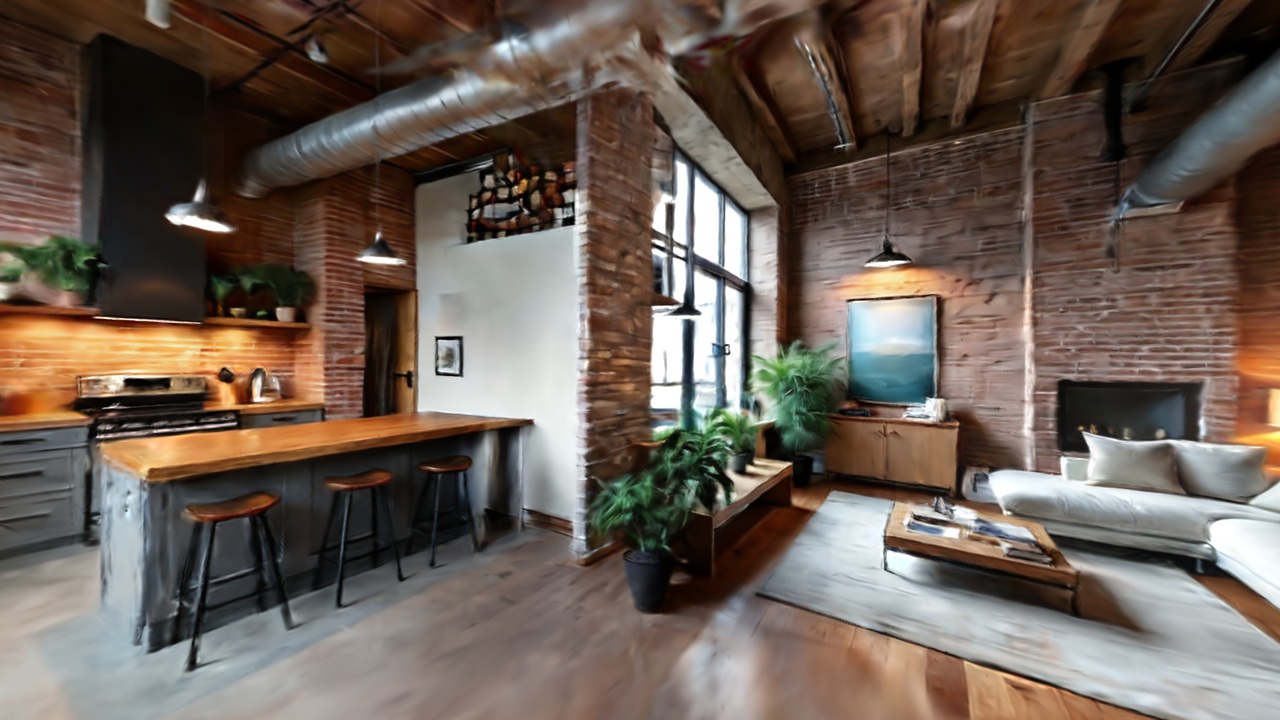} &
\includegraphics[height=0.15\textwidth]{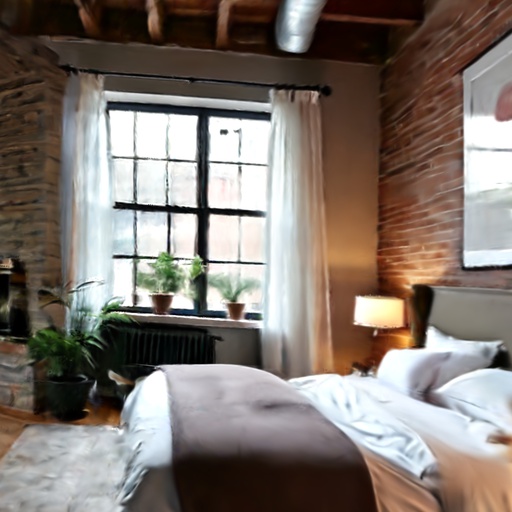} &
\includegraphics[height=0.15\textwidth]{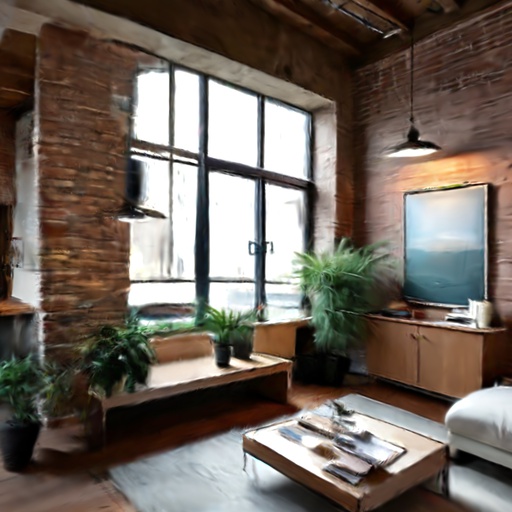} &
\includegraphics[height=0.15\textwidth]{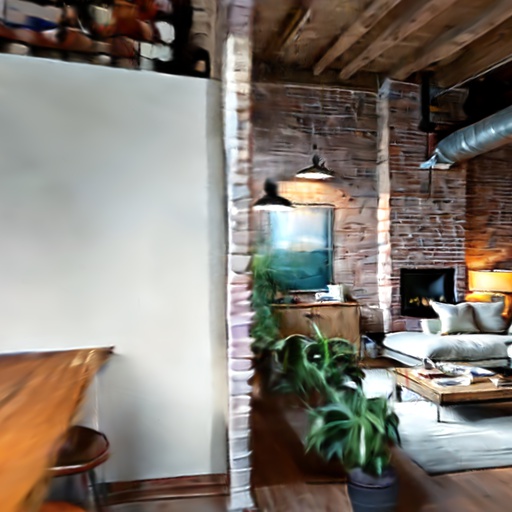} &
\includegraphics[height=0.15\textwidth]{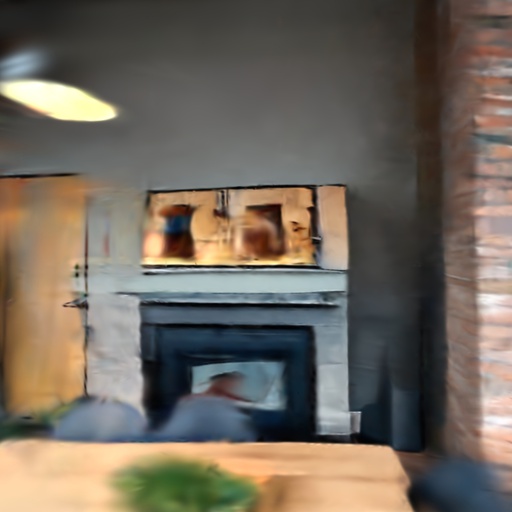} \\

\multirow{2}{*}[0.07\textwidth]{\rotatebox{90}{\textit{"Gothic Revival mansion"}}} &
\includegraphics[height=0.15\textwidth]{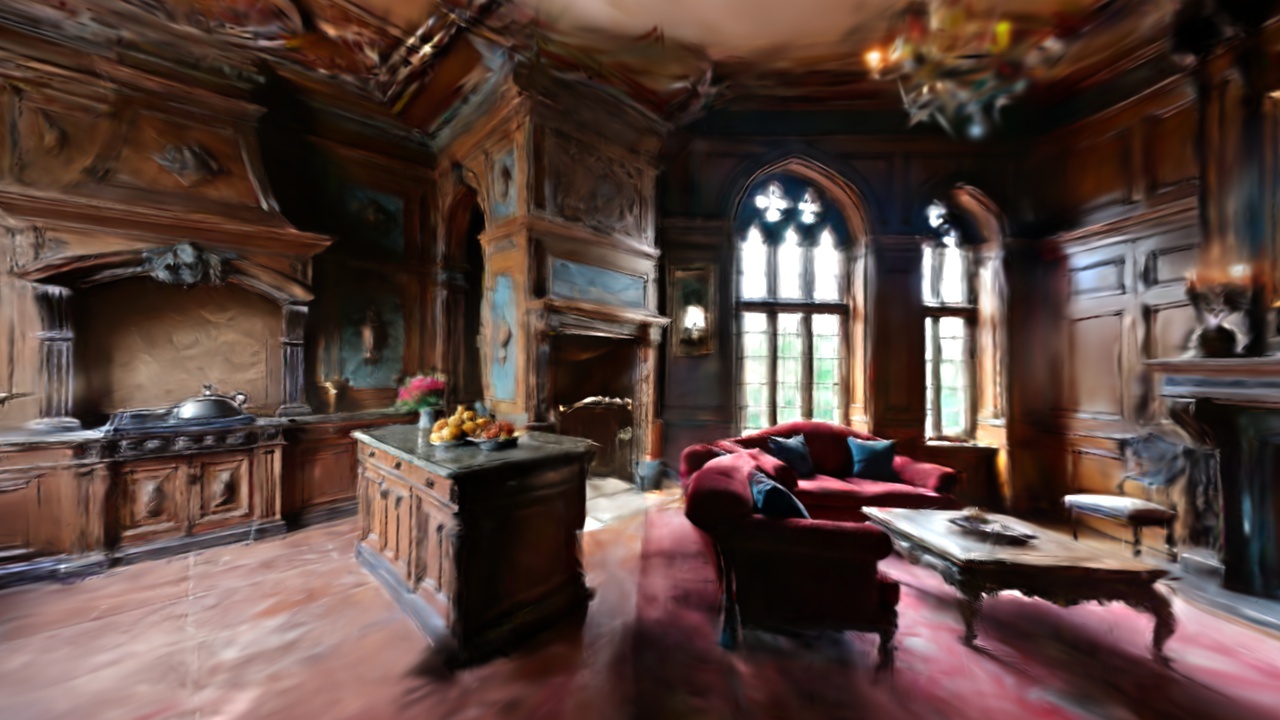} &
\includegraphics[height=0.15\textwidth]{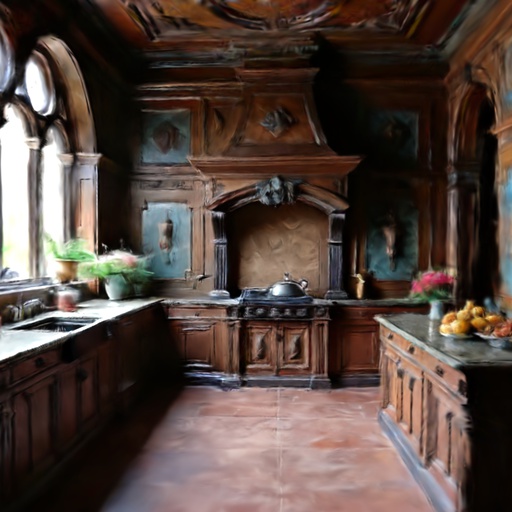} &
\includegraphics[height=0.15\textwidth]{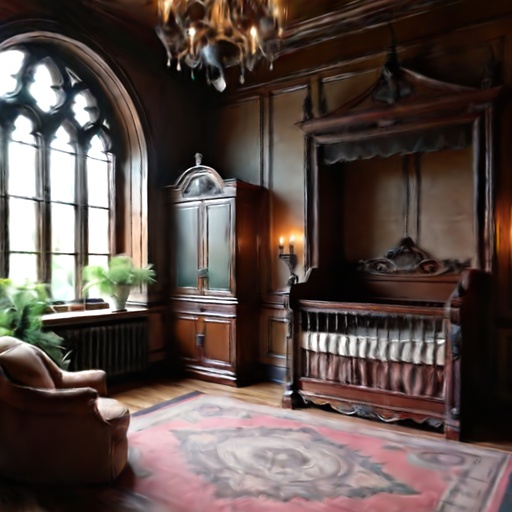} &
\includegraphics[height=0.15\textwidth]{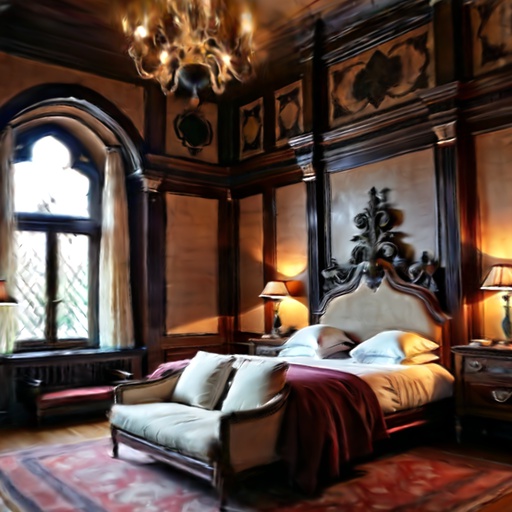} &
\includegraphics[height=0.15\textwidth]{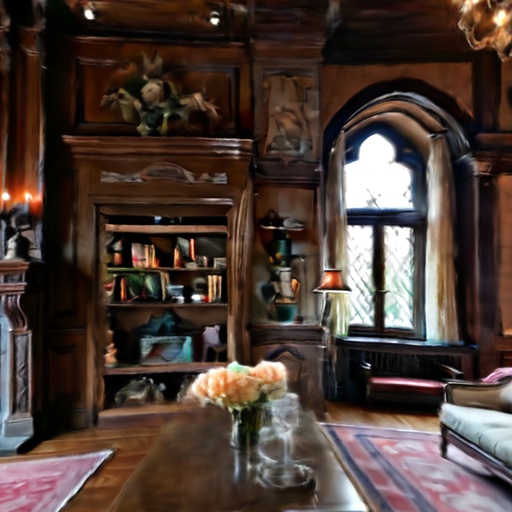} \\

&  %
\includegraphics[height=0.15\textwidth]{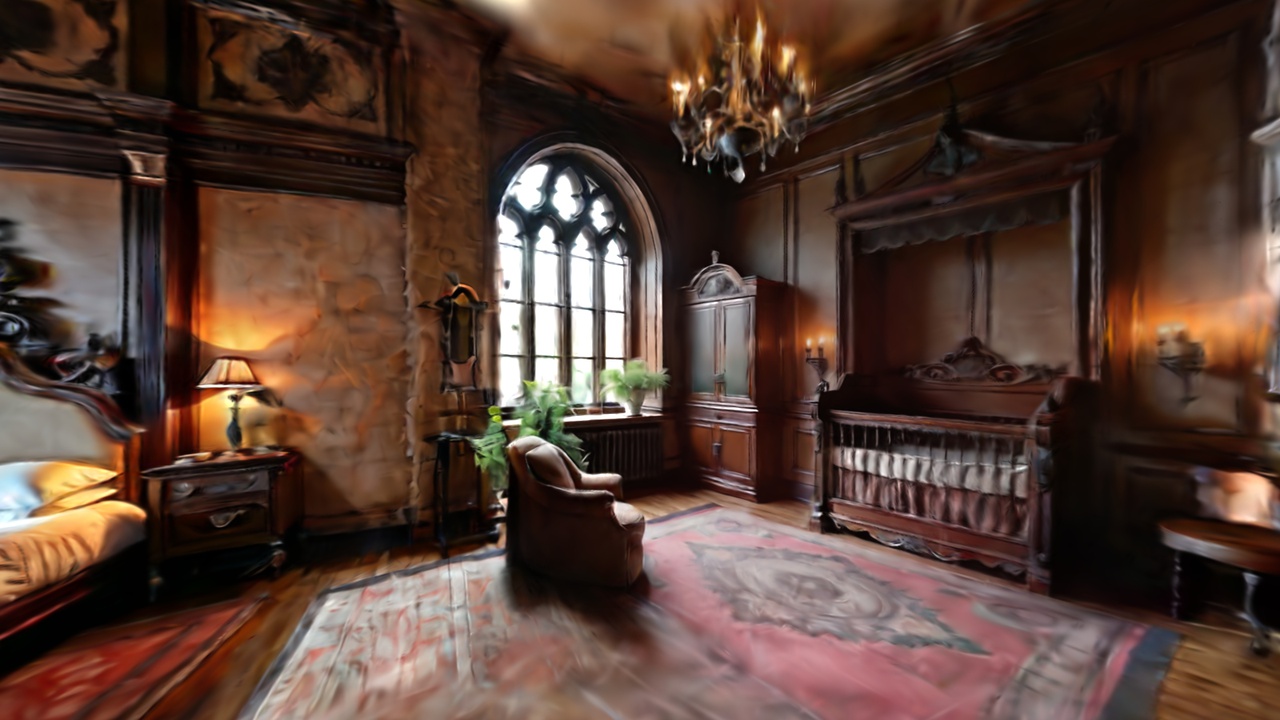} &
\includegraphics[height=0.15\textwidth]{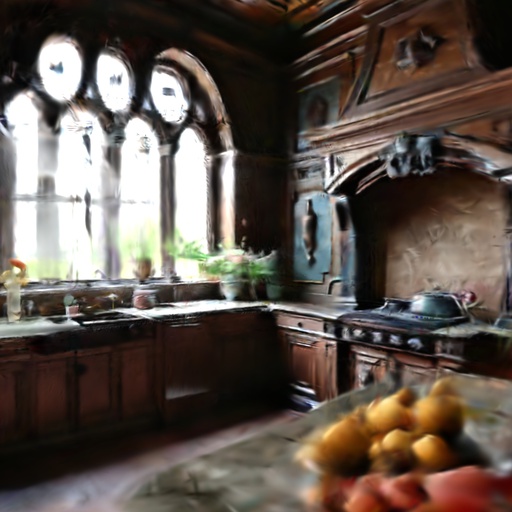} &
\includegraphics[height=0.15\textwidth]{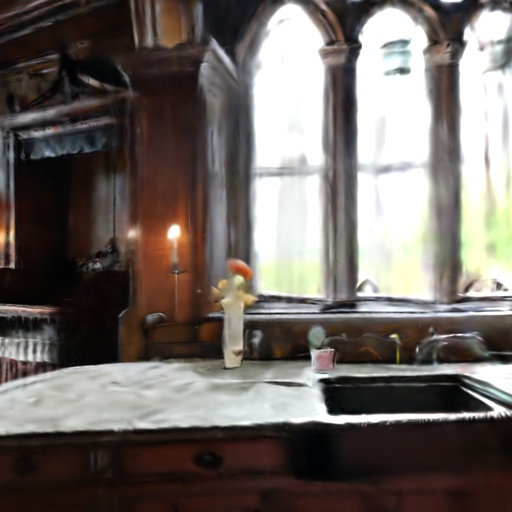} &
\includegraphics[height=0.15\textwidth]{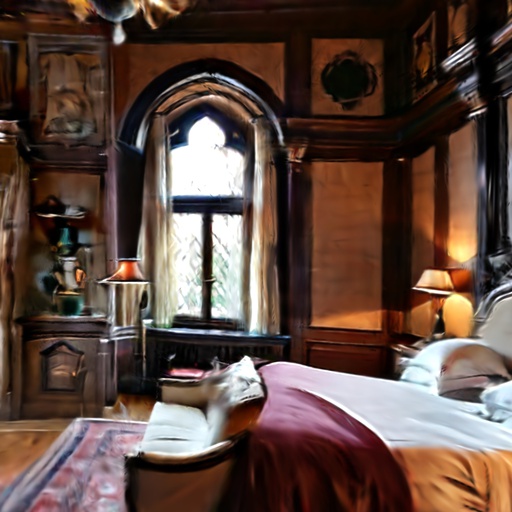} &
\includegraphics[height=0.15\textwidth]{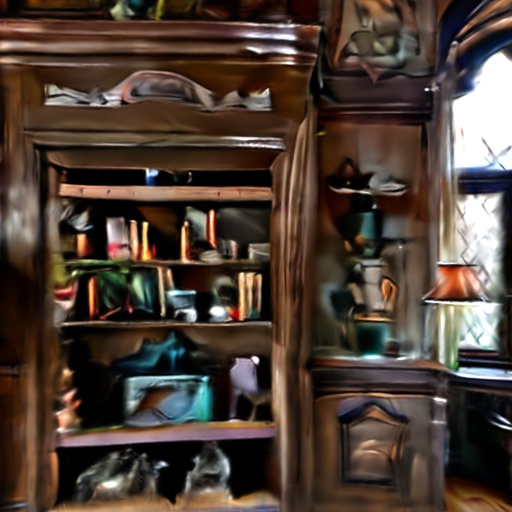} \\

\multirow{2}{*}[0.07\textwidth]{\rotatebox{90}{\textit{"Boho style beach house"}}} &
\includegraphics[height=0.15\textwidth]{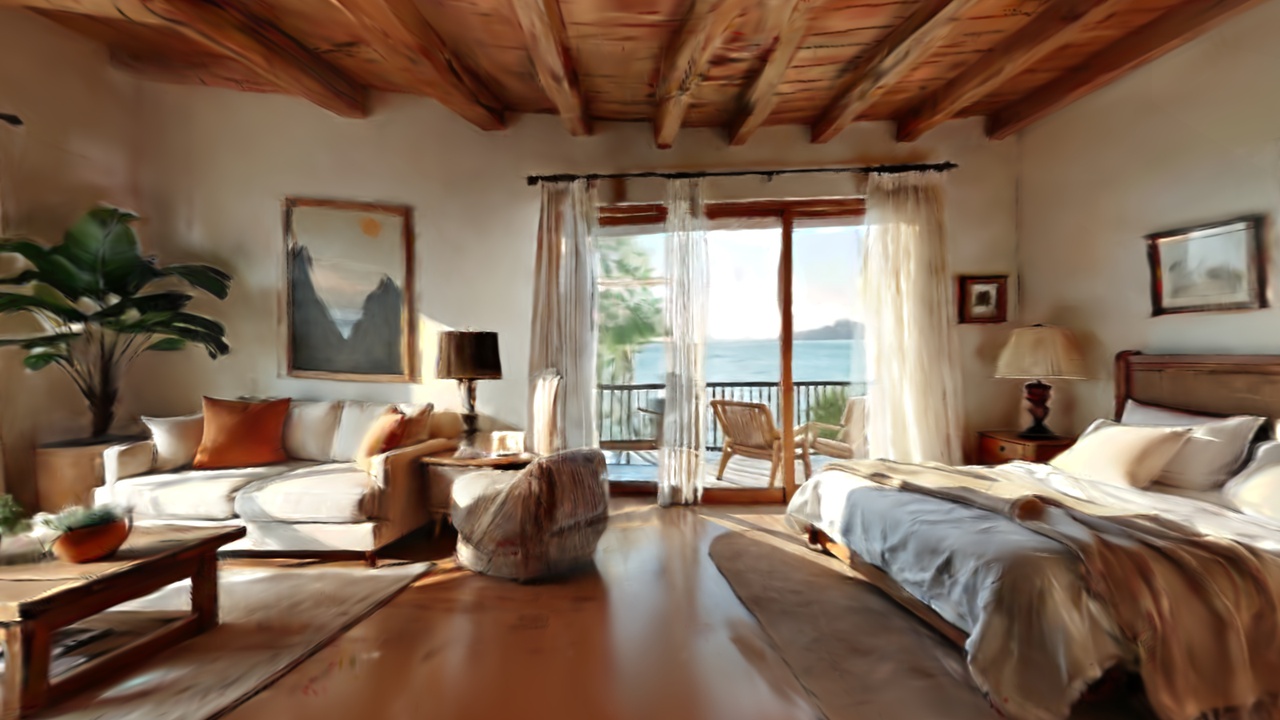} &
\includegraphics[height=0.15\textwidth]{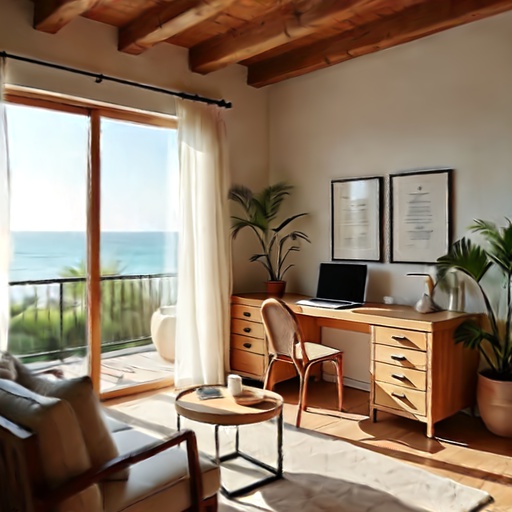} &
\includegraphics[height=0.15\textwidth]{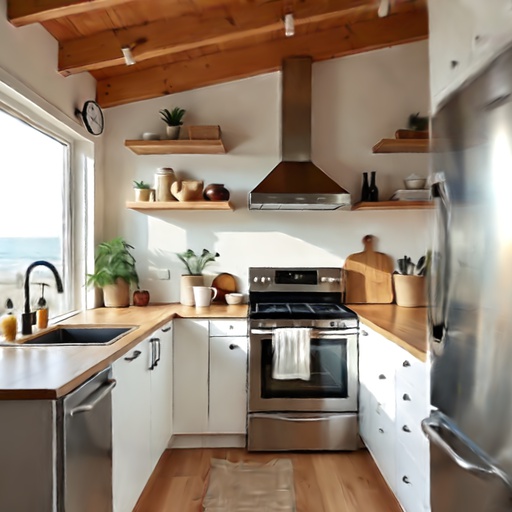} &
\includegraphics[height=0.15\textwidth]{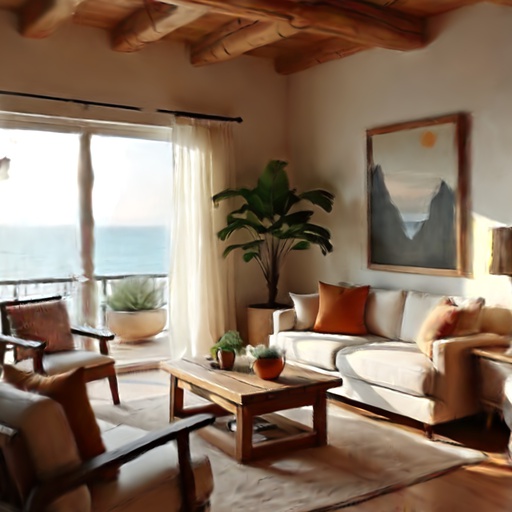} &
\includegraphics[height=0.15\textwidth]{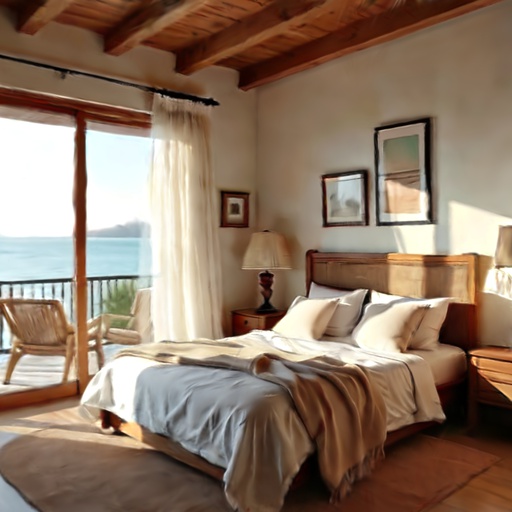} \\

&  %
\includegraphics[height=0.15\textwidth]{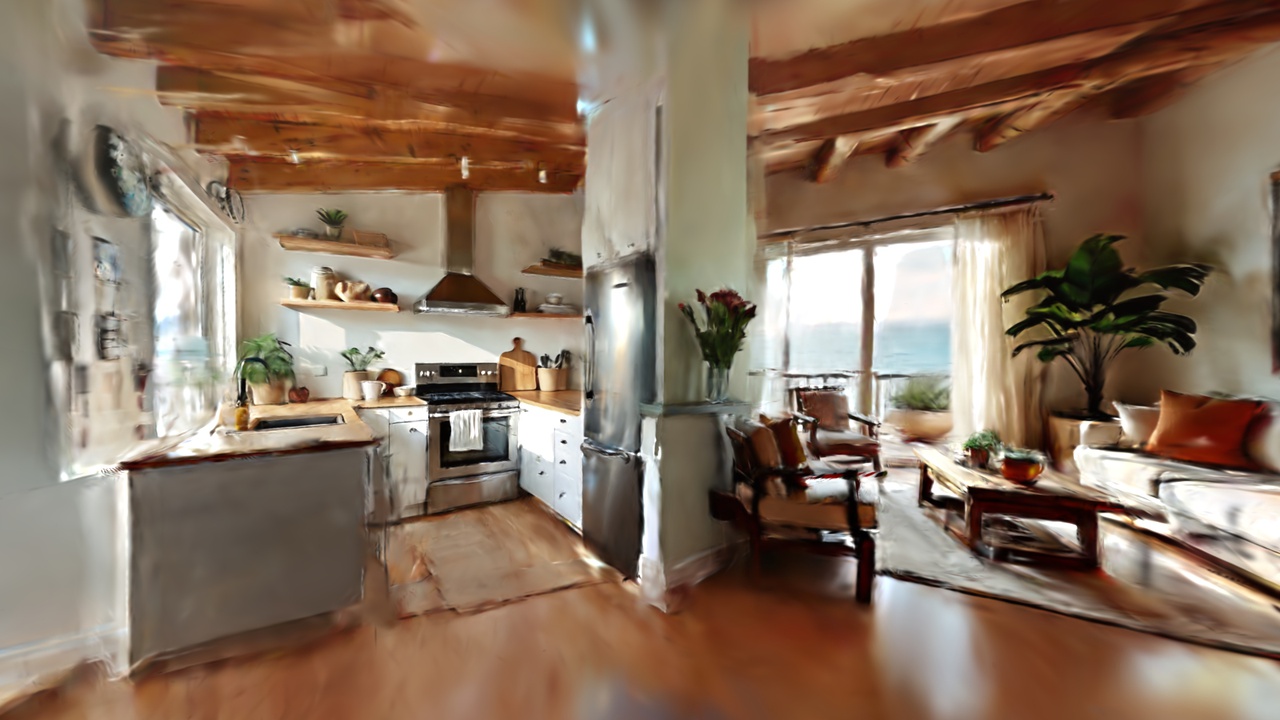} &
\includegraphics[height=0.15\textwidth]{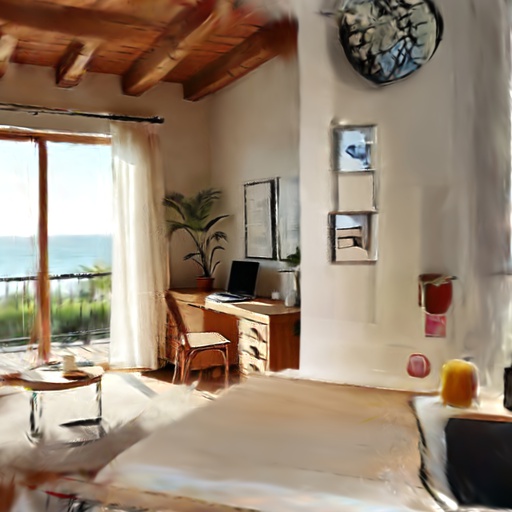} &
\includegraphics[height=0.15\textwidth]{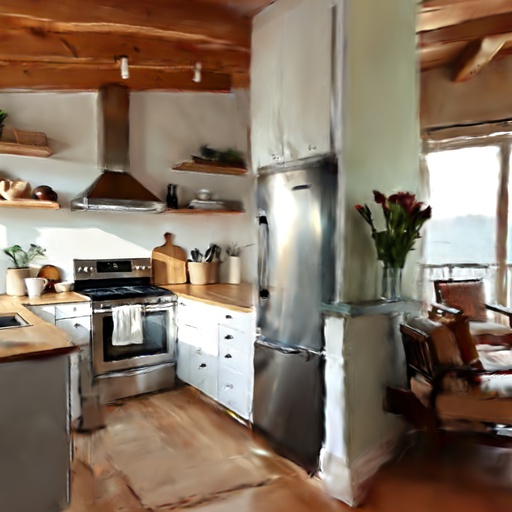} &
\includegraphics[height=0.15\textwidth]{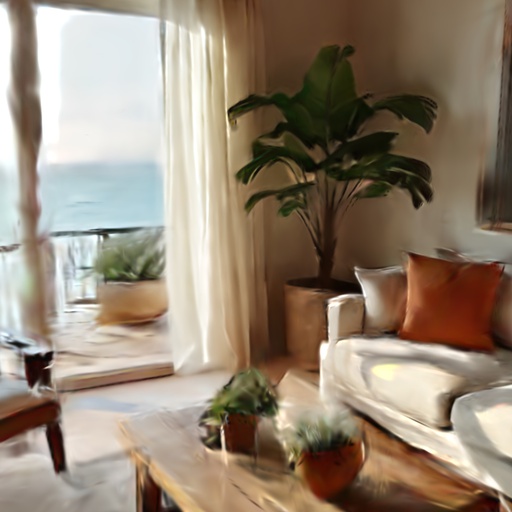} &
\includegraphics[height=0.15\textwidth]{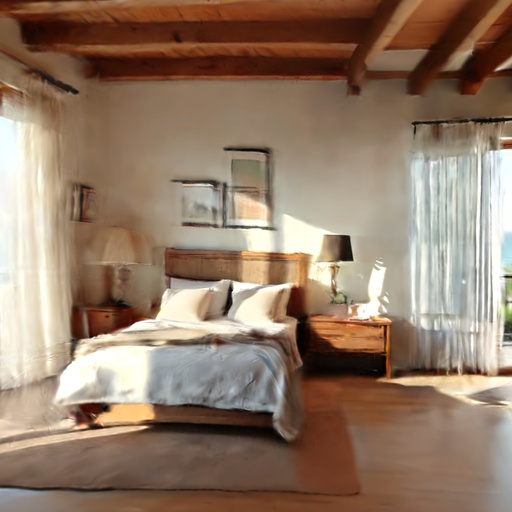} \\
\end{tabular}

\caption{
\textbf{Additional world generation results of our method.}
We visualize scene overview renderings close to the scene center facing in two opposite directions.
Additionally, we render novel views far beyond the scene center, that explore the different generated areas in more detail.
This showcases the possibility to interactively explore our worlds from arbitrary viewpoints.
Please see the supplementary video for animated flythroughs of our generated scenes.
}
\Description{
\textbf{Additional world generation results of our method.}
We visualize scene overview renderings close to the scene center facing in two opposite directions.
Additionally, we render novel views far beyond the scene center, that explore the different generated areas in more detail.
This showcases the possibility to interactively explore our worlds from arbitrary viewpoints.
Please see the supplementary video for animated flythroughs of our generated scenes.
}
\label{fig:ours-scenes-suppl-3}
\end{figure*}